\documentclass{article}
\pdfoutput=1
\usepackage{url}
\usepackage{amsmath, amssymb, bm}
\usepackage{bbm}
\usepackage{color,soul}
\usepackage{graphicx}
\usepackage{booktabs}
\usepackage{soul}
\usepackage[export]{adjustbox}
\usepackage{algorithm,algorithmic}
\usepackage{xspace}
\newcommand*{\eg}{e.g.\@\xspace}

\newcommand*{\wrt}{w.r.t.\@\xspace}
\newcommand*{\iid}{i.i.d.\@\xspace}
\makeatletter
\newcommand*{\etc}{%
	\@ifnextchar{.}%
	{etc}%
	{etc.\@\xspace}%
}
\makeatother

\newcommand{\approptoinn}[2]{\mathrel{\vcenter{
			\offinterlineskip\halign{\hfil$##$\cr
				#1\propto\cr\noalign{\kern2pt}#1\sim\cr\noalign{\kern-2pt}}}}}

\newcommand{\appropto}{\mathpalette\approptoinn\relax}

\def\*#1{\mathbf{#1}}
\DeclareMathOperator*{\argmax}{arg\,max}
\newcommand{\dataset}{\*D}
\newcommand{\param}{\bm{\omega}}
\newcommand{\parami}{\hat{\bm{\omega}}_i}

\newcommand{\stochi}{\parami}
\newcommand{\dparam}{\text{d}\param}
\newcommand{\samplesmall}{x}
\newcommand{\labsmall}{y}
\newcommand{\sample}{\*\samplesmall}
\newcommand{\lab}{\*\labsmall}
\newcommand{\samplei}{\sample_i}
\newcommand{\hids}{h}
\newcommand{\hid}{\*\hids}
\newcommand{\labi}{\lab_i}
\newcommand{\fomegax}{f_{\param}(\sample)}
\newcommand{\fomegaxsample}{f_{\parami}(\sample)}
\newcommand{\fomegaxy}{f_{\param}(\sample,\lab)}
\newcommand{\estlabel}{\bm{\hat{\lab}}}
\newcommand{\posterior}{p(\param|\dataset)}
\newcommand{\prior}{p(\param)}
\newcommand{\inference}{p(\lab|\sample,\dataset) = \int \fomegaxy \posterior d\param}
\newcommand{\approxparam}{\bm{\theta}}
\newcommand{\q}{q_{\approxparam}}
\newcommand{\qparam}{\q(\param)}
\newcommand{\KLposterior}{\text{KL}(\qparam||\posterior)}
\newcommand{\KLprior}{\text{KL}(\qparam||\prior)}
\newcommand{\LogLikelihood}{\ln p(\labi|f_{\param}(\samplei))}
\newcommand{\LogLikelihoodf}{\ln f_{\param}(\samplei,\labi)}
\newcommand{\LogLikelihoodz}{\ln f_{\{\approxparam,\stochi\}}(\samplei,\labi)}
\newcommand{\expLogLikelihood}{\int\qparam\LogLikelihood\dparam}
\newcommand{\expLogLikelihoodf}{\int\qparam\LogLikelihoodf\dparam}

\newcommand{\expLogLikelihoodMCIf}{\ln f_{\parami}(\samplei,\labi)}

\newcommand{\dnweight}{\*W}
\newcommand{\dnweighti}[1]{\dnweight^{#1}}
\newcommand{\activ}{a}
\newcommand{\expec}{\mathbb{E}}
\newcommand{\mean}{\bm{\mu}}
\newcommand{\means}{\mu}
\newcommand{\var}{\bm{\sigma}}
\newcommand{\vars}{\sigma}
\newcommand{\batch}{\*{B}}
\newcommand{\reg}{\Omega}
\newcommand{\regparam}{\reg(\param)}
\newcommand{\regapproxparam}{\reg(\approxparam)}

\newcommand{\qthetaomega}{q_{\bm{\theta}}(\param)}
\newcommand{\KLomega}{\text{KL}(\qthetaomega||\posterior)}
\newcommand{\likelihood}{p(\*Y|\*X,\param)}
\newcommand{\domega}{\text{d}\param}
\newcommand{\KLomegap}{\text{KL}(\qthetaomega||\prior)}
\newcommand{\LVItheta}{\mathcal{L}_{\text{VA}}(\bm{\theta})}
\newcommand{\expLogLikelihoodi}{\int_{\param}\qthetaomega\ln p(\labi|f_{\param}(\samplei))\domega}
\newcommand{\subsample}{-\frac{N}{M}\sum_{i\in B}}
\newcommand{\expLogLikelihoodKWProb}{\ln p(\labi|f_{g(\bm{\theta},\epsilon)}(\samplei))}
\newcommand{\expLogLikelihoodKW}{\subsample\int_{\epsilon} p(\epsilon)\expLogLikelihoodKWProb\text{d}\epsilon}

\newcommand{\myparagraph}[1]{\paragraph{#1}\mbox{}\\}
\newcommand{\weightedsum}{\dnweight^{(j)}\sample_{\text{m}}}
\newcommand{\samplevarnum}{\Sigma_{\text{m}=1}^{\text{M}}(\weightedsum-\mu_{\text{B}})^2}
\newcommand{\samplevar}{\frac{\samplevarnum}{M}}
\newcommand{\bigosimple}{\mathcal{O}\Bigg[\Big(\samplevar-\sigma^2\Big)^2\Bigg]}
\newcommand{\bigo}{\mathcal{O}\Bigg[\sqrt{M}\Big(\samplevar-\sigma^2\Big)^2\Bigg]}

\newcommand{\approxpredictive}{p^{*}(\lab|\sample,\dataset)}

\newcommand{\prioromegai}{p(\omega_{i})}
\newcommand{\qthetaomegai}{q_{\bm{\theta}}(\omega_{i})}
\newcommand{\qthetaomegaj}{q_{\bm{\theta}}(\omega_{j})}
\newcommand{\KLomegapomegai}{\text{KL}(\qthetaomegai||\prioromegai)}%
\newcommand{\pdvtheta}{\frac{\partial}{\partial \theta}}
\newcommand{\pdvthetak}{\frac{\partial}{\partial \theta_k}}

\newcommand{\fomegaxi}{f_{\param}(\sample_i)}
\newcommand{\paramj}{\hat{\bm{\omega}}_j}
\newcommand{\fomegaxisamplej}{f_{\paramj}(\sample_i)}
\newcommand{\fomegaxxi}{f_{\param}(x_i)}

\newcommand{\xmean}{\sum_{\bm{x}\in\textbf{D}}\frac{\bm{x}_k}{N}}

\newcommand{\qthetamean}{q_{\bm{\theta}}(\means_{\batch}^u)}
\newcommand{\qthetastd}{q_{\bm{\theta}}(\vars_{\batch}^u)}
\newcommand{\priormean}{p(\means_{\batch}^u)}
\newcommand{\priorstd}{p(\vars_{\batch}^u)}
\newcommand{\KLmean}{\text{KL}(\qthetamean||\priormean)}
\newcommand{\KLstd}{\text{KL}(\qthetastd||\priorstd)}
\newcommand{\pdvWuk}{\frac{\partial}{\partial \textbf{W}^{(u,k)}}}
\newcommand{\pdvWik}{\frac{\partial}{\partial \textbf{W}^{(u,i)}}}
\newcommand{\covterms}{\sum_{i=1}^{K}\textbf{W}^{(u,i)}\text{Cov}(x_i, x_k)}
\newcommand{\wux}{\textbf{W}^{(u)}\bm{x}}
\newcommand{\wuxbar}{\textbf{W}^{(u)}\bm{\bar{x}}}
\newcommand{\xbark}{\bm{\bar{x}}_k}
\newcommand{\fourthmoment}{\expec[(\wux-\means^u)^4]}

\makeatletter
\newcommand{\ALOOP}[1]{\ALC@it\algorithmicloop\ #1%
	\begin{ALC@loop}}
	\newcommand{\ENDALOOP}{\end{ALC@loop}\ALC@it\algorithmicendloop}
\renewcommand{\algorithmicrequire}{\textbf{Input:}}
\renewcommand{\algorithmicensure}{\textbf{Output:}}

\makeatother

\usepackage{hyperref}

\usepackage[accepted]{icml2018}

\icmltitlerunning{Bayesian Uncertainty Estimation for Batch Normalized Deep Networks}

\begin{document}

\twocolumn[
\icmltitle{Bayesian Uncertainty Estimation for Batch Normalized Deep Networks}



\icmlsetsymbol{equal}{*}

\begin{icmlauthorlist}
\icmlauthor{Mattias Teye}{kth,bud,equal}
\icmlauthor{Hossein Azizpour}{kth,equal}
\icmlauthor{Kevin Smith}{kth,sci}
\end{icmlauthorlist}

\icmlaffiliation{kth}{School of Electrical Engineering and Computer Science, KTH Royal Institute of Technology, Stockholm, Sweden}
\icmlaffiliation{sci}{Science for Life Laboratory}
\icmlaffiliation{bud}{Current address: Electronic Arts, SEED, Stockholm, Sweden. This work was carried out at Budbee AB.}

\icmlcorrespondingauthor{Kevin Smith}{ksmith@kth.se}

\icmlkeywords{Machine Learning, ICML}

\vskip 0.3in
]



\printAffiliationsAndNotice{\icmlEqualContribution} 

\begin{abstract}
  We show that training a deep network using batch normalization is equivalent to approximate inference in Bayesian models. We further demonstrate that this finding allows us to make meaningful estimates of the model uncertainty using conventional architectures, without modifications to the network or the training procedure. Our approach is thoroughly validated by measuring the quality of uncertainty in a series of empirical experiments on different tasks. It outperforms baselines with strong statistical significance, and displays competitive performance with recent Bayesian approaches.
\end{abstract}

\section{Introduction}
Deep learning has dramatically advanced the state of the art in a number of domains. Despite their unprecedented discriminative power, deep networks are prone to make mistakes. Nevertheless, they can already be found in settings where errors carry serious repercussions such as autonomous vehicles \citep{chen2016monocular} and high frequency trading. We can soon expect automated systems to screen for various types of cancer \citep{Esteva2017, shen2017end} and  diagnose biopsies \citep{djuric2017precision}. As autonomous systems based on deep learning are increasingly deployed in settings with the potential to cause physical or economic harm, we need to develop a better understanding of when we can be confident in the estimates produced by deep networks, and when we should be less certain. 

Standard deep learning techniques used for supervised learning lack methods to account for uncertainty in the model.
This can be problematic when the network encounters conditions it was not exposed to during training, or if the network is confronted with adversarial examples \cite{goodfellow2014explaining}. 
When exposed to data outside the distribution it was trained on, the network is forced to extrapolate, which can lead to unpredictable behavior. 

\begin{figure}[t]
\centering
\includegraphics[width=1\linewidth]{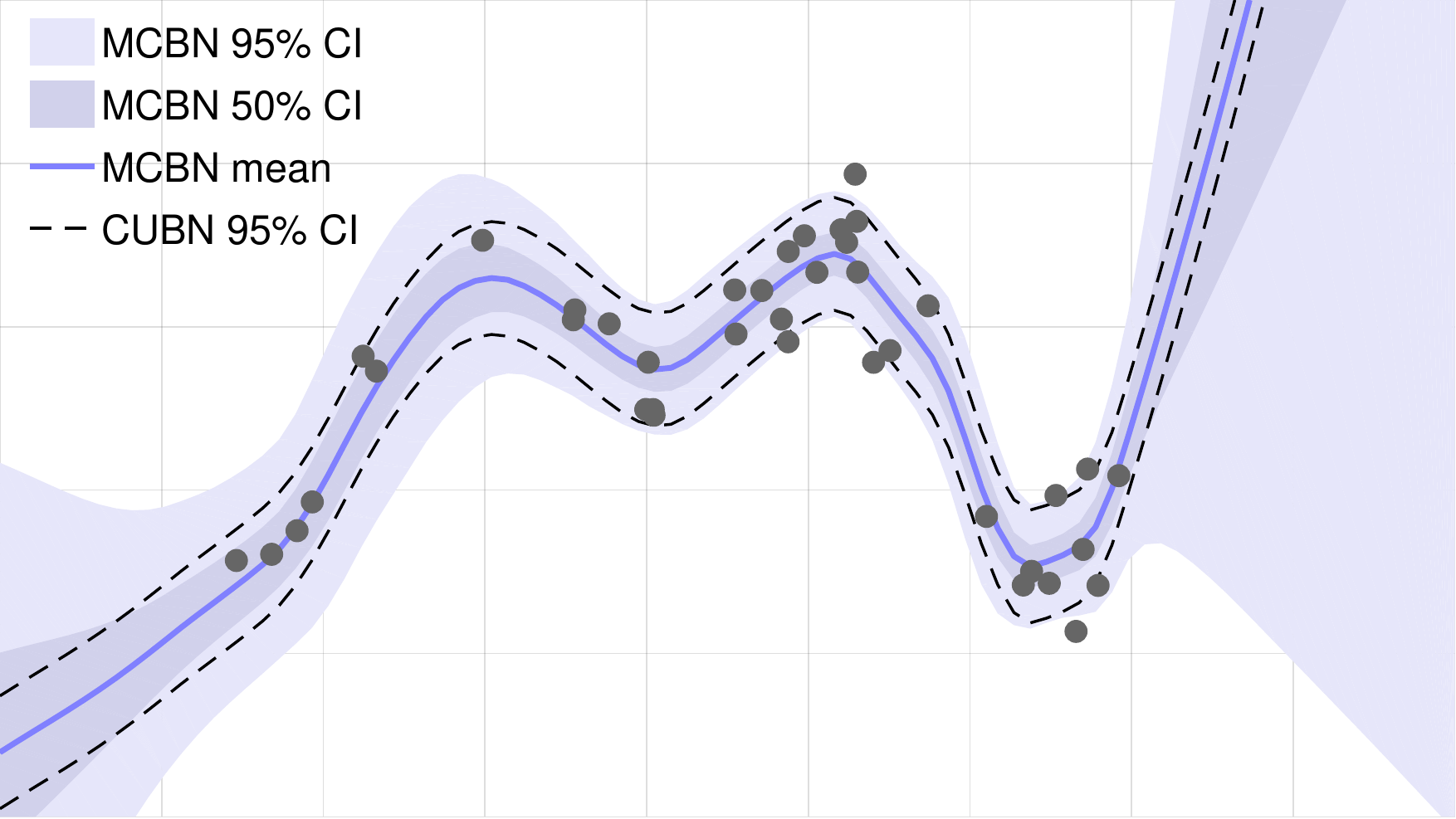}

\vspace{-3mm}
\caption{\textbf{Training a deep network using batch normalization is equivalent to approximate inference  in  Bayesian  models.} Thus, uncertainty estimates can be obtained from any network using BN through a simple procedure. At inference, several mini-batches are constructed by taking random samples to accompany the query. The mean and variance of the outputs are used to estimate the predictive distribution (MCBN). Here, we show results on a toy dataset from a network with three hidden layers (30 units per layer). Training data is depicted as dots. The solid line is the predictive mean of 500 stochastic forward passes and the shaded areas represent the model's uncertainty. The dashed lines depict a minimal baseline for uncertainty (CUBN), see Section \ref{sec:metrics}.} 
\label{fig:toydata}
\vspace{-4mm}
\end{figure}

If the network can provide information about its uncertainty in addition to its point estimate, disaster may be avoided. In this work, we focus on estimating such predictive uncertainties in deep networks (Figure \ref{fig:toydata}).

The Bayesian approach provides a theoretical framework for modeling uncertainty \citep{Ghahramani2015}, which has prompted several attempts to extend neural networks (NN) into a Bayesian setting. Most notably, Bayesian neural networks (BNNs) have been studied since the 1990's \citep{neal2012bayesian}, but do not scale well and struggle to compete with modern deep learning architectures. Recently, \cite{Gal2015a} developed a practical solution to obtain uncertainty estimates by casting \textit{dropout} training in conventional deep networks as a Bayesian approximation of a Gaussian Process (its correspondence to a general approximate Bayesian model was shown in \citep{Gal2016Uncertainty}). They showed that \textit{any network} trained with dropout is an approximate Bayesian model, and uncertainty estimates can be obtained by computing the variance on multiple predictions with different dropout masks.

The inference in this technique, called \textit{Monte Carlo Dropout} (MCDO), has an attractive quality: it can be applied to any pre-trained networks with dropout layers. Uncertainty estimates come (nearly) for free. However, not all architectures use dropout, and most modern networks have adopted other regularization techniques. \textit{Batch normalization} (BN), in particular, has become widespread thanks to its ability to stabilize learning with improved generalization \citep{Ioffe2015}.

An interesting aspect of BN is that the mini-batch statistics used for training each iteration depend on randomly selected batch members. We exploit this stochasticity and show that training using batch normalization, like dropout, can be cast as an approximate Bayesian inference.
We demonstrate how this finding allows us to make meaningful estimates of the model uncertainty in a technique we call \textit{Monte Carlo Batch Normalization} (MCBN)  (Figure \ref{fig:toydata}). The method we propose can be applied to any network using standard batch normalization. 

We validate our approach by empirical experiments on a variety of datasets and tasks, including regression and image classification. We measure uncertainty quality relative to a baseline of fixed uncertainty, and show that MCBN outperforms the baseline on nearly all datasets with strong statistical significance. We also show that the uncertainty quality of MCBN is on par with other recent approximate Bayesian networks. 

\section{Related Work}

Bayesian models provide a natural framework for modeling uncertainty, and several approaches have been developed to adapt NNs to Bayesian reasoning. A common approach is to place a prior distribution (often a Gaussian) over each parameter. The resulting model corresponds to a Gaussian process for infinite parameters \citep{neal1995bayesian}, and a Bayesian NN \citep{mackay1992practical} for a finite number of parameters. Inference in BNNs is difficult however \citep{Gal2016Uncertainty}, so focus has thus shifted to techniques that approximate the posterior, \textit{approximate BNNs}. Methods based on variational inference (VI) typically rely on a fully factorized approximate distribution \citep{Kingma2014, hinton1993keeping}, but often do not scale. To alleviate these difficulties, \cite{Graves2011a} proposed a model using sampling methods to estimate a factorized posterior. Probabilistic backpropagation (PBP), estimates a factorized posterior via expectation propagation \citep{hernandez2015probabilistic}. 

Using several strategies to address scaling issues, Deep Gaussian Processes show superior performance in terms of RMSE and uncertainty quality compared to state-of-the-art approximate BNNs   \citep{Bui2016}\footnote{By uncertainty quality, we refer to predictive probability distributions as measured by PLL and CRPS.}. Another recent approach to Bayesian learning, Bayesian hypernetworks, use a NN to learn a distribution of parameters over another network \citep{krueger2017bayesian}. Multiplicative Normalizing Flows for variational Bayesian networks (MNF) \citep{pmlr-v70-louizos17a} is a recent model that formulates a posterior dependent on auxiliary variables. MNF achieves a highly flexible 
posterior by the application of normalizing flows to the auxiliary variables.

Although these recent techniques address some of the difficulties with approximate BNNs, they all require modifications to the architecture or the way networks are trained, as well as specialized knowledge from practitioners. Recently, \cite{Gal2015a} showed that a network trained with dropout implicitly performs the VI objective. Therefore \textit{any} network trained with dropout can be treated as an approximate Bayesian model by making multiple predictions through the network while sampling different dropout masks for each prediction. The mean and variance of the predictions are used in the estimation of the mean and variance of the predictive distribution \footnote{This technique is referred to as ``MC Dropout'' in the original work, though we refer to it here as MCDO.}. 


\section{Method}

In the following, we introduce Bayesian models and a variational approximation using Kullback-Leibler (KL) divergence following \cite{Gal2016Uncertainty}. We continue by showing that a batch normalized deep network can be seen as an approximate Bayesian model. Employing theoretical insights and empirical analysis, we study the induced prior on the parameters when using batch normalization. Finally, we describe the procedure for estimating the uncertainty of a batch normalized  network's output.\footnote{While the method applies to FC or Conv layers, the induced prior from weight decay (Section \ref{section:prior}) is studied for FC layers.}

\subsection{Bayesian Modeling}

We assume a finite training set $\*D=\{(\samplei,\labi)\}_{i=1:N}$ where each $(\samplei,\labi)$ is a sample-label pair. Using $\dataset$, we are interested in learning an inference function $\fomegaxy$ with parameters $\param$. In deterministic models, the estimated label $\estlabel$ is obtained as follows:
\begin{align*}
\estlabel=\argmax_{\lab}\fomegaxy
\end{align*}
In probabilistic models we let $\fomegaxy = p(\lab|\sample,\param)$. In Bayesian modeling, in contrast to finding a point estimate of the model parameters, the idea is to estimate an (approximate) posterior distribution of the model parameters $\posterior$ to be used for probabilistic prediction:
\begin{align*}
\inference
\end{align*}

The predicted label, $\estlabel$, can then be accordingly obtained by sampling $p(\lab|\sample,\dataset)$ or taking its maxima.

\paragraph{Variational Approximation}
In approximate Bayesian modeling, a common approach is to learn a parameterized approximating distribution $\qparam$ that minimizes $\KLposterior$; the Kullback-Leibler divergence of the true posterior \wrt its approximation. Minimizing this KL divergence is equivalent to the following minimization while being free of the data term $p(\dataset)$ \footnote{Achieved by constructing the Evidence Lower Bound, called ELBO, and assuming \iid observation noise; details can be found in Appendix Section 1.1.}:
\begin{align*}
\mathcal{L}_{\text{VA}}(\approxparam):=& -\sum_{i=1}^{N}\expLogLikelihoodf\\
&+\KLprior
\end{align*}

During optimization, we want to take the derivative of the expected likelihood \wrt the learnable parameters $\bm{\theta}$. We use the same MC estimate as in \cite{Gal2016Uncertainty} (explained in Appendix Section 1.1), such that one realized $\parami$ is taken for each sample $i$ \footnotemark. Optimizing over mini-batches of size $M$, the approximated objective becomes:
\footnotetext{While a MC integration using a single sample is a weak approximation, in an iterative optimization for $\approxparam$ several samples will be taken over time.}

{\small
	\begin{align}
	\hat{\mathcal{L}}_{\text{VA}}(\approxparam):= -\frac{N}{M}\sum_{i=1}^{M}\expLogLikelihoodMCIf+\KLprior
	\label{eq:approxVA}
	\end{align}
}
The first term is the data likelihood and the second term is the divergence of the prior \wrt the approximated posterior. 

\subsection{Batch Normalized Deep Nets as Bayesian Modeling}

We now describe the optimization procedure of a deep network with batch normalization and draw the resemblance to the approximate Bayesian modeling in Eq (\ref{eq:approxVA}).

The inference function of a feed-forward deep network with $L$ layers can be described as:
\begin{align*}
\fomegax = \dnweighti{L}\activ(\dnweighti{L-1}...\activ(\dnweighti{2}\activ(\dnweighti{1}\sample))
\end{align*}

where $\activ(.)$ is an element-wise nonlinearity function and $\dnweighti{l}$ is the weight vector at layer $l$. Furthermore, we denote the input to layer $l$ as $\sample^l$ with $\sample^1=\sample$ and we then set $\hid^l=\dnweighti{l}\sample^{l}$. Parenthesized super-index for matrices (e.g. $\dnweight^{(j)}$) and vectors (e.g. $\samplesmall^{(j)}$) indicates $j$th row and element respectively. Super-index $u$ refers to a specific unit at layer $l$, (\eg $\dnweighti{u}=\dnweighti{l,(j)}, \hids^u=\hids^{l,(j)}$). \footnotemark 
\footnotetext{For a (softmax) classification network, $\fomegax$ is a vector with $\fomegaxy=\fomegax^{(\lab)}$, for regression networks with \iid Gaussian noise we have $\fomegaxy=\mathcal{N}(\fomegax,\tau^{-1}\*I )$.}

\paragraph{Batch Normalization}
Each layer of a deep network is constructed by several linear units whose parameters are the rows of the weight matrix $\dnweight$. Batch normalization is a unit-wise operation proposed in \cite{Ioffe2015} to standardize the distribution of each unit's input. For FC layers, it converts a unit's input $\hids^u$ in the following way:
\begin{align*}
\hat{\hids}^{u}=\frac{\hids^{u}-\expec[\hids^{u}]}{\sqrt{\text{Var}[\hids^{u}]}}
\end{align*}
where the expectations are computed over the training set during evaluation, and mini-batch during training (in deep networks, the weight matrices are often optimized using back-propagated errors calculated on mini-batches of data)\footnotemark. Therefore, during training, the estimated mean and variance on the mini-batch $\batch$ is used, which we denote by $\mean_{\batch}$ and $\var_{\batch}$ respectively. This makes the inference at training time for a sample $\sample$ a stochastic process, varying based on other samples in the mini-batch.

\footnotetext{It also learns an affine transformation for each unit with parameters $\bm{\gamma}$ and $\bm{\beta}$, omitted for brevity: $\hat{\samplesmall}^{(j)}_{\text{affine}}=\gamma^{(j)}\hat{\samplesmall}^{(j)}+\beta^{(j)}$.}

\paragraph{Loss Function and Optimization}
Training deep networks with mini-batch optimization involves a (regularized) risk minimization with the following form:

\begin{align*}
\mathcal{L}_{\text{RR}}(\param):= \frac{1}{M}\sum_{i=1}^{M}l(\estlabel_i, \lab_i)+\regparam
\end{align*}

where the first term is the empirical loss on the training data and the second term is a regularization penalty acting as a prior on model parameters $\param$. If the loss $l$ is cross-entropy for classification or sum-of-squares for regression problems (assuming \iid Gaussian noise on labels), the first term is equivalent to minimizing the negative log-likelihood:

\begin{align*}
\mathcal{L}_{\text{RR}}(\param):= -\frac{1}{M\tau}\sum_{i=1}^{M}\LogLikelihoodf+\regparam
\end{align*}

with $\tau=1$ for classification. In a network with batch normalization, the model parameters include $\{\dnweighti{1:L},\bm{\gamma}^{1:L},\bm{\beta}^{1:L}, \mean_{\batch}^{1:L},\var_{\batch}^{1:L}\}$. If we decouple the learnable parameters $\approxparam=\{\dnweighti{1:L},\bm{\gamma}^{1:L},\bm{\beta}^{1:L}\}$ from the stochastic parameters $\param=\{\mean_{\batch}^{1:L},\var_{\batch}^{1:L}\}$, we get the following objective at each step of the mini-batch optimization:
\begin{align}
\mathcal{L}_{\text{RR}}(\approxparam):= -\frac{1}{M\tau}\sum_{i=1}^{M}\LogLikelihoodz+\regapproxparam
\label{eq:NN_loglik}
\end{align}
where $\stochi$ is the means and variances for sample $i$'s mini-batch at a certain training step. Note that while $\stochi$ formally needs to be \iid for each training example, a batch normalized network samples the stochastic parameters once per training step (mini-batch). For a large number of epochs, however, the distribution of sampled batch members for a given training example converges to the i.i.d. case.

In a batch normalized network, $\qparam$ corresponds to the joint distribution of the weights, induced by the randomness of the normalization parameters $\mean_{\batch}^{1:L},\var_{\batch}^{1:L}$, as implied by the repeated sampling from $\*D$ during training. This is an approximation of the true posterior, where we have restricted the posterior to lie within the domain of our parametric network and source of randomness. With that, \textit{we can estimate the uncertainty of predictions from a trained batch normalized network using the inherent stochasticity of BN} (Section \ref{sec:predictivedistrib}).

\subsection{Prior \texorpdfstring{$\prior$}{}}
\label{section:prior}
Equivalence between the VA and BN training procedures requires $\pdvtheta$ of Eq.~(\ref{eq:approxVA}) and Eq.~(\ref{eq:NN_loglik}) to be equivalent up to a scaling factor. This is the case if $\pdvtheta\KLprior=N\tau\pdvtheta\regapproxparam$.

To reconcile this condition, one option is to let the prior $p(\bm{\omega})$ imply the regularization term $\regapproxparam$. Eq.~(\ref{eq:approxVA}) reveals that the contribution of $\KLprior$ to the optimization objective is inversely scaled with $N$. For BN, this corresponds to a model with a small $\regapproxparam$ when $N$ is large. In the limit as $N\rightarrow\infty$, the optimization objectives of Eq.~(\ref{eq:approxVA}) and Eq.~(\ref{eq:NN_loglik}) become identical with no regularization.\footnote{To prove the existence and find an expression of $\KLprior$, in Appendix Section 1.3 we find that BN approximately induces Gaussian distributions over BN units' means and standard deviations, centered around the population values given by \textbf{D}. We assume a factorized distribution and Gaussian priors, and find the corresponding $\KLprior$ components in Appendix Section 1.4 Eq. (7). These could be used to construct a custom $\regapproxparam$ for any Gaussian choice of $p(\bm{\omega})$.}

Another option is to let some $\regapproxparam$ imply $p(\bm{\omega})$. In Appendix Section 1.4 we explore this with L2-regularization, also called weight decay ($\regapproxparam=\lambda \sum_{l=1:L}||W^l||^2$). We find that unlike in MCDO \cite{Gal2016Uncertainty}, some simplifying assumptions are necessary to reconcile the VA and BN objectives with weight decay: no scale and shift applied to BN layers, uncorrelated units in each layer, BN applied on all layers, and large $N$ and $M$. Given these conditions:
\begin{align*}
p(\means_{\batch}^u)&=\mathcal{N}(\mu_{\mu,p}, \sigma_{\mu,p})\\
p(\vars_{\batch}^u)&=\mathcal{N}(\mu_{\sigma,p}, \sigma_{\sigma,p})
\end{align*}
where $\mu_{\mu,p}=0$, $\sigma_{\mu,p}\rightarrow\infty$, $\mu_{\sigma,p}=0$ and $\sigma_{\sigma,p}\rightarrow\frac{1}{2N\tau\lambda_l}$. 

This corresponds to a wide and narrow distribution on BN units' means and std. devs respectively, where $N$ accounts for the narrowness of the prior.
Due to its popularity in deep learning, our experiments in Section \ref{section:experiments} are performed with weight decay.

\subsection{Predictive Uncertainty in Batch Normalized Deep Nets}\label{sec:predictivedistrib}
In the absence of the true posterior, we rely on the approximate posterior to express an approximate predictive distribution:
\begin{align*}
\approxpredictive := \int \fomegaxy \qparam d\param
\end{align*}
Following \cite{Gal2016Uncertainty} we estimate the first (for regression and classification) and second (for regression) moments of the predictive distribution empirically (see Appendix Section 1.5 
for details):
\begin{align*}
\expec_{p^*}[\lab]&\approx\frac{1}{T}\sum_{i=1}^T \fomegaxsample \\
\text{Cov}_{p^*}[\lab]&\approx\tau^{-1}\textbf{I}+\frac{1}{T}\sum_{i=1}^T \fomegaxsample^\intercal\fomegaxsample\\
&-\expec_{p^*}[\lab]^\intercal\expec_{p^*}[\lab]
\end{align*}
where each $\parami$ corresponds to sampling the net's stochastic parameters $\param=\{\mean_{\batch}^{1:L},\var_{\batch}^{1:L}\}$ the same way as during training. Sampling $\parami$ therefore involves sampling a batch $\batch$ from the \textit{training set} and updating the parameters in the BN units, just as if we were taking a training step with $\batch$. From a VA perspective, training the network amounted to minimizing $\KLposterior$ wrt $\bm{\theta}$. Sampling $\parami$ from the training set, and keeping the size of $\batch$ consistent with the mini-batch size used during training, ensures that $\qparam$ during inference remains identical to the approximate posterior optimized during training.

The network is trained just as a regular BN network, but instead of replacing $\param=\{\mean_{\batch}^{1:L},\var_{\batch}^{1:L}\}$ with population values from $\*D$ for inference, we update these parameters stochastically, once for each forward pass.\footnote{As an alternative to using the training set $\*D$ to sample $\parami$, we could sample from the implied $\qparam$ as modeled in the Appendix. This would alleviate having to store $\*D$ for use during prediction. In our experiments we used $\*D$ to sample $\parami$ however, and leave the evaluation of the modeled $\qparam$ for future research.} Pseudocode for estimating predictive mean and variance is given in Algorithm 1.



\section{Experiments and Results}\label{section:experiments}

We assess the uncertainty quality of MCBN quantitatively and qualitatively. Our quantitative analysis relies on CIFAR10 for image classification and eight standard regression datasets, listed in Appendix Table 1. Publicly available from the UCI Machine Learning Repository \citep{UniversityofCaliforniab} and Delve \citep{Ghahramani1996}, these datasets have been used to benchmark comparative models in recent related literature (see \cite{hernandez2015probabilistic}, \cite{Gal2015a}, \cite{Bui2016} and \cite{Li2017}). We report results using standard metrics, and also propose useful upper and lower bounds to normalize these metrics for an easier interpretation in Section \ref{sec:benchmark}.

Our qualitative results include the toy dataset in Figure \ref{fig:toydata} in the style of \citep{karpathy}, {\it a new visualization of uncertainty quality} that plots test errors sorted by predicted variance (Figure \ref{fig:plots} and Appendix), and image segmentation results (Figure \ref{fig:plots} and Appendix).


\begin{algorithm}[t]
  \caption{MCBN Algorithm}\label{algo:mbcn}
  \begin{algorithmic}[1]
    \REQUIRE sample $x$, number of inferences $T$, batchsize $b$
    \ENSURE mean prediction $\hat{y}$, predictive uncertainty $\sigma^2$
    \STATE $\textbf{y} = \{\}$
    \ALOOP {for $T$ iterations}
        \STATE $B \thicksim D$ // mini batch
        \STATE $\hat{\param}=\{\mean_B, \var_B\}$ \quad  // mini batch mean and variance
        \STATE $\textbf{y} = \textbf{y} \cup f_{\hat{\param}}(x)$
    \ENDALOOP
    \STATE $\hat{y} = \mathbb{E} [\textbf{y}]$
    \STATE $\sigma^2 = \text{Cov}[\textbf{y}] + \tau^{-1}\*I$\enskip// for regression
  \end{algorithmic}
\end{algorithm}

\subsection{Metrics} \label{sec:metrics}
We evaluate uncertainty quality based on two standard metrics, described below: Predictive Log Likelihood (PLL) and Continuous Ranked Probability Score (CRPS). To improve the interpretability of the metrics, we propose to normalize them by upper and lower bounds.


\paragraph{Predictive Log Likelihood (PLL)}  Predictive Log Likelihood is widely accepted as the main uncertainty quality metric for regression \citep{hernandez2015probabilistic,Gal2015a,Bui2016,Li2017}. A key property of PLL is that it makes no assumptions about the form of the distribution. The measure is defined for a probabilistic model $\fomegax$ and a single observation $(\*y_i, \*x_i)$ as:
$$\text{PLL}(\fomegax, (\*y_i, \*x_i))=\log p(\*y_i|\fomegaxi)$$
where $p(\*y_i|\fomegaxi)$ is the model's predicted PDF evaluated at $\*y_i$, given the input $x_i$. A more detailed description is given in the Appendix Section 1.5. 
The metric is unbounded and maximized by a perfect prediction (mode at $\*y_i$) with no variance. As the predictive mode moves away from $\*y_i$, increasing the variance tends to increase PLL (by maximizing probability mass at $\*y_i$). While PLL is an elegant measure, it has been criticized for allowing outliers to have an overly negative effect on the score \citep{Selten1998}.

\paragraph{Continuous Ranked Probability Score (CRPS)}  Continuous Ranked Probability Score is a measure that takes the full predicted PDF into account with less sensitivity to outliers. A prediction with low variance that is slightly offset from the true observation will receive a higher score form CRPS than PLL. In order for CRPS to be analytically tractable, we need to assume a Gaussian unimodal predictive distribution. CRPS is defined as
$$\text{CRPS}(\fomegaxxi, (y_i, x_i))=\int_{-\infty}^{\infty}\big(F(y)-\mathbf{1}(y\geq y_i)\big)^2\text{d}y$$
where $F(y)$ is the predictive CDF, and $\mathbf{1}(y\geq y_i)=1$ if $y\geq y_i$ and $0$ otherwise (for univariate distributions) \citep{Gneiting2007}. CRPS is interpreted as the sum of the squared area between the CDF and 0 where $y<y_i$ and between the CDF and 1 where $y\geq y_i$. A perfect prediction with no variance yields a CRPS of 0; for all other cases the value is larger. CRPS has no upper bound.

\subsection{Benchmark models and normalized metrics} \label{sec:benchmark}

It is difficult to interpret the quality of uncertainty from raw PLL and CRPS values. We propose to normalize the metrics between useful lower and upper bounds. The normalized measures estimate the performance of an uncertainty model between the trivial solution (constant uncertainty) and \textit{optimal uncertainty} for each prediction. For the lower bound, we define a baseline that predicts constant variance regardless of input. The variance is set to a fixed value that optimizes CRPS on validation data. We call this model Constant Uncertainty BN (CUBN). It reflects our best guess of constant variance on test data -- thus, any improvement in uncertainty quality over CUBN indicates a sensible estimate of uncertainty. We similarly define a baseline for dropout, Constant Uncertainty Dropout (CUDO). The  modeling of variance (uncertainty) by MCBN and CUBN are visualized in Figure \ref{fig:toydata}.

An upper bound on uncertainty performance can also be defined for a probabilistic model $f$ with respect to CRPS or PLL. For each observation $(y_i, x_i)$, a value for the predictive variance $T_i$ can be chosen that maximizes PLL or minimizes CRPS\footnote{$T_i$ can be found analytically for PLL, but must be found numerically for CRPS.}. Using CUBN as a lower bound and the optimized CRPS score as the upper bound, uncertainty estimates can be normalized between these bounds (1 indicating optimal performance, and 0 indicating same performance as fixed uncertainty). We call this normalized measure
$
\overline{\textrm{CRPS}} = \frac{\text{CRPS}(f, (y_i, x_i)) - \text{CRPS}(f_{CU}, (y_i, x_i))}{\min_{T} \text{CRPS}(f, (y_i, x_i)) - \text{CRPS}(f_{CU}, (y_i, x_i))} \times 100,$
and the PLL analogue
$
\overline{\textrm{PLL}} = \frac{\text{PLL}(f, (y_i, x_i)) - \text{PLL}(f_{CU}, (y_i, x_i))}{\max_{T} \text{PLL}(f, (y_i, x_i)) - \text{PLL}(f_{CU}, (y_i, x_i))} \times 100.$


\begin{table*}[t]
\small
\caption{\textbf{Uncertainty quality measured on eight regression datasets.} MCBN, MCDO and MNF are compared over 5 random 80-20 splits of the data with 5 different random seeds each split. We report $\overline{\textrm{CRPS}}$ and $\overline{\textrm{PLL}}$, uncertainty metrics CRPS and PLL normalized to a lower bound of constant variance and upper bound that maximizes the metric expressed as a percentage (described in Section \ref{sec:benchmark}). Higher numbers mean the model is closer to the upper bound. We check if the reported values for $\overline{\textrm{CRPS}}$ and $\overline{\textrm{PLL}}$ significantly exceed the lower bound using a one sample t-test (significance level indicated by *'s). See text for further details.}
\centering
\vspace{2mm}
\begin{tabular}{l@{\hskip 6mm}r@{\hskip 1mm}cr@{\hskip 1mm}cr@{\hskip 1mm}c@{\hskip 12mm}r@{\hskip 1mm}cr@{\hskip 1mm}cr@{\hskip 1mm}c}
\toprule
                           & \multicolumn{6}{c}{$\overline{\textrm{CRPS}}$}                                & \multicolumn{6}{c}{$\overline{\textrm{PLL}}$}                                 \\
Dataset                    & \multicolumn{2}{c}{MCBN} & \multicolumn{2}{c}{MCDO} & \multicolumn{2}{c}{MNF} & \multicolumn{2}{c}{MCBN} & \multicolumn{2}{c}{MCDO} & \multicolumn{2}{c}{MNF} \\
\midrule
Boston             & {8.50}          & ****      & 3.06        & ****       & 5.88         & ****      & {10.49}        & ****      & 5.51         & ****      & 1.76        & ns       \\
Concrete                   & 3.91         & ****      & 0.93        & *          & {3.13}        & ***      & -36.36       & **        & {10.92}        & ****      & -2.16        & ns      \\
Energy          & {5.75}         & ****      & 1.37        & ns         & 1.10             & ns           & {10.89}        & ****      & -14.28       & *         & -33.88            & ns           \\
Kin8nm             & {2.85}         & ****      & 1.82        & ****       & 0.53            & ns          & {1.68}         & ***       & -0.26        & ns        & 0.42            & ns           \\
Power                & {0.24}         & ***       & -0.44       & ****       & -0.89            & ****           & 0.33         & **        & {3.52}         & ****      & -0.87            & ****           \\
Protein & {2.66}         & ****      & 0.99        & ****       & 0.57         & ****          & 2.56         & ****      & {6.23}         & ****      & 0.52            & ****         \\
Wine (Red)         & 0.26         & **        & {2.00}           & ****       & 0.94            & ****          & 0.19         & *         & {2.91}         & ****      & 0.83            & ****          \\
Yacht        & -56.39       & ***       & {21.42}       & ****       & 24.92      & ****      & 45.58        & ****      & -41.54       & ns        & 46.19       & ****      \\ 
\bottomrule
\end{tabular}
\label{table:main_results}
\vspace{-1mm}
\end{table*}

\subsection{Test setup}

Our evaluation compares MCBN to MCDO \cite{Gal2015a} and MNF \cite{pmlr-v70-louizos17a} using the datasets and metrics described above. Our setup is similar to  \cite{hernandez2015probabilistic}, which was also followed by \cite{Gal2015a}. However, our comparison implements a different hyperparameter selection, allows for a larger range of dropout rates, and uses larger networks with two hidden layers.


For the regression task, all models share a similar architecture: two hidden layers with 50 units each, and ReLU activations, with the exception of Protein Tertiary Structure dataset (100 units per hidden layer). Inputs and outputs were normalized during training. Results were averaged over five random splits of $20\%$ test and $80\%$ training and cross-validation (CV) data. For each split, 5-fold CV by grid search with a RMSE minimization objective was used to find training hyperparameters and optimal n.o. epochs, out of a maximum of $2000$. For BN-based models, the hyperparameter grid consisted of a weight decay factor ranging from $0.1$ to $1^{-15}$ by a $\log10$ scale, and a batch size range from $32$ to $1024$ by a $\log2$ scale. For DO-based models, the hyperparameter grid consisted of the same weight decay range, and dropout probabilities in $\{0.2, 0.1, 0.05, 0.01, 0.005, 0.001\}$. DO-based models used a batch size of 32 in all evaluations. For MNF\footnote{Where we used an adapted version of the authors' code.}, the n.o. epochs was optimized, the batch size was set to 100, and early stopping test performed each epoch (compared to every 20th for MCBN, MCDO).


For MCBN and MCDO, the model with optimal training hyperparameters was used to optimize $\tau$ numerically. This optimization was made in terms of average CV CRPS for MCBN, CUBN, MCDO, and CUDO respectively.

Estimates for the predictive distribution were obtained by taking $T=500$ stochastic forward passes through the network. For each split, test set evaluation was done 5 times with different seeds. Implementation was done in TensorFlow with the Adam optimizer and a learning rate of 0.001. 

For the image classification test we use CIFAR10 \cite{krizhevsky2009learning} which includes 10 object classes with 5,000 and 1,000 images in the training and test sets, respectively. Images are 32x32 RGB format. We trained a ResNet32 architecture with a batch size of 32, learning rate of 0.1, weight decay of 0.0002, leaky ReLU slope of 0.1, and 5 residual units. SGD with momentum was used as the optimizer.  

Code for reproducing our experiments is available  at \href{https://github.com/icml-mcbn/mcbn}{https://github.com/icml-mcbn/mcbn}.


\subsection{Test results}

The regression experiment comparing uncertainty quality is summarized in Table \ref{table:main_results}. We report  $\overline{\textrm{CRPS}}$ and $\overline{\textrm{PLL}}$, expressed as a percentage, which reflects how close the model is to the upper bound, and check to see if the model significantly exceeds the lower bound using a one sample t-test (significance level is indicated by *'s). Further details are provided in Appendix Section 1.7. 

In Figure \ref{fig:plots} \textit{(left)}, we present a novel visualization of uncertainty quality for regression problems. Data are sorted by estimated uncertainty in the $x$-axis. Grey dots show the errors in model predictions, and the shaded areas show the model uncertainty. A running mean of the errors appears as a gray line. If uncertainty estimation is working well, a correlation should exist between the mean error (gray line) and uncertainty (shaded area). This indicates that the uncertainty estimation recognizes samples with larger (or smaller) potential for predictive errors.

\begin{figure*}[t]
\vspace{-1mm}

\begin{tabular}{@{}c@{\hskip 1mm}c@{\hskip 1mm}c@{\hskip 1mm}c@{}}
\includegraphics[height=34mm]{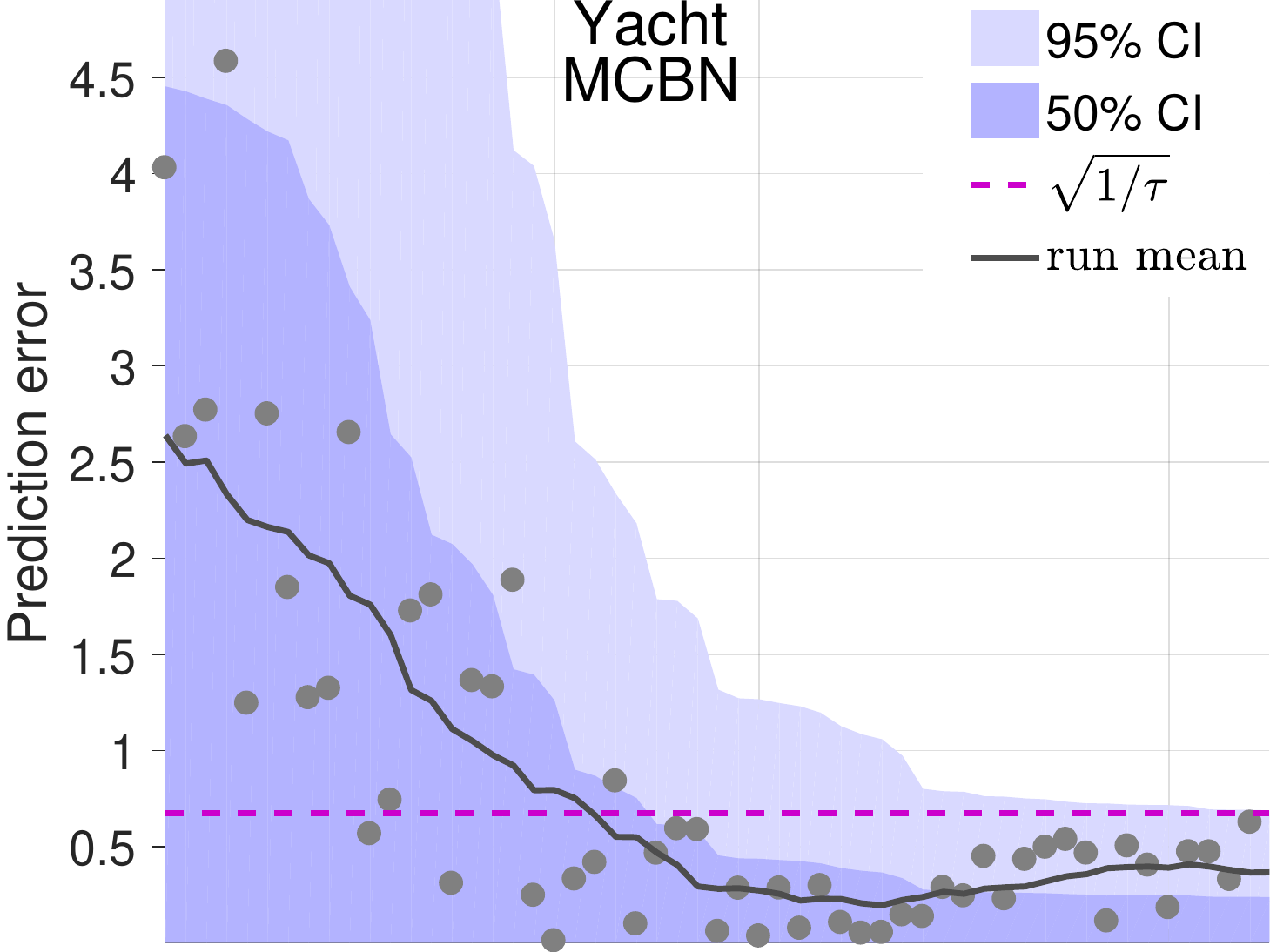} &
\includegraphics[height=34mm]{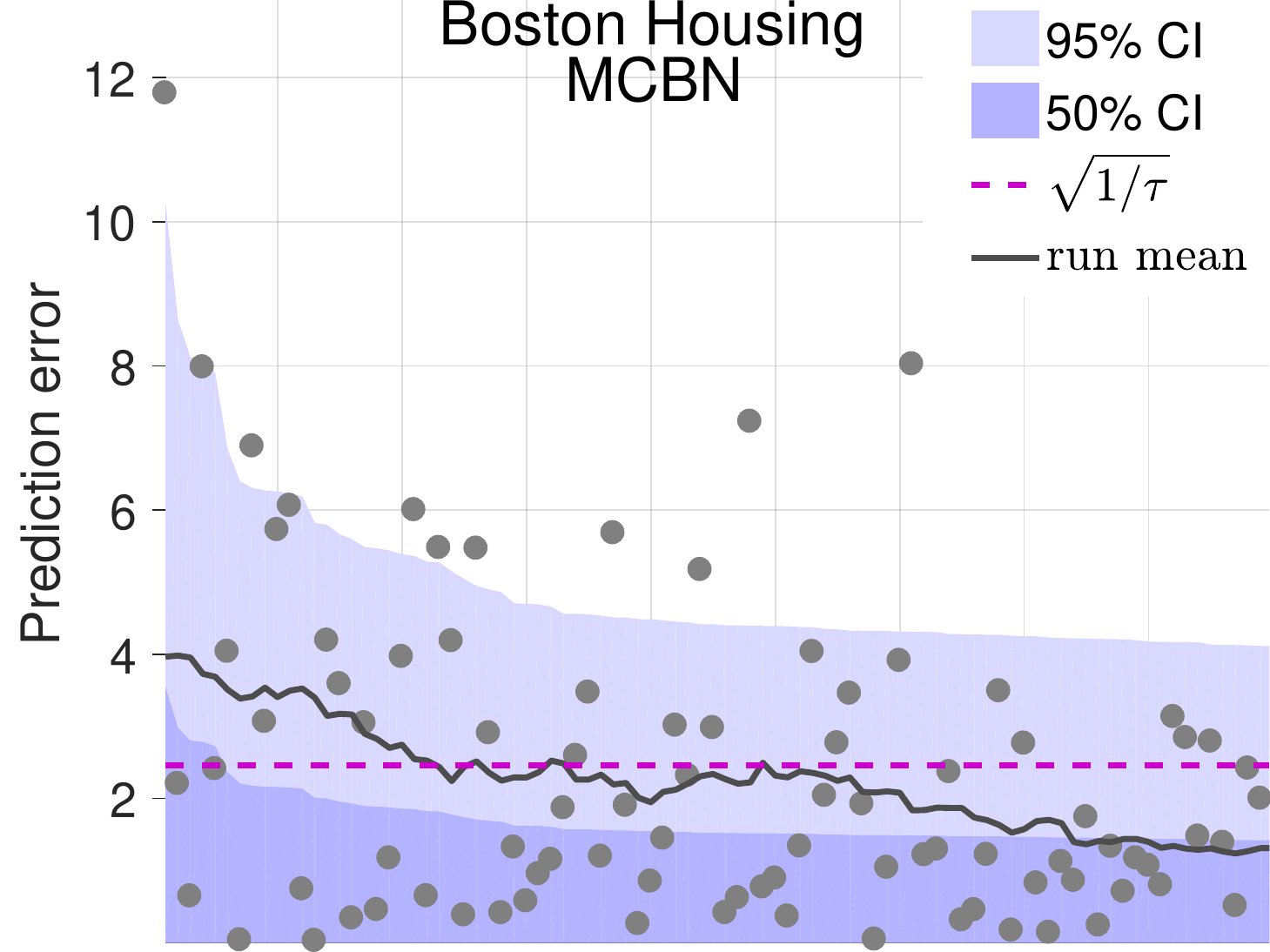} &
\includegraphics[height=34mm]{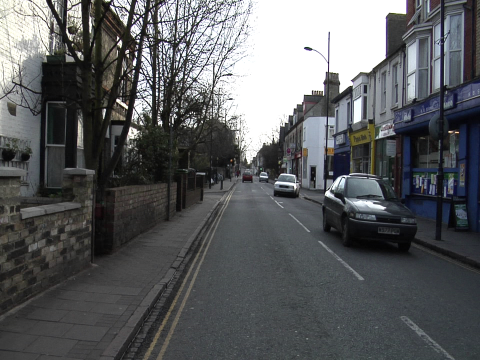} &
\includegraphics[height=34mm]{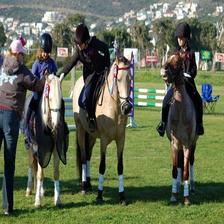}
\\

\includegraphics[height=34mm]{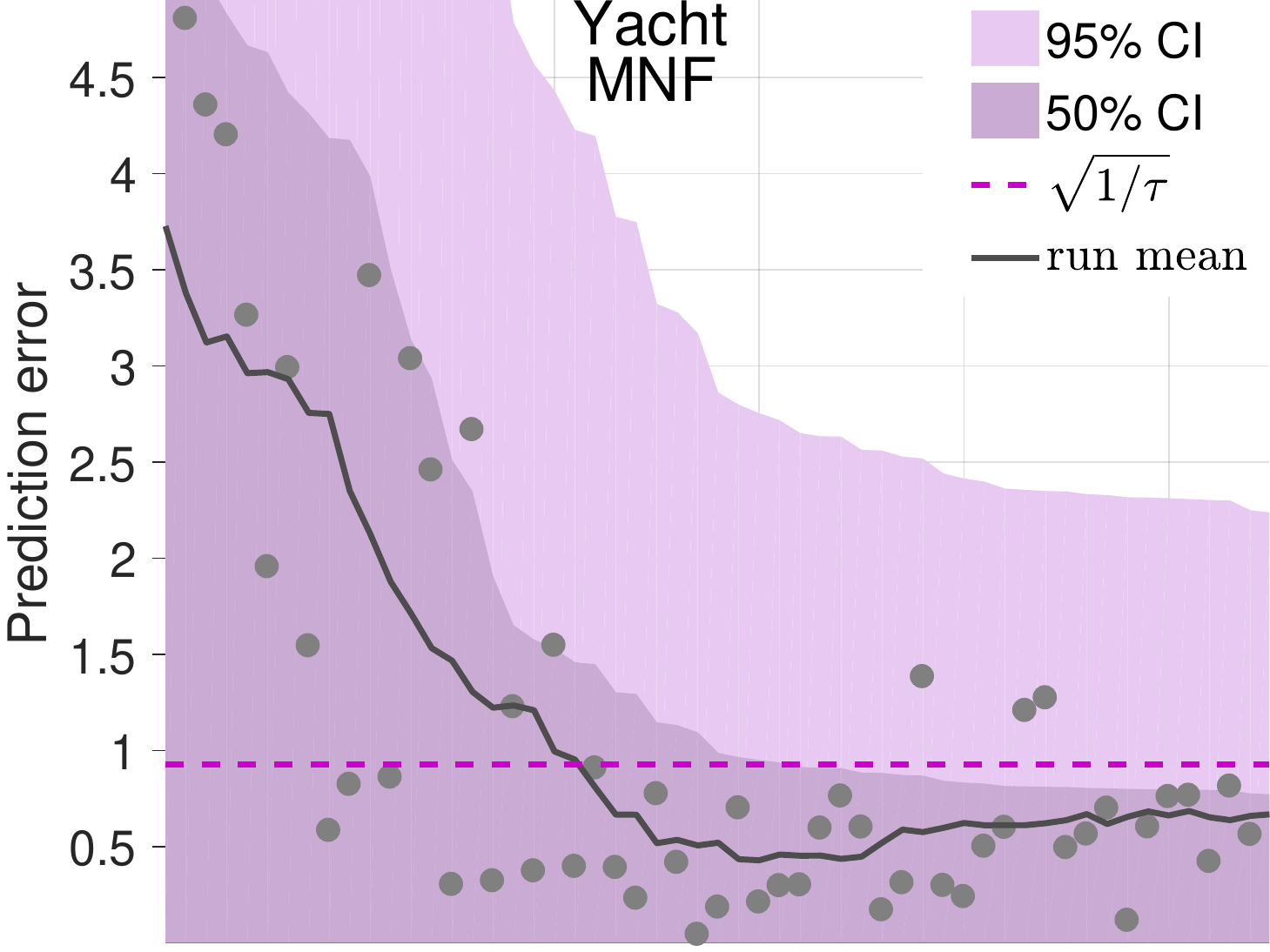} &
\includegraphics[height=34mm]{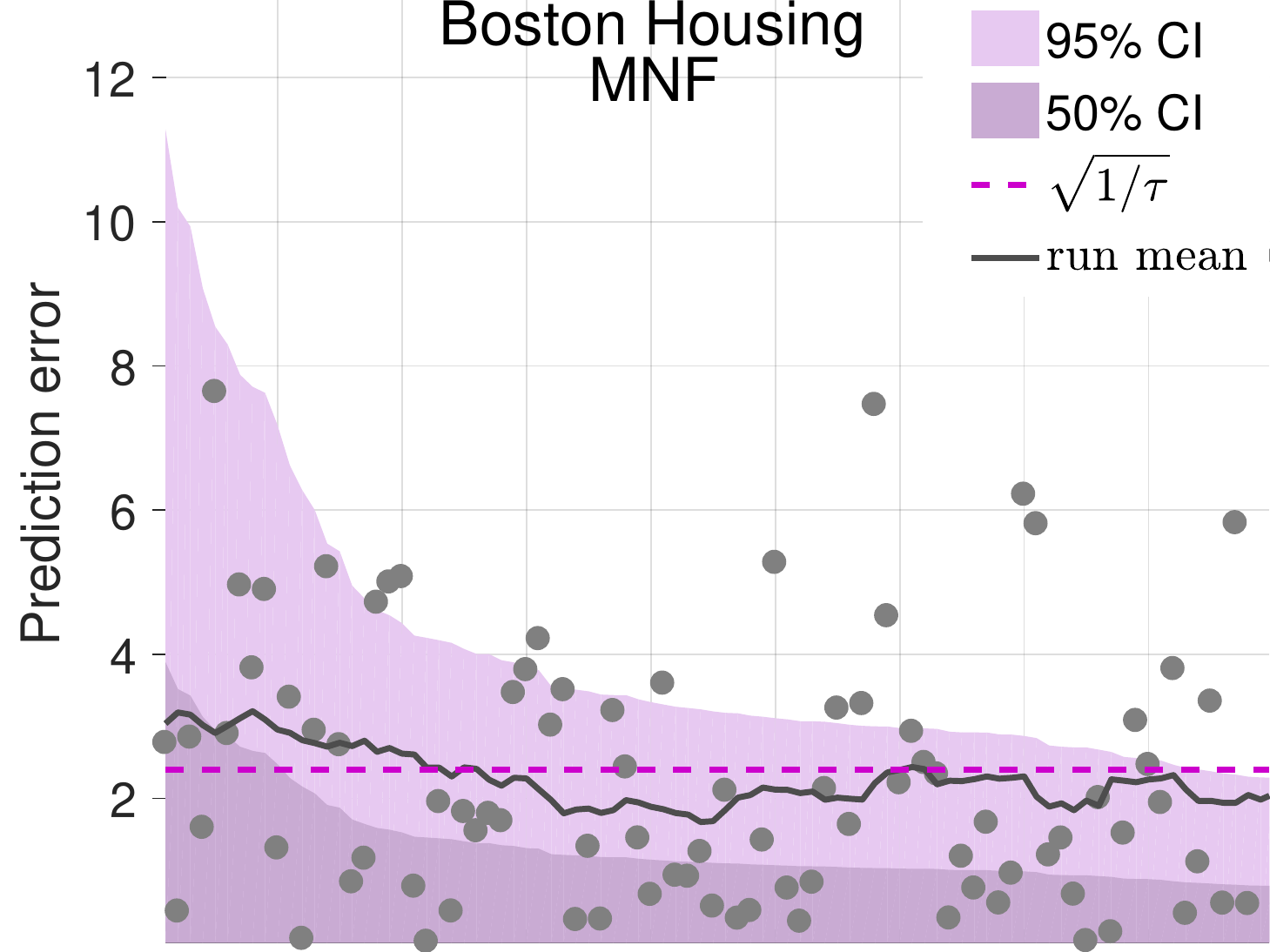} &
\includegraphics[height=34mm]{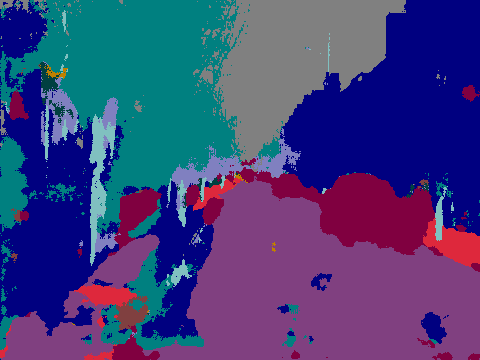}  &
\includegraphics[height=34mm]{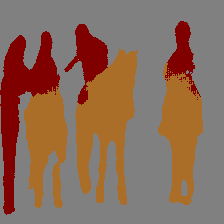}
\\
\includegraphics[height=34mm]{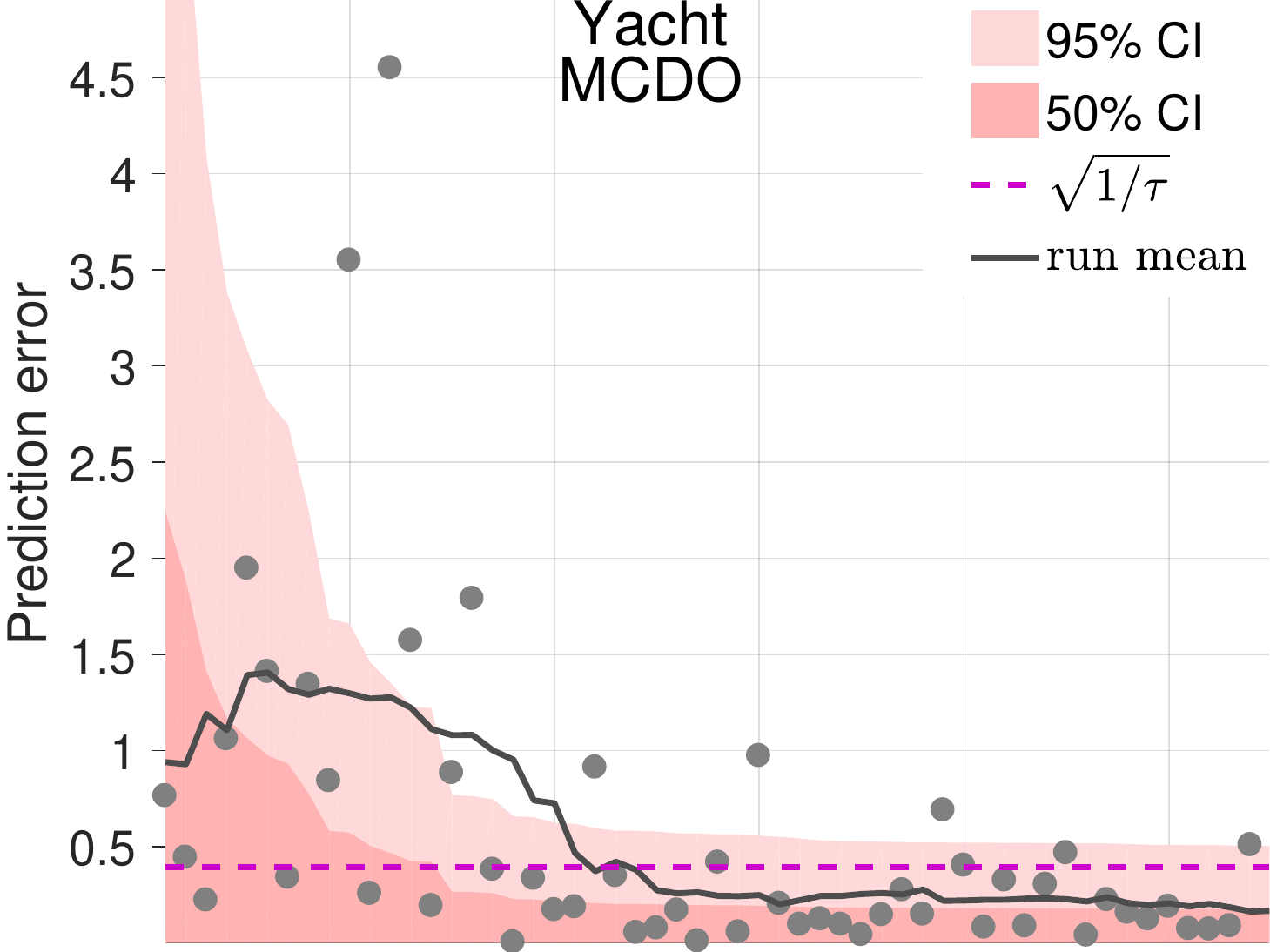} &
\includegraphics[height=34mm]{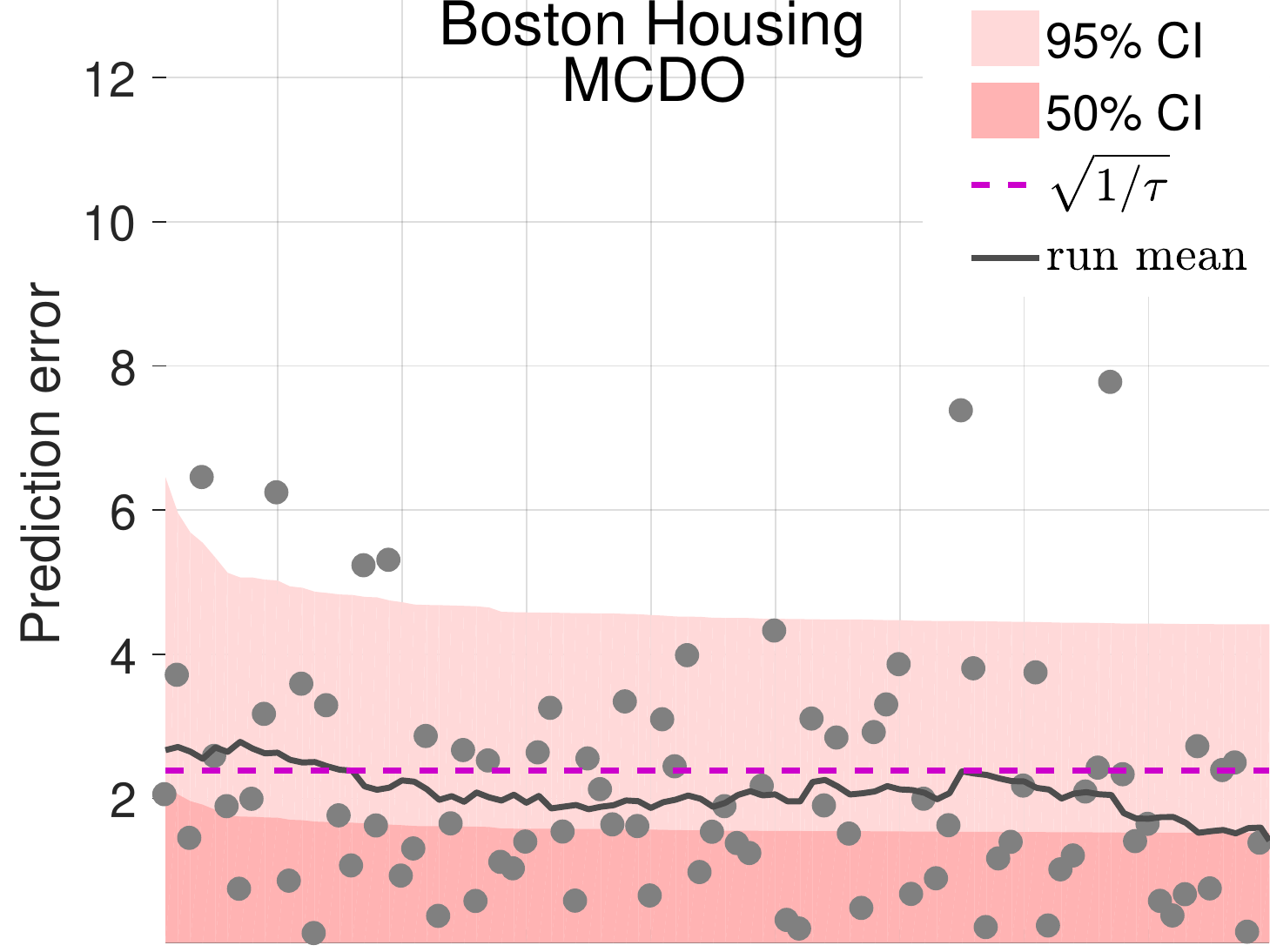} &
\includegraphics[height=34mm]{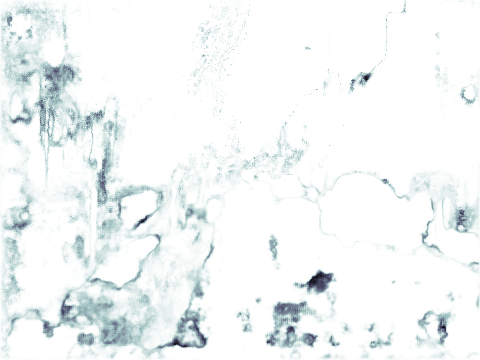} &
\includegraphics[height=34mm]{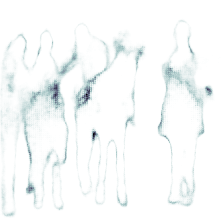}
\\
\hspace{5mm} \scriptsize{Data sorted by estimated uncertainty} &
\hspace{5mm} \scriptsize{Data sorted by estimated uncertainty} &
\multicolumn{2}{c}{\scriptsize{Image segmentation uncertainty (CamVid and PASCAL-VOC)}} \\
\end{tabular}
\caption{ \textbf{Uncertainty-error plots (left) and segmentation and uncertainty results applying MCBN to Bayesian SegNet (right).} \textit{(left)} Errors in predictions (gray dots) sorted by estimated uncertainty on select datasets. The shaded areas show model uncertainty for MCBN (blue), MNF (violet) and MCDO (red). The light area indicates 95\% CI, dark area 50\% CI. Gray dots show absolute prediction errors on the test set, and the gray line depicts a running mean of the errors. The dashed line indicates the optimized constant uncertainty. A correlation between estimated uncertainty (shaded area) and mean error (gray) indicates the uncertainty estimates are meaningful for estimating errors.  \textit{(right)} Applying MCBN to Bayesian SegNet \citep{KendallBC15} on scenes from CamVid ($3^{rd}$ column) and PASCAL-VOC ($4^{th}$ column). Top: original image. Middle: the Bayesian estimated segmentation. Bottom: estimated uncertainty using MCBN for all classes. The uncertainty maps for both datasets are reasonable, but qualitatively better for PASCAL-VOC due to the larger mini-batch size (36) compared to CamVid (10). Smaller batch sizes were used for CamVid due to memory limitations (CamVid images are 360x480 while VOC are 224x224). See Appendix for complete results.} \label{fig:plots}
\end{figure*}

We applied MCBN on the image classification task of CIFAR10. The baseline in this case is the softmax distribution using the moving average for BN units. Log likelihood (PLL) is the metric used to compare with the baseline. The baseline achieves a PLL of \textbf{-0.32} on the test set, while MCBN obtains a PLL of \textbf{-0.28}. Table \ref{table:cifar_results} shows the performance of MCBN when using different number of stochastic forward passes (the MCBN batchsize is fixed to the training batch size at 32). PLL improves as the number of the stochastic passes increases, until it is significantly better than the softmax baseline. 

To demonstrate how model uncertainty can be obtained from an existing network with minimal effort, we applied MCBN to an image segmentation task using Bayesian SegNet with the main CamVid and PASCAL-VOC models in \cite{KendallBC15}. We simply ran multiple forward passes with different mini-batches randomly taken from the train set. The models obtained from the online model zoo have BN blocks after each layer. We recalculate mean and variance for the first 2 blocks only and use the training statistics for the rest of the blocks. Mini-batches of size 10 and 36 were used for CamVid and VOC respectively due to memory limits. The results in Figure \ref{fig:plots}  \textit{(right)} were obtained from 20 stochastic forward passes, showing high uncertainty near object boundaries. The VOC results are more appealing because of larger mini-batches. 

\begin{table}[b!]
\vspace{-8mm}
\centering
\small
\caption{\textbf{Uncertainty quality for image classification varying number of stochastic forward passes.} Uncertainty quality for image classification measured by PLL. ResNet32 is trained on CIFAR10 with batch size 32. PLL improves as the sampling increases until it is significantly better than the softmax baseline (\textbf{-0.32}).}
\vspace{1mm}
\begin{tabular}{lcccccccc}
\toprule
& \multicolumn{8}{c}{Number of stochastic forward passes}\\
&1&2&4&8&16&32&64&128\\ 
\midrule
PLL&-.36&-.32&-.30&-.29&-.29&-.28&-.28&\textbf{-.28}\\
\bottomrule
\end{tabular}
\label{table:cifar_results}
\end{table}


We provide additional experimental results in the Appendix. 
Appendix Tables 2 and 3 show the mean $\overline{\textrm{CRPS}}$ and $\overline{\textrm{PLL}}$ values from the regression experiment. Table 4 provides the raw CRPS and PLL scores. In Table 5 we provide RMSE results of the MCBN and MCDO networks in comparison with non-stochastic BN and DO networks. These results indicate that the procedure of multiple forward passes in MCBN and MCDO show slight improvements in the predictive accuracy compared to their non-Bayesian counterparts. In Tables 6 and 7, we investigate the effect of varying batch size while keeping other hyperparameters fixed. We see that performance deteriorates with small batch sizes ($\leq$16), a known issue of BN \citep{DBLP:journals/corr/Ioffe17}. Similarly, results varying the number of stochastic forward passes $T$ is reported in Tables 8 and 9. While performance benefits from large $T$, in some cases $T=50$ (i.e. 1/10 of $T$ in the main evaluation) performs well. Uncertainty-error plots for all the datasets are provided in the Appendix.

\section{Discussion}


The results presented in Tables \ref{table:main_results}-\ref{table:cifar_results} and Appendix Tables 2-9 
indicate that MCBN generates meaningful uncertainty estimates that correlate with actual errors in the model's prediction. In Table \ref{table:main_results}, we show statistically significant improvements over CUBN in the majority of the datasets, both in terms of $\overline{\textrm{CRPS}}$ and $\overline{\textrm{PLL}}$. The visualizations in Figure \ref{fig:plots} and in the Appendix Figures 2-3 show correlations between the estimated model uncertainty and errors of the network's predictions. We perform the same experiments using MCDO and MNF, and find that MCBN generally performs on par with both methods. Looking closer, MCBN outperforms MCDO and MNF in more cases than not, measured by $\overline{\textrm{CRPS}}$. However, care must be used. The learned parameters are different, leading to different predictive means and confounding direct comparison. 

The results on the Yacht Hydrodynamics dataset seem contradictory. The $\overline{\textrm{CRPS}}$ score for MCBN are extremely negative, while the $\overline{\textrm{PLL}}$ score is extremely positive. The opposite trend is observed for MCDO. To add to the puzzle, the visualization in Figure \ref{fig:plots} depicts an extremely promising uncertainty estimation that models the predictive errors with high fidelity. We hypothesize that this strange behavior is due to the small size of the data set, which only contains 60 test samples, or due to the Gaussian assumption of CRPS. There is also a large variability in the model's accuracy on this dataset, which further confounds the measurements for such limited data. 

One might criticize the overall quality of uncertainty estimates observed in all the models we tested, due to the magnitude of $\overline{\textrm{CRPS}}$ and $\overline{\textrm{PLL}}$ in Table \ref{table:main_results}. The scores rarely exceed 10\% improvement over the lower bound. However, we caution that these measures should be taken in context. The upper bound is very difficult to achieve in practice -- it  is optimized for \textit{each test sample individually} -- and the lower bound is a quite reasonable estimate.

The study of MCBN sensitivity to batch size revealed that a certain batch size is required for the best performance, dependent on the data. When doing inference on a GPU, large batch sizes may cause memory issues for cases where the input is large and the network has a large number of parameters, as is common for state-of-the-art image classification networks. However, there are various workarounds to this problem. One can store BN statistics, instead of batches, to reduce memory issues. Furthermore, we can use the Gaussian estimate of the BN statistics as discussed previously, which makes memory and computation extremely efficient.



\section{Conclusion}
In this work, we have shown that training a deep network using batch normalization is equivalent to approximate inference in Bayesian models. 
We show evidence that the uncertainty estimates from MCBN correlate with actual errors in the model's prediction, and are useful for practical tasks such as regression, image classification, and image segmentation. Our experiments show that MCBN yields a significant improvement over the optimized constant uncertainty baseline, on par with MCDO and MNF. Our evaluation also suggests new normalized metrics based on useful upper and lower bounds, and a new visualization which provides an intuitive explanation of uncertainty quality. 

Finally, it should be noted that over the past few years, batch normalization has become an integral part of most -- if not all -- cutting edge deep networks. We have shown that it is possible to obtain meaningful uncertainty estimates from existing models without modifying the network or the training procedure. With a few lines of code, robust uncertainty estimates can be obtained by computing the variance of multiple stochastic forward passes through an existing network.


\bibliography{mcbn_icml}
\bibliographystyle{icml2018}

\clearpage
\setcounter{section}{0}
\setcounter{equation}{2}
\onecolumn

\section{Appendix}

\subsection{Variational Approximation}\label{appendix:VA}
Assume we were to come up with a family of distributions parameterized by $\bm{\theta}$ in order to approximate the posterior, $\qthetaomega$. Our goal is to set $\bm{\theta}$ such that $\qthetaomega$ is as similar to the true posterior $\posterior$ as possible.

For clarity, $\qthetaomega$ is a distribution over stochastic parameters $\bm{\omega}$ that is determined by a set of learnable parameters $\bm{\theta}$ and some source of randomness. The approximation is therefore limited by our choice of parametric function $\qthetaomega$ as well as the randomness.\footnote{In an approx. Bayesian view of a NN, $\qthetaomega$ would correspond to the distribution of weights in the network given by some SRT.} Given $\bm{\omega}$ and an input $\sample$, an output distribution $p(\lab|\sample,\param)=p(\lab|f_{\param}(\sample))=\fomegaxy$ is induced by observation noise (the conditionality of which is omitted for brevity).

One strategy for optimizing $\bm{\theta}$ is to minimize $\KLomega$, the KL divergence of $\posterior$ \wrt $\qthetaomega$. Minimizing $\KLomega$ is equivalent to maximizing the ELBO:
$$\int_{\bm{\omega}}\qthetaomega\ln\likelihood\domega-\KLomegap$$
Assuming \iid observation noise, this is equivalent to minimizing:
$$\LVItheta:=-\sum_{n=1}^{N}\expLogLikelihood+\KLomegap$$ 
Instead of making the optimization on the full training set, we can use a subsampling (yielding an unbiased estimate of $\LVItheta$) for iterative optimization (as in mini-batch optimization):
$$\hat{\mathcal{L}}_{\text{VA}}(\bm{\theta}):=\subsample\expLogLikelihoodi+\KLomegap$$
During optimization, we want to take the derivative of the expected log likelihood \wrt the learnable parameters $\bm{\theta}$. \cite{Gal2016Uncertainty} provides an intuitive interpretation of a MC estimate  for NNs trained with a SRT (equivalent to the reparametrisation trick in \cite{Kingma2014}), and this interpretation is followed here. We let an auxillary variable $\bm{\epsilon}$ represent the stochasticity during training such that $\param=g(\bm{\theta},\bm{\epsilon})$. The function $g$ and the distribution of $\bm{\epsilon}$ are such that $p(g(\bm{\theta},\bm{\epsilon})) = \qthetaomega$.\footnote{In a NN trained with BN, it is easy to see that $g$ exists if we let $\bm{\epsilon}$ represent a mini-batch selection from the training data, since all BN units' means and variances are completely determined by $\bm{\epsilon}$ and $\theta$.} Assuming $\qthetaomega$ can be written $\int_{\bm{\epsilon}} q_{\bm{\theta}}(\bm{\omega}|\bm{\epsilon})p(\bm{\epsilon})\text{d}\bm{\epsilon}$ where $q_{\bm{\theta}}(\bm{\omega}|\bm{\epsilon}) = \delta(\bm{\omega}-g(\bm{\theta},\bm{\epsilon}))$, this auxiliary variable yields the estimate (full proof in \cite{Gal2016Uncertainty}):
$$\hat{\mathcal{L}}_{\text{VA}}(\bm{\theta})=\expLogLikelihoodKW+\KLomegap$$
where taking a single sample MC estimate of the integral yields the optimization objective in Eq. 1.

\subsection{KL Divergence of factorized Gaussians}\label{appendix:factorized_gaussians}
If $\qthetaomega$ and $\prior$ factorize over all stochastic parameters:
\begin{align}\label{eq:kldivergencefactorization}
\begin{split}
\KLomegap=&-\int_{\bm{\omega}}\prod_{i}\big[\qthetaomegai\big]\ln\frac{\prod_{i}p(\omega_{i})}{\prod_{i}\qthetaomegai}\text{d}\bm{\omega} \\
=&-\int_{\bm{\omega}}\prod_{i}\big[\qthetaomegai\big]\sum_{i}\bigg[\ln\frac{p(\omega_{i})}{\qthetaomegai}\bigg]\prod_{i}\text{d}\omega_{i} \\
=&\sum_{j}\Big[-\int_{\bm{\omega}}\prod_{i}\big[\qthetaomegai\big]\ln\frac{p(\omega_{j})}{\qthetaomegaj}\prod_{i}\text{d}\omega_{i}\Big]
\\
=&\sum_{j}\Big[-\int_{\omega_{j}}\qthetaomegaj\ln\frac{p(\omega_{j})}{\qthetaomegaj}\text{d}\omega_{j}\int_{\omega_{i\neq j}}\prod_{i\neq j}\qthetaomegai\text{d}\omega_{i}\Big] \\
=&\sum_{i}-\int_{\omega_{i}}\qthetaomegai\ln\frac{p(\omega_{i})}{\qthetaomegai}\text{d}\omega_{i} \\
=&\sum_{i}\KLomegapomegai
\end{split} 
\end{align}
such that $\KLomegap$ is the sum of the KL divergence terms for the individual stochastic parameters $\omega_i$. If the factorized distributions are Gaussians, where $\qthetaomegai=\mathcal{N}(\mu_q,\sigma_q^2)$ and $p(\omega_i)=\mathcal{N}(\mu_p, \sigma_p^2)$ we get:
\begin{align}\label{eq:kldivergencegaussian}
\begin{split}
\KLomegapomegai =&\int_{\omega_{i}}\qthetaomegai\ln\frac{\qthetaomegai}{\prioromegai}\text{d}\omega_{i} \\
=&-H(\qthetaomegai)-\int_{\omega_{i}}\qthetaomegai\ln\prioromegai\text{d}\omega_{i} \\
=&-\frac{1}{2}(1+\ln(2\pi\sigma_q^2)) \\
&-\int_{\omega_{i}}\qthetaomegai\ln\frac{1}{(2\pi\sigma_p^2)^{1/2}}\exp\Big\{-\frac{(\omega_i-\mu_p)^2}{2\sigma_p^2}\Big\}\text{d}\omega_{i}\\
=&-\frac{1}{2}(1+\ln(2\pi\sigma_q^2))\\
&+\frac{1}{2}\ln(2\pi\sigma_p^2)+\frac{\expec_q[\omega_i^2]-2\mu_p\expec_q[\omega_i]+\mu_p^2}{2\sigma_p^2} \\
=&\ln\frac{\sigma_p}{\sigma_q}+\frac{\sigma_q^2+(\mu_q-\mu_p)^2}{2\sigma_p^2}-\frac{1}{2}
\end{split}
\end{align}
for each KL divergence term. Here $H(\qthetaomegai)=\frac{1}{2}(1+\ln(2\pi\sigma_q^2))$ is the differential entropy of $\qthetaomegai$.

\subsection{Distribution of \texorpdfstring{$\means_{\batch}^{u}, \vars_{\batch}^{u}$}{}}\label{appendix:dist_mu_sigma}
\begin{figure}[t]
	\centering
	\begin{tabular}{@{}c@{\hskip 1mm}c@{}}
		\includegraphics[width=0.45\linewidth]{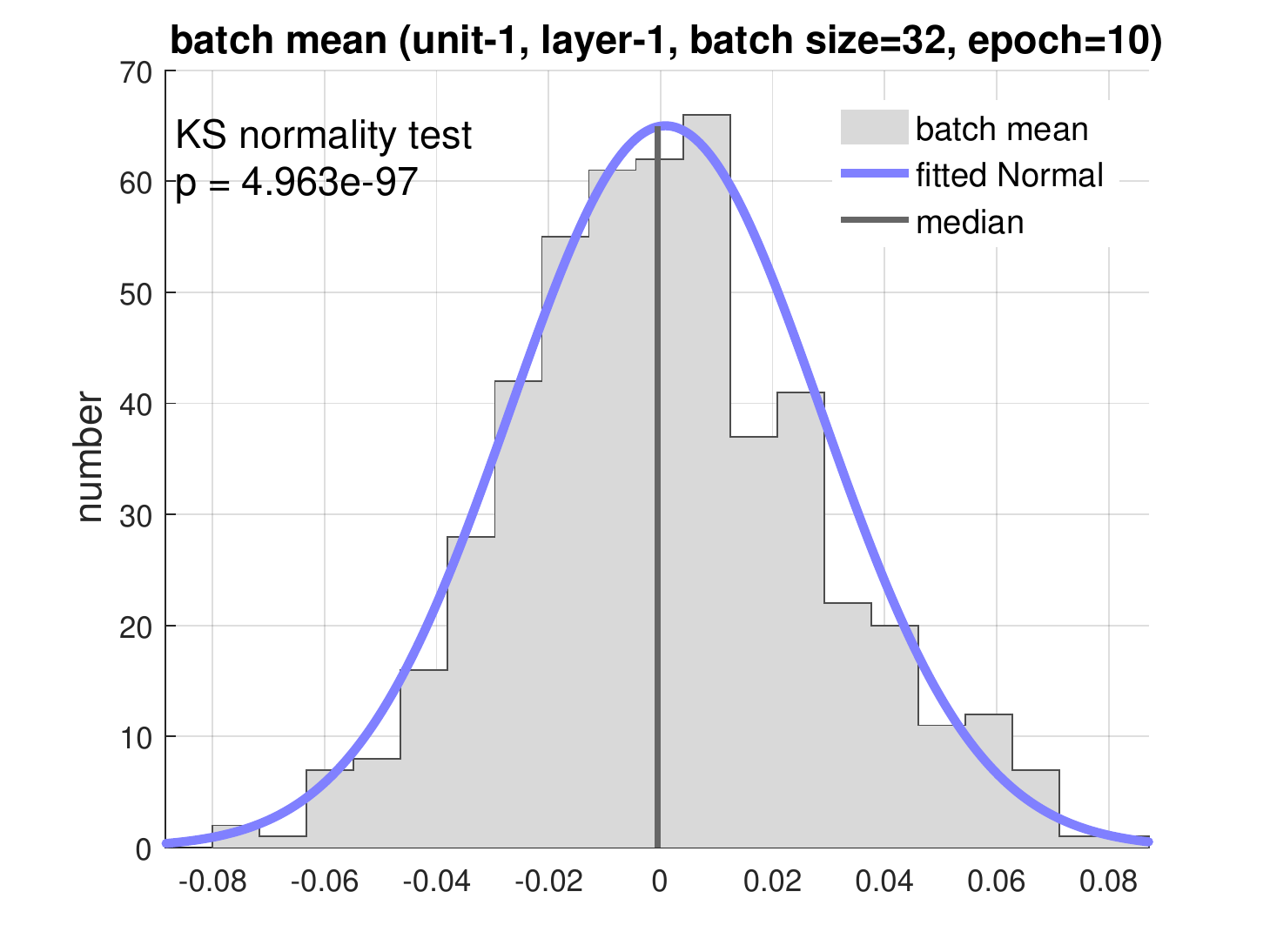} &
		\includegraphics[width=0.45\linewidth]{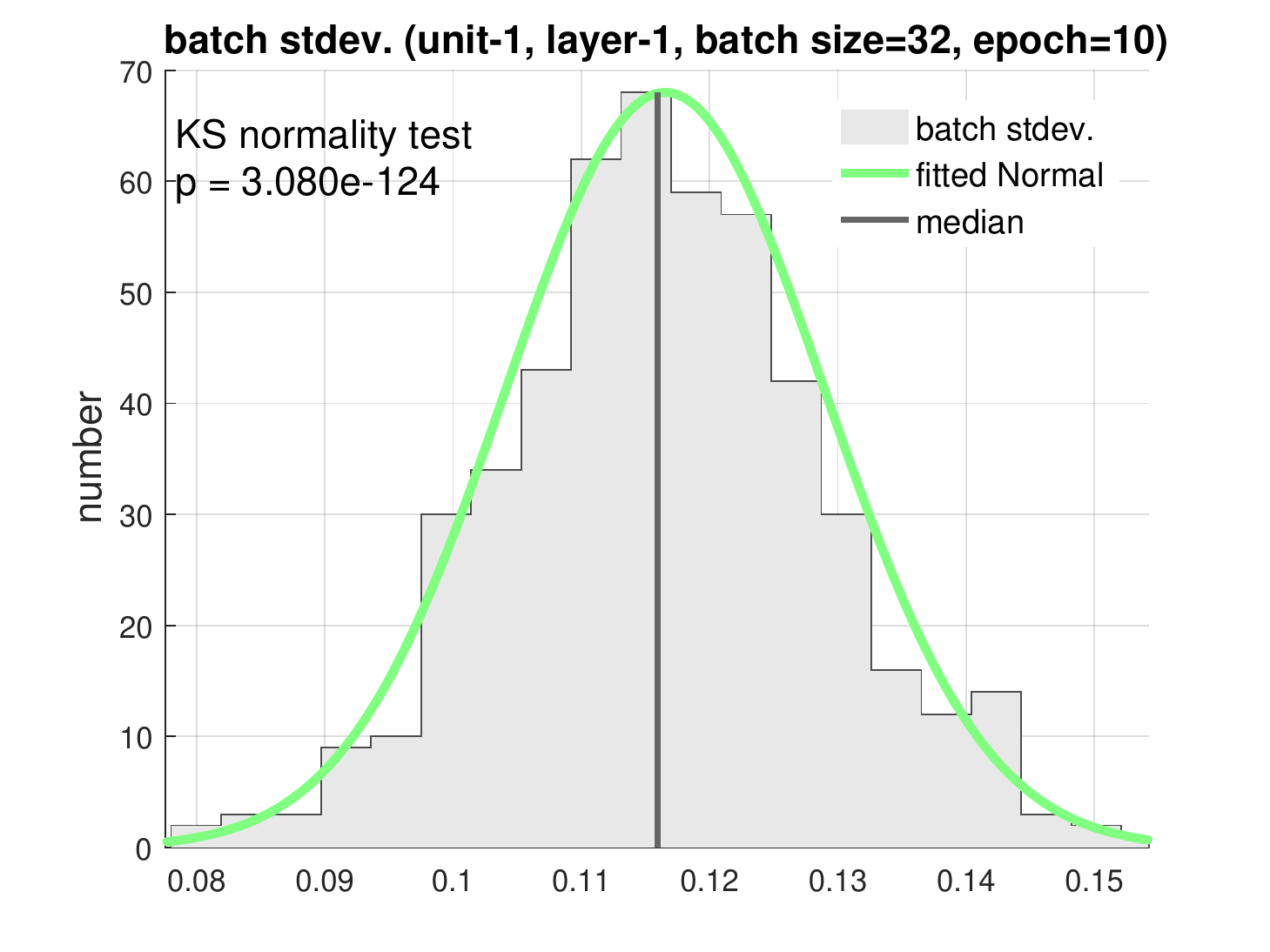} \\
	\end{tabular}
	\vspace{-3mm}
	\caption{Batch statistics used to train the network are normal. A one-sample Kolmogorov-Smirnov test checks that $\means_{\batch}$ and $\vars_{\batch}$ come from a standard normal distribution. More examples are available in Appendix \ref{appendix:extendedstatistics}.}
	\label{fig:batchstatistics}
	\vspace{-3mm}
\end{figure}
Here we approximate the distribution of mean and standard deviation of a mini-batch, separately to two Gaussians -- This has also been empirically verified, see Figure \ref{fig:batchstatistics} for 2 sample plots and the appendix section \ref{appendix:extendedstatistics} for more. For the mean we get:
$$\mu_\text{B}=\frac{\Sigma_{\text{m=1}}^{\text{M}}\weightedsum}{\text{M}}$$
where $\*x_{\text{m}}$ are the examples in the sampled batch. We will assume these are sampled \iid \footnotemark. Samples of the random variable $\weightedsum$ are then \iid. Then by central limit theorem (CLT) the following holds for sufficiently large M (often $\geq30$):
\footnotetext{Although in practice with deep learning, mini-batches are sampled without replacement, stochastic gradient descent samples with replacement in its standard form.}
$$\mu_\text{B}\sim\mathcal{N}(\mu, \frac{\sigma^2}{\text{M}})$$
For standard deviation:
$$\sigma_{\text{B}}=\sqrt{\frac{\Sigma_{\text{m}=1}^{\text{M}}(\weightedsum-\mu_{\text{B}})^2}{M}}$$
Then
$$\sqrt{M}(\sigma_{\text{B}}-\sigma)=\sqrt{M}\Big(\sqrt{\samplevar}-\sqrt{\sigma^2}\Big)$$
We want to rewrite $\sqrt{\samplevar}$. We take a Taylor expansion of $f(x)=\sqrt{x}$ around $a=\sigma^2$. With $x=\samplevar$:
\begin{align*}
\sqrt{x}&=\sqrt{\sigma^2}+\frac{1}{2\sqrt{\sigma^2}}(x-\sigma^2)+\mathcal{O}[(x-\sigma^2)^2]
\end{align*}
so
\begin{align*}
\sqrt{M}(\sigma_{\text{B}}-\sigma)&=\sqrt{M}\Bigg(\frac{1}{2\sqrt{\sigma^2}}\Big(\samplevar-\sigma^2\Big)+\\
&\qquad\qquad\bigosimple\Bigg) \\
&=\frac{\sqrt{M}}{2\sigma}\Big(\frac{1}{M}\samplevarnum-\sigma^2\Big)+\\
&\qquad\qquad\bigo\\
&=\frac{1}{2\sigma\sqrt{M}}\Big(\samplevarnum-M\sigma^2\Big)+\\
&\qquad\qquad\bigo
\end{align*}
consider $\samplevarnum$. We know that $E[\weightedsum]=\mu$ and write
\begin{align*}
&\samplevarnum\\
=&\Sigma_{\text{m=1}}^{\text{M}}((\weightedsum-\mu)-(\mu_{\text{B}}-\mu))^2\\
=&\Sigma_{\text{m=1}}^{\text{M}}((\weightedsum-\mu)^2+(\mu_{\text{B}}-\mu)^2-2(\weightedsum-\mu)(\mu_{\text{B}}-\mu))\\
=&\Sigma_{\text{m=1}}^{\text{M}}(\weightedsum-\mu)^2+M(\mu_{\text{B}}-\mu)^2-2(\mu_{\text{B}}-\mu)\Sigma_{\text{m=1}}^{\text{M}}(\weightedsum-\mu)\\
=&\Sigma_{\text{m=1}}^{\text{M}}(\weightedsum-\mu)^2-M(\mu_{\text{B}}-\mu)^2\\
=&\Sigma_{\text{m=1}}^{\text{M}}((\weightedsum-\mu)^2-(\mu_{\text{B}}-\mu)^2)
\end{align*}
then
\begin{align*}
\sqrt{M}(\sigma_{\text{B}}-\sigma)&=\frac{1}{2\sigma\sqrt{M}}\Big(\Sigma_{\text{m=1}}^{\text{M}}((\weightedsum-\mu)^2-(\mu_{\text{B}}-\mu)^2)-M\sigma^2\Big)+\\
&\qquad\qquad\bigo\\
&=\frac{1}{2\sigma\sqrt{M}}\Big(\Sigma_{\text{m=1}}^{\text{M}}(\weightedsum-\mu)^2-\Sigma_{\text{m=1}}^{\text{M}}(\mu_{\text{B}}-\mu)^2-M\sigma^2\Big)+\\
&\qquad\qquad\bigo\\
&=\frac{1}{2\sigma\sqrt{M}}\Big(\Sigma_{\text{m=1}}^{\text{M}}((\weightedsum-\mu)^2-\sigma^2)-\Sigma_{\text{m=1}}^{\text{M}}(\mu_{\text{B}}-\mu)^2\Big)+\\
&\qquad\qquad\bigo\\
&=\frac{1}{2\sigma\sqrt{M}}\Sigma_{\text{m=1}}^{\text{M}}((\weightedsum-\mu)^2-\sigma^2)\\
&\quad-\frac{1}{2\sigma\sqrt{M}}\Sigma_{\text{m=1}}^{\text{M}}(\mu_{\text{B}}-\mu)^2\\
&\quad+\bigo\\
&=\underbrace{\frac{1}{2\sigma\sqrt{M}}\Sigma_{\text{m=1}}^{\text{M}}((\weightedsum-\mu)^2-\sigma^2)}_{\text{term A}}\\
&\quad-\underbrace{\frac{\sqrt{M}}{2\sigma}(\mu_{\text{B}}-\mu)^2}_{\text{term B}}\\
&\quad+\underbrace{\bigo}_{\text{term C}}
\end{align*}
We go through each term in turn
\myparagraph{Term A}
We have
$$\text{Term A}=\frac{1}{2\sigma\sqrt{M}}\Sigma_{\text{m=1}}^{\text{M}}((\weightedsum-\mu)^2-\sigma^2)$$
where $\Sigma_{\text{m=1}}^{\text{M}}(\weightedsum-\mu)^2$ is the sum of $M$ RVs $(\weightedsum-\mu)^2$. Note that since $E[\weightedsum]=\mu$ it holds that $E[(\weightedsum-\mu)^2]=\sigma^2$. Since $(\weightedsum-\mu)^2$ is sampled approximately iid (by assumptions above), for large enough M by CLT it holds approximately that
$$\Sigma_{\text{m=1}}^{\text{M}}(\weightedsum-\mu)^2\sim\mathcal{N}(M\sigma^2, M\text{Var(}(\weightedsum-\mu)^2))$$
where
\begin{align*}
\text{Var(}(\weightedsum-\mu)^2)&=E[(\weightedsum-\mu)^{2*2}]-E[(\weightedsum-\mu)^2]^2 \\
&=E[(\weightedsum-\mu)^4]-\sigma^4
\end{align*}
Then
\begin{align*}
\Sigma_{\text{m=1}}^{\text{M}}((\weightedsum-\mu)^2-\sigma^2)\sim\mathcal{N}(0, M*E[(\weightedsum-\mu)^4]-M\sigma^4)
\end{align*}
so
$$\text{Term A}\sim\mathcal{N}(0, \frac{E[(\weightedsum-\mu)^4]-\sigma^4}{4\sigma^2})$$
\myparagraph{Term B}
We have
$$\text{Term B}=\frac{\sqrt{M}}{2\sigma}(\mu_{\text{B}}-\mu)^2=\frac{1}{2\sigma}\sqrt{M}(\mu_{\text{B}}-\mu)(\mu_{\text{B}}-\mu)$$
Consider $(\mu_{\text{B}}-\mu)$. As $\mu_{\text{B}}\xrightarrow[]{p}\mu$ when $M\rightarrow\infty$ we have $\mu_{\text{B}}-\mu\xrightarrow[]{p}0$. We also have 
$$\sqrt{M}(\mu_{\text{B}}-\mu)=\frac{\Sigma_{\text{m}=1}^{\text{M}}\weightedsum}{\sqrt{M}}-\sqrt{M}\mu$$
which by CLT is approximately Gaussian for large $M$. We can then make use of the Cramer-Slutzky Theorem, which states that if $(X_n)_{n\geq1}$ and $(Y_n)_{n\geq1}$ are two sequences such that $X_n\xrightarrow[]{d}X$ and $Y_n\xrightarrow[]{p}a$ as $n\rightarrow\infty$ where $a$ is a constant, then as $n\rightarrow\infty$, it holds that $X_n*Y_n\xrightarrow[]{d}X*a$. Thus, Term B is approximately 0 for large M.
\myparagraph{Term C}
We have
$$\text{Term C}=\bigo$$
Since $E[(\weightedsum-\mu)^2]=\sigma^2$ we can make the same use of Cramer-Slutzky as for \textit{Term B}, such that Term C is approximately 0 for large M.
\myparagraph{Finalizing the distribution}
We have approximately
$$\sqrt{M}(\sigma_{\text{B}}-\sigma)\sim\mathcal{N}(0, \frac{E[(\weightedsum-\mu)^4]-\sigma^4}{4\sigma^2})$$
so
$$\sigma_{\text{B}}\sim\mathcal{N}(\sigma, \frac{E[(\weightedsum-\mu)^4]-\sigma^4}{4\sigma^2 M})$$

\subsection{Prior}
\label{appendix:prior}
Here we make use of the stochasticity from BN modeled in the Appendix section \ref{appendix:dist_mu_sigma}, to evaluate the implied prior on the stochastic variables for a BN network. Specifically, we consider a BN network with fully connected layers and BN applied to each layer, trained with L2-regularization (weight decay). In the following, we make use of the simplifying assumptions of no scale and shift tranformations, BN applied to each layer, and independent input units to each layer. 

Equivalence between the objectives of Eq. (1) and (2) requires:
\begin{align}\label{eq:required_condition}
\begin{split}
\pdvthetak\KLprior&=N\tau\pdvthetak\regapproxparam \\
&=N\tau\pdvthetak\sum_{l=1}^L\lambda_l||\textbf{W}^l||^2
\end{split}
\end{align}
where $\theta_k\in\bm{\theta}$, and $\bm{\theta}$ is the set of weights in the network. To proceed with the LHS of Eq. (\ref{eq:required_condition}) we first need to find the approximate posterior $\qthetaomega$ that batch normalization induces. As shown in Appendix \ref{appendix:dist_mu_sigma}, with some weak assumptions and approximations the Central Limit Theorem (CLT) yields Gaussian distributions of the stochastic variables $\means_{\batch}^{u}, \vars_{\batch}^{u}$, for large enough $M$. For any BN unit $u$:
\begin{align}
\begin{split}
&\means_{\batch}^u \appropto \mathcal{N}(\means^u, \frac{(\vars^u)^2}{M}),\\
&\vars_{\batch}^u \appropto \mathcal{N}(\vars^u, \frac{\fourthmoment-(\vars^u)^4}{4(\vars^u)^2M})
\label{eq:approx_distrib}  
\end{split}
\end{align}
where $\means^u$ and $\vars^u$ are \textit{population-level} moments (i.e. moments over \textbf{D}). 

We assume that $\qparam$ and $\prior$ factorize over all stochastic variables.\footnote{The empirical distributions have been numerically checked to be linearly independent and the joint distribution is close to a bi-variate Gaussian.} We use $i$ as an index of the set of stochastic variables. As shown in Eq. (\ref{eq:kldivergencefactorization}) in Appendix \ref{appendix:factorized_gaussians}, the factorized distributions yield:
\begin{equation*}
\KLomegap=\sum_{i}\KLomegapomegai 
\end{equation*}
Note that each BN unit produces two $\KLomegapomegai$ terms: one for $\omega_i=\means_{\batch}^u$ and one for $\omega_i=\vars_{\batch}^u$. We consider these terms for one particular BN unit $u$, and drop the index $i$ for brevity. We use a Gaussian prior $p(\omega_i)=\mathcal{N}(\mu_p, \sigma_p^2)$ and, for consistency, use the notation $\qthetaomegai=\mathcal{N}(\mu_q,\sigma_q^2)$. As shown in Eq. (\ref{eq:kldivergencegaussian}) in Appendix \ref{appendix:factorized_gaussians}:
\begin{equation*}
\KLomegapomegai=\ln\frac{\sigma_p}{\sigma_q}+\frac{\sigma_q^2+(\mu_q-\mu_p)^2}{2\sigma_p^2}-\frac{1}{2}
\end{equation*}
Since $\theta_k$ changes during training, a prior cannot depend on $\theta_k$ so $\pdvthetak(\mu_p)=\pdvthetak(\sigma_p)=0$. Letting $(\cdot)'$ denote $\pdvthetak(\cdot)$:
\begin{equation}\label{eq:kldivergencederiv}
\pdvthetak\KLomegapomegai=\frac{\sigma_q\sigma_q'+\mu_q\mu_q'-\mu_p\mu_q'}{\sigma_p^2}-\frac{\sigma_q'}{\sigma_q}
\end{equation}

We need not consider $\theta_k$ past a previous layer's BN, since a normalization step is performed before scale and shift. In the general case with a given Gaussian $\prior$, Eq. \ref{eq:kldivergencederiv} evaluated on all BN units' means and standard deviations \wrt all $\theta_k$ up to a previous layer's BN, would yield an expression for a custom $N\tau\pdvthetak\regapproxparam$ that could be used for an exact VI treatment of BN.

In our reconciliation of weight decay however, given our assumptions of no scale and shift and BN applied to each layer, we need only consider the \textit{weights} in the same layer as the BN unit. This means that the stochastic variables in layer $l$ are only affected by weights in $\theta_k\in\bm{W}^{l}$ (i.e. not the scale and shift variables operating on the input to the layer). We denote a weight connecting the $k$:th input unit to the $u$:th BN unit by $\textbf{W}^{(u, k)}$. For such weights, we need to derive $\mu_q'$ and $\sigma_q'$, for two cases: $\omega_i=\means_{\batch}^u$ and $\omega_i=\vars_{\batch}^u$. We denote the priors of the mean and std. dev for $\means_{\batch}^u$ by $\mu_{\mu,q}$ and $\sigma_{\mu,q}$, and for $\vars_{\batch}^u$ by $\mu_{\sigma,q}$ and $\sigma_{\sigma,q}$. Using the distributions modeled in Eq. \ref{eq:approx_distrib}:
\paragraph{Case 1: $\omega_i=\means_{\batch}^u$}
\noindent
\begin{align}
\begin{split}
\nonumber\mu_{\mu,q}&=\sum_{\bm{x}\in\textbf{D}}\frac{\wux}{N}=\wuxbar\\
\mu_{\mu,q}'&=\xmean=\xbark\\
\nonumber\sigma_{\mu,q}&=\sqrt{\frac{(\vars^u)^2}{M}}=\sqrt{\frac{\sum_{\bm{x}\in\textbf{D}}(\wux-\mu_{q})^2}{NM}}\\
\sigma_{\mu,q}'&=\frac{1}{2}\sigma_{q}^{-1}\sum_{\bm{x}\in\textbf{D}}\frac{2(\wux-\mu_{q})(\bm{x}_k-\xbark)}{NM}
=\sigma_{q}^{-1}\bigg(\covterms\bigg)M^{-1}\end{split}
\end{align}
where there are $K$ input units to the layer.
\paragraph{Case 2: $\omega_i=\sigma_{\batch}^u$}
\noindent
\begin{align}
\begin{split}
\nonumber\mu_{\sigma,q}&=\sqrt{\frac{\sum_{\bm{x}\in\textbf{D}}(\wux-\mu_{q})^2}{N}}=\sigma_{\mu,q}M^{\frac{1}{2}}\\
\mu_{\sigma,q}'&=\sigma_{\mu,q}^{-1}M^{-\frac{1}{2}}\bigg(\covterms)\bigg)\\
\nonumber\sigma_{\sigma,q}&=\frac{\fourthmoment-(\vars^u)^4}{4(\vars^u)^2M}\\
\sigma_{\sigma,q}'&=\frac{\fourthmoment'\sigma^u-2(\sigma^u)^4(\sigma^u)'-2(\sigma^u)'\fourthmoment}{4(\sigma^u)^3M}\\
\end{split}
\end{align}
Combining these results with Eq. \ref{eq:kldivergencederiv} we find that taking $\KLomegapomegai$ for the mean and variance of a single BN unit $u$ wrt the weight from input unit $k$:
\begin{align*}
\begin{split}
&\pdvWuk\KLmean+\pdvWuk\KLstd\\
=&\frac{\sigma_{\mu,q}\sigma_{\mu,q}'+\mu_{\mu,q}\mu_{\mu,q}'-\mu_{\mu,p}\mu_{\mu,q}'}{\sigma_{\mu,p}^2}-\frac{\sigma_{\mu,q}'}{\sigma_{\mu,q}}\\
+&\frac{\sigma_{\sigma,q}\sigma_{\sigma,q}'+\mu_{\sigma,q}\mu_{\sigma,q}'-\mu_{\sigma,p}\mu_{\sigma,q}'}{\sigma_{\sigma,p}^2}-\frac{\sigma_{\sigma,q}'}{\sigma_{\sigma,q}}\\
=&\frac{\mathcal{O}(M^{-1})+\xbark\wuxbar-\mu_{\mu,p}\xbark}{\sigma_{\mu,p}^2}-\mathcal{O}(M^{-1})\\
+&\frac{\mathcal{O}(M^{-2})+\covterms-\mu_{\sigma,p}\mathcal{O}(M^{-\frac{1}{2}})}{\sigma_{\sigma,p}^2}\\
-&\frac{\fourthmoment'\sigma^u-2(\sigma^u)^4(\sigma^u)'-2(\sigma^u)'\fourthmoment}{\fourthmoment\sigma^u-(\sigma^u)^5}
\end{split}
\end{align*}
where we summarize the terms scaled by $M$ with $\mathcal{O}$-notation. We see that if we let $M\rightarrow\infty$, $\mu_{\mu,p}=0$, $\sigma_{\mu,p}\rightarrow\infty$, $\mu_{\sigma,p}=0$ and $\sigma_{\sigma,p}$ is small enough, then:
\begin{equation*}
\pdvWuk\bigg(\KLmean+\KLstd\bigg)\approx\frac{\covterms}{\sigma_{\sigma,p}^2}
\end{equation*}
such that each BN layer yields the following:
\begin{equation}\label{eq:kl_layer_terms}
\sum_{u}\sum_{i=1}^K\pdvWik\bigg(\KLmean+\KLstd\bigg)\approx\sum_{u}\frac{\sum_{i=1}^K\textbf{W}^{u,i}\sum_{i2=1}^K\text{Cov}(x_i,x_{i2})}{\sigma_{\sigma,p,u}^2}
\end{equation}
where we denote the prior for the std. dev. of the std. dev. of BN unit $u$ by $\sigma_{\sigma,p,u}$. Given our assumptions of no scale and shift from the previous layer, and independent input features in every layer, Eq. \ref{eq:kl_layer_terms} reduces to:
\begin{equation*}
\sum_{u}\sum_{i=1}^K\frac{\textbf{W}^{u,i}}{\sigma_{\sigma,p}^2}
\end{equation*}
if the same prior is chosen for each BN unit in the layer. We therefore find that Eq. \ref{eq:required_condition} is reconciled by $p(\means_{\batch}^u)\rightarrow\mathcal{N}(0, \infty)$ and $p(\vars_{\batch}^u)\rightarrow\mathcal{N}(0, \frac{1}{2N\tau\lambda_l})$, if $\frac{1}{2N\tau\lambda_l}$ is small enough, which is the case if $N$ is large.

\subsection{predictive distribution properties}\label{appendix:predictivemoments}
This section provides derivations of properties of the predictive distribution $\approxpredictive$ in section 3.4, 
following \cite{Gal2016Uncertainty}. We first find the first two modes of the approximate predictive distribution (with the second mode applicable to regression), then show how to estimate the predictive log likelihood, a measure of uncertainty quality used in the evaluation.
\paragraph{Predictive mean}
Assuming Gaussian iid noise defined by model precision $\tau$, i.e. $\fomegaxy=p(\*y|\fomegax)=\mathcal{N}(\*y;\fomegax,\tau^{-1}\textbf{I})$:
\begin{align*}
\expec_{p^*}[\lab]&=\int\lab \approxpredictive d\lab \\
&=\int_{\*y}\*y\Big(\int_{\bm{\omega}} \fomegaxy \qparam d\param\Big)\text{d}\*y \\
&=\int_{\*y}\*y\Big(\int_{\bm{\omega}} \mathcal{N}(\*y;\fomegax,\tau^{-1}\textbf{I})\qthetaomega \text{d}\bm{\omega}\Big)\text{d}\*y \\
&=\int_{\bm{\omega}}\Big(\int_{\*y}\*y\mathcal{N}(\*y;\fomegax,\tau^{-1}\textbf{I})\text{d}\*y\Big)\qthetaomega \text{d}\bm{\omega} \\
&=\int_{\bm{\omega}}\fomegax\qthetaomega\text{d}\bm{\omega} \\
&\approx \frac{1}{T}\sum_{i=1}^T \fomegaxsample
\end{align*}
where we take the MC Integral with $T$ samples of $\bm{\omega}$ for the approximation in the final step. 
\paragraph{Predictive variance}
For regression, our goal is to estimate:
\begin{align*}
\text{Cov}_{p^*}[\*y]=\expec_{p^*}[\*y^{\intercal}\*y]-\expec_{p^*}[\*y]^\intercal \expec_{p^*}[\*y]
\end{align*}
We find that:
\begin{align*}
\expec_{p^*}[\*y^{\intercal}\*y]&=\int_{\*y}\*y^{\intercal}\*y\approxpredictive\text{d}\*y \\
&=\int_{\*y}\*y^{\intercal}\*y\Big(\int_{\bm{\omega}} \fomegaxy \qparam d\param\Big)\text{d}\*y \\
&=\int_{\bm{\omega}}\Big(\int_{\*y}\*y^{\intercal}\*y \fomegaxy\text{d}\*y\Big)\qthetaomega \text{d}\bm{\omega} \\
&=\int_{\bm{\omega}}\Big(\text{Cov}_{\fomegaxy}(\*y) +\expec_{\fomegaxy}[\*y]^\intercal \expec_{\fomegaxy}[\*y]\Big)\qthetaomega \text{d}\bm{\omega} \\
&=\int_{\bm{\omega}}\Big(\tau^{-1}\textbf{I}+\fomegax^\intercal\fomegax\Big)\qthetaomega \text{d}\bm{\omega} \\
&=\tau^{-1}\textbf{I}+E_{\qthetaomega}[\fomegax^\intercal\fomegax] \\
&\approx \tau^{-1}\textbf{I}+\frac{1}{T}\sum_{i=1}^T \fomegaxsample^\intercal\fomegaxsample
\end{align*}
where we use MC integration with $T$ samples for the final step. The predictive covariance matrix is given by:
\begin{align*}
\text{Cov}_{p^*}[\*y]\approx \tau^{-1}\textbf{I}+\frac{1}{T}\sum_{i=1}^T \fomegaxsample^\intercal\fomegaxsample-\expec_{p^*}[\lab]^\intercal\expec_{p^*}[\lab]
\end{align*}
which is the sum of the variance from observation noise and the sample covariance from $T$ stochastic forward passes though the network.

The form of $p^*$ can be approximated by a Gaussian for each output dimension (for regression). We assume bounded domains for each input dimension, wide layers throughout the network, and a uni-modal distribution of weights centered at 0. By the Liapounov CLT condition, the first layer then receives approximately Gaussian inputs (a proof can be found in \cite{Lehmann1999}). Having sampled $\mu_{\batch}^u$ and $\sigma_{\batch}^u$ from a mini-batch, each BN unit's output is bounded. CLT thereby continues to hold for deeper layers, including $\fomegax=\*W^L\*x^L$. A similar motivation for a Gaussian approximation of Dropout has been presented by \cite{Wang2013}.
\paragraph{Predictive Log Likelihood}
We use the Predictive Log Likelihood (PLL) as a measure to estimate the model's uncertainty quality. For a certain test point $(\*y_i, \*x_i)$, the PLL definition and approximation can be expressed as:
\begin{align*}
\text{PLL}(\fomegax, (\*y_i, \*x_i))&=\log p(\*y_i|\fomegaxi)\\
&=\log\int f_{\param}(\samplei,\labi)\posterior\dparam\\
&\approx\log\int f_{\param}(\samplei,\labi)\qparam\dparam\\
&\approx\log\frac{1}{T}\sum_{j=1}^Tp(\*y_i|\fomegaxisamplej)\\
\end{align*}
where $\paramj$ represents a sampled set of stochastic parameters from the approximate posterior distrubtion $\qparam$ and we take a MC integration with $T$ samples. For regression, due to the iid Gaussian noise, we can further develop the derivation into the form we use when sampling:
\begin{align*}
\text{PLL}(\fomegax, (\*y_i, \*x_i))&=\log\frac{1}{T}\sum_{j=1}^T\mathcal{N}(\*y_i|\fomegaxisamplej, \tau^{-1}\*I)\\
&=\text{logsumexp}_{j=1,\dots,T}\big(-\frac{1}{2}\tau||\*y_i-\fomegaxisamplej||^2\big)\\
&-\log T-\frac{1}{2}\log2\pi+\frac{1}{2}\log\tau
\end{align*}
Note that PLL makes no assumption on the form of the approximate predictive distribution.

\subsection{Data}

To assess the uncertainty quality of the various methods studied we rely on eight standard regression datasets, listed in Table \ref{table:datasets}. Publicly available from the UCI Machine Learning Repository \citep{UniversityofCaliforniab} and Delve \citep{Ghahramani1996}, these datasets have been used to benchmark comparative models in recent related literature (see \cite{hernandez2015probabilistic}, \cite{Gal2015a}, \cite{Bui2016} and \cite{Li2017}).

For image classification, we applied MCBN using ResNet32 to CIFAR10.

For the image segmentation task, we applied MCBN using Bayesian SegNet on data from CamVid and PASCAL-VOC using models published in \cite{KendallBC15}.

\begin{table}[H]
	\caption{\textbf{Regression dataset summary.} Properties of the eight regression datasets used to evaluate MCBN. $N$ is the dataset size and $Q$ is the n.o. input features. Only one target feature was used -- we used heating load for the Energy Efficiency dataset, which contains multiple target features.}
	\vspace{2mm}
	\centering
	\begin{tabular}{lll}
		\toprule
		\textbf{Dataset name}         & $N$ & $Q$  \\ 
		\midrule
		Boston Housing                & 506        & 13                                            \\
		Concrete Compressive Strength & 1,030      & 8                                              \\
		Energy Efficiency             & 768        & 8                                 \\
		Kinematics 8nm                & 8,192      & 8                                              \\
		Power Plant                   & 9,568       & 4                                              \\
		Protein Tertiary Structure    & 45,730      & 9                                              \\
		Wine Quality (Red)            & 1,599       & 11                                             \\
		Yacht Hydrodynamics           & 308        & 6                     \\
		\bottomrule
	\end{tabular}
	\label{table:datasets}
	\vspace{-3mm}
\end{table}

\subsection{Extended experimental results}\label{appendix:extendedresults}
Below, we provide extended results measuring uncertainty quality. In Tables \ref{table:crps_results} and \ref{table:pll_results}, we provide tables showing the mean $\overline{\textrm{CRPS}}$ and $\overline{\textrm{PLL}}$ values for MCBN and MCDO. These results indicate that MCBN performs on par or better than MCDO across several datasets. In Table \ref{table:rawcrpspll} we provide the raw PLL and CRPS results for MCBN and MCDO. In Table \ref{table:rmse_results} we provide RMSE results of the MCBN and MCDO networks in comparison with non-stochastic BN and DO networks. These results indicate that the procedure of multiple forward passes in MCBN and MCDO show slight improvements in the accuracy of the network.

\begin{figure}[H]
	\centering
	\begin{tabular}{@{}c@{\hskip 1mm}c@{\hskip 1mm}c@{}}
		\includegraphics[width=0.3\linewidth]{figures/r_bostonHousing_MCBN} &
		\includegraphics[width=0.3\linewidth]{figures/r_bostonHousing_MNF} &
		\includegraphics[width=0.3\linewidth]{figures/r_bostonHousing_MCDO} \\
		\vspace{2mm}
		\includegraphics[width=0.3\linewidth]{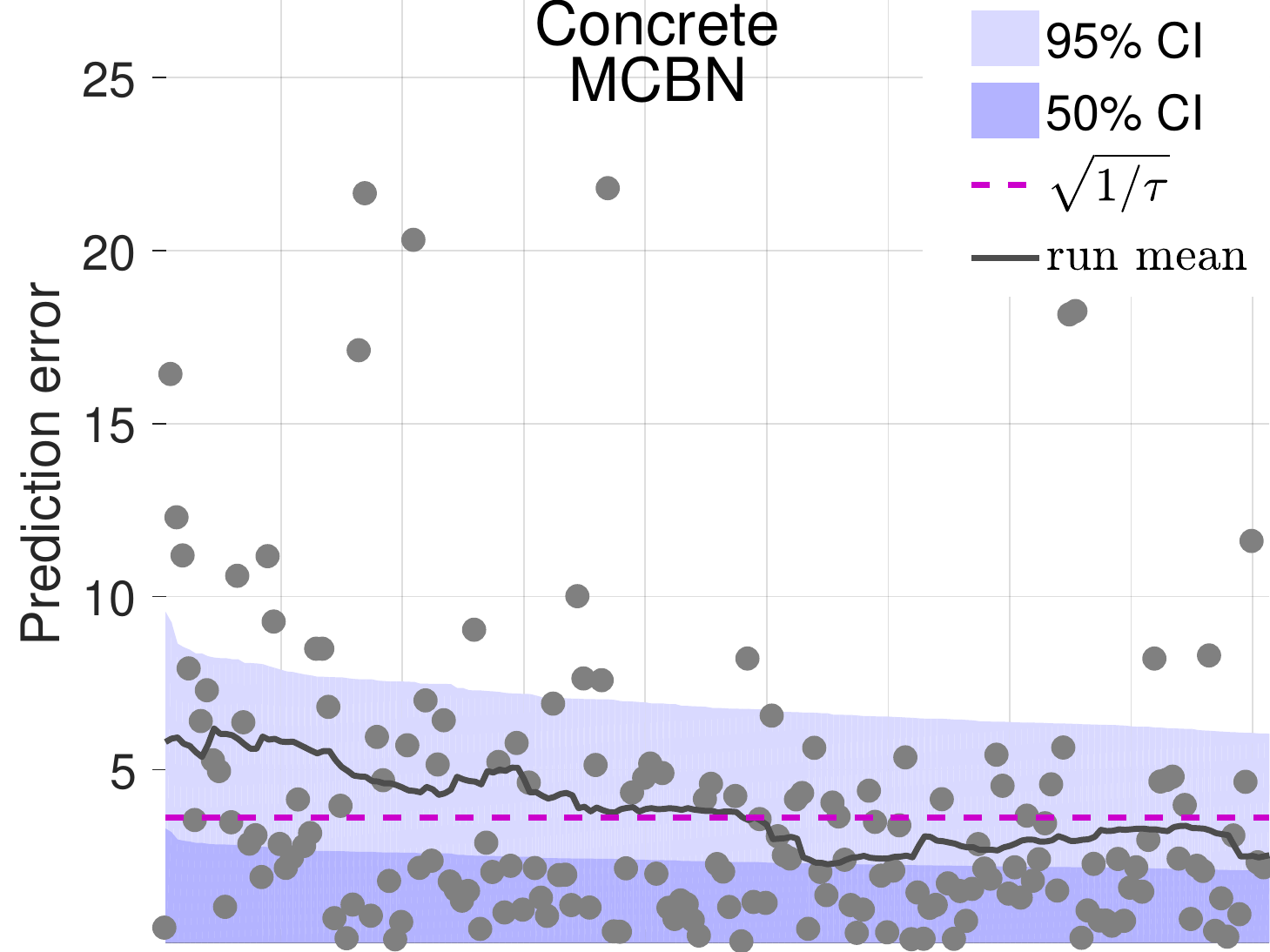} &
		\includegraphics[width=0.3\linewidth]{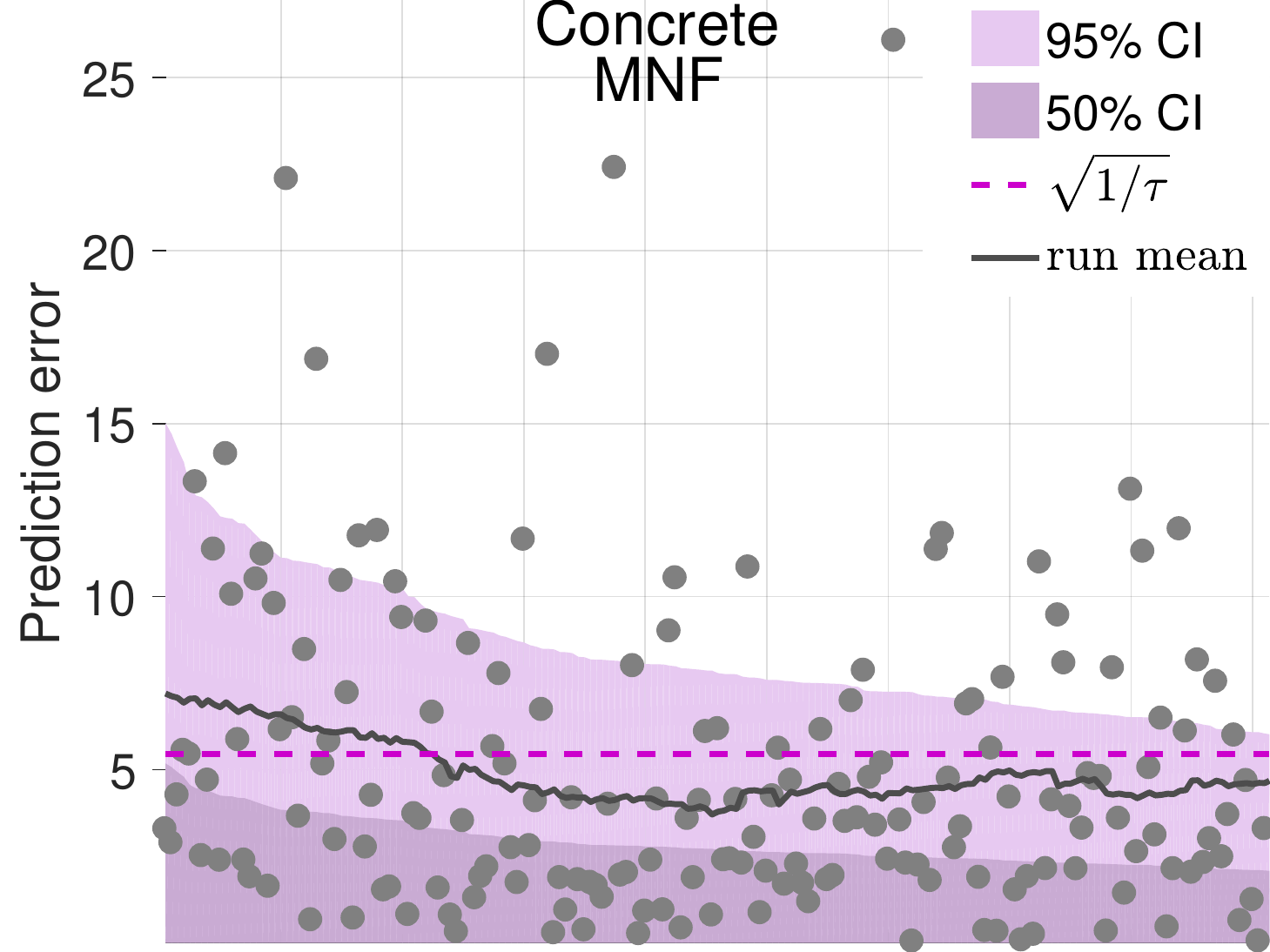} &
		\includegraphics[width=0.3\linewidth]{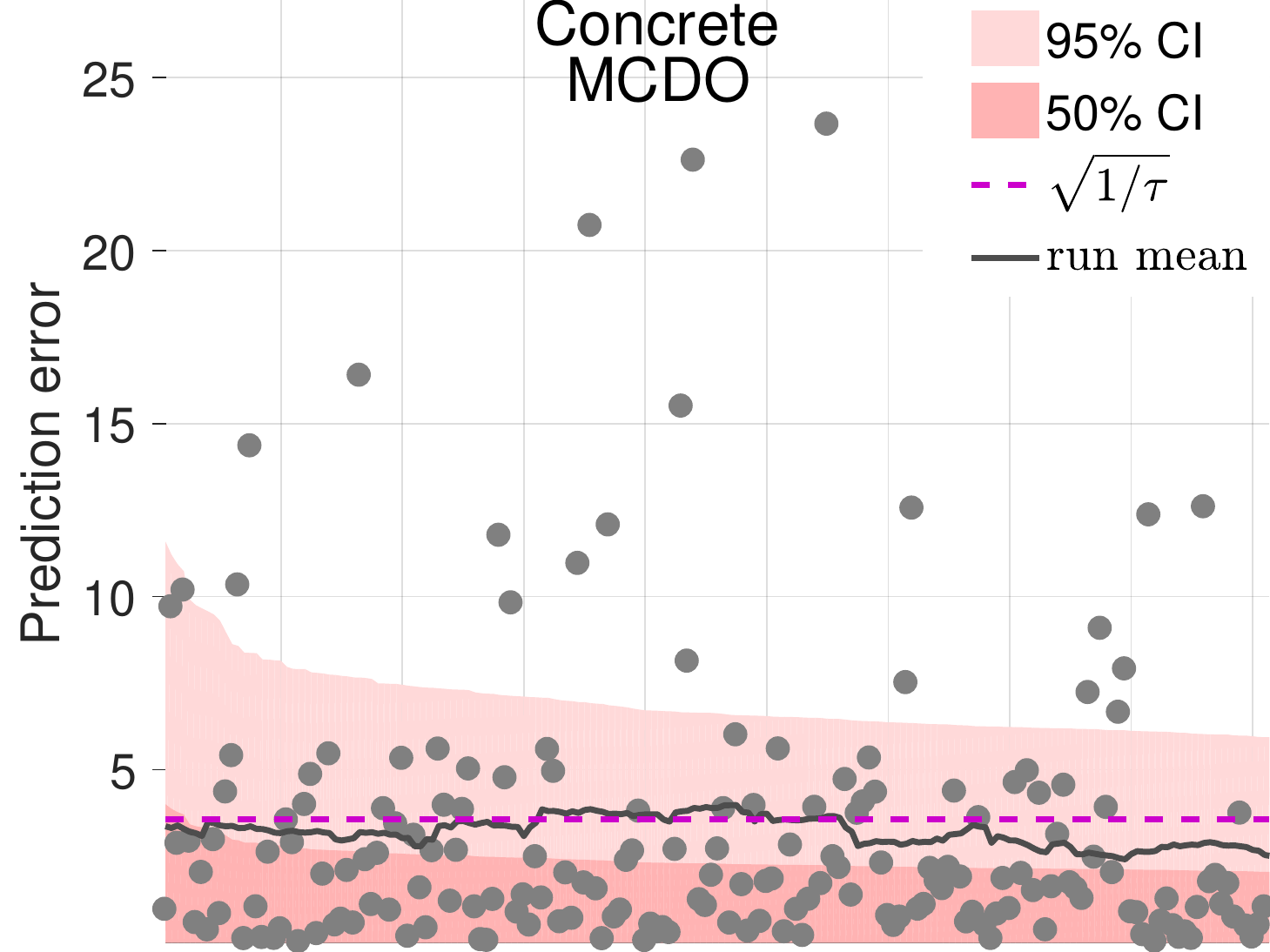} \\
		\vspace{2mm}
		\includegraphics[width=0.3\linewidth]{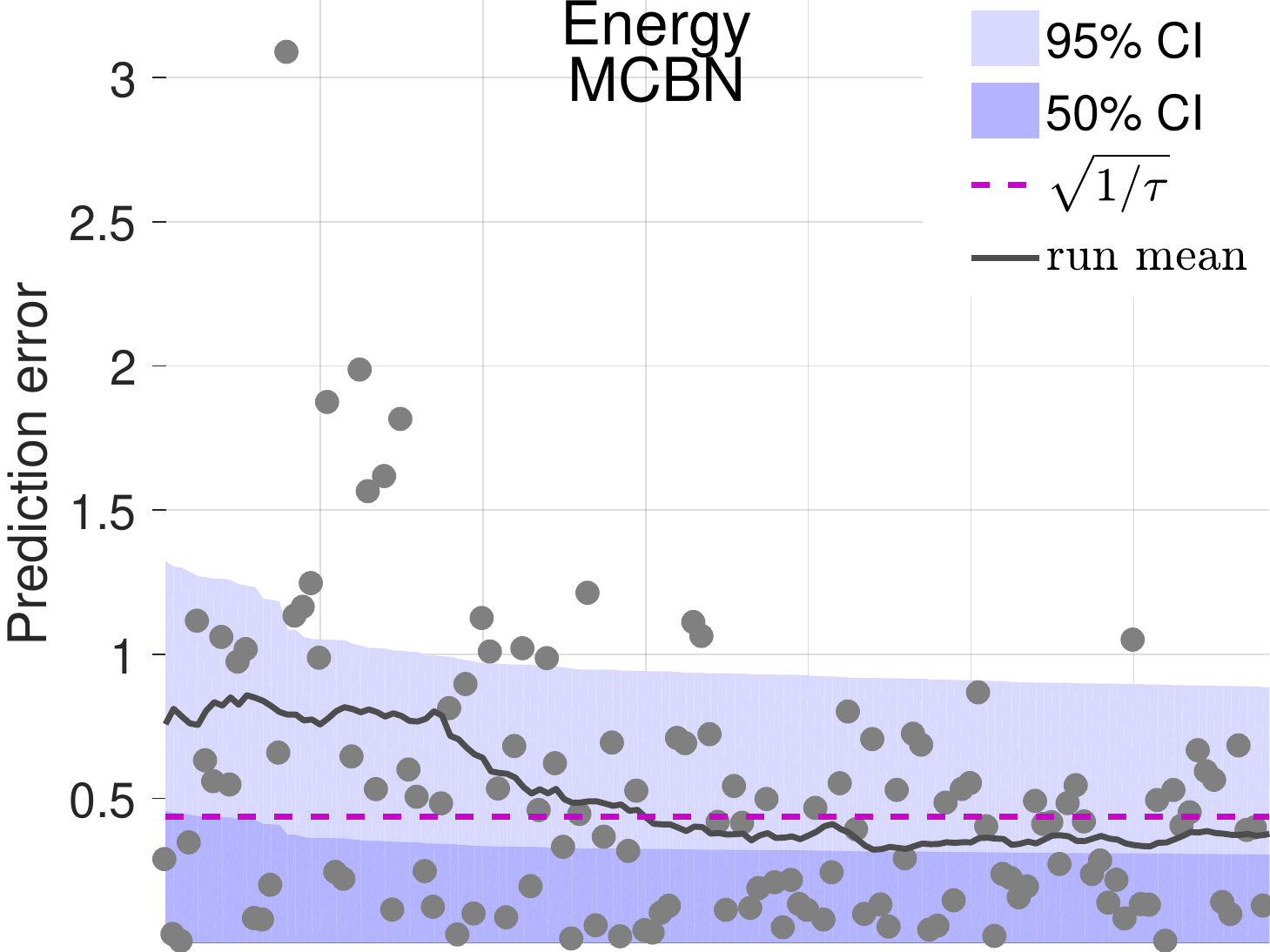} &
		\includegraphics[width=0.3\linewidth]{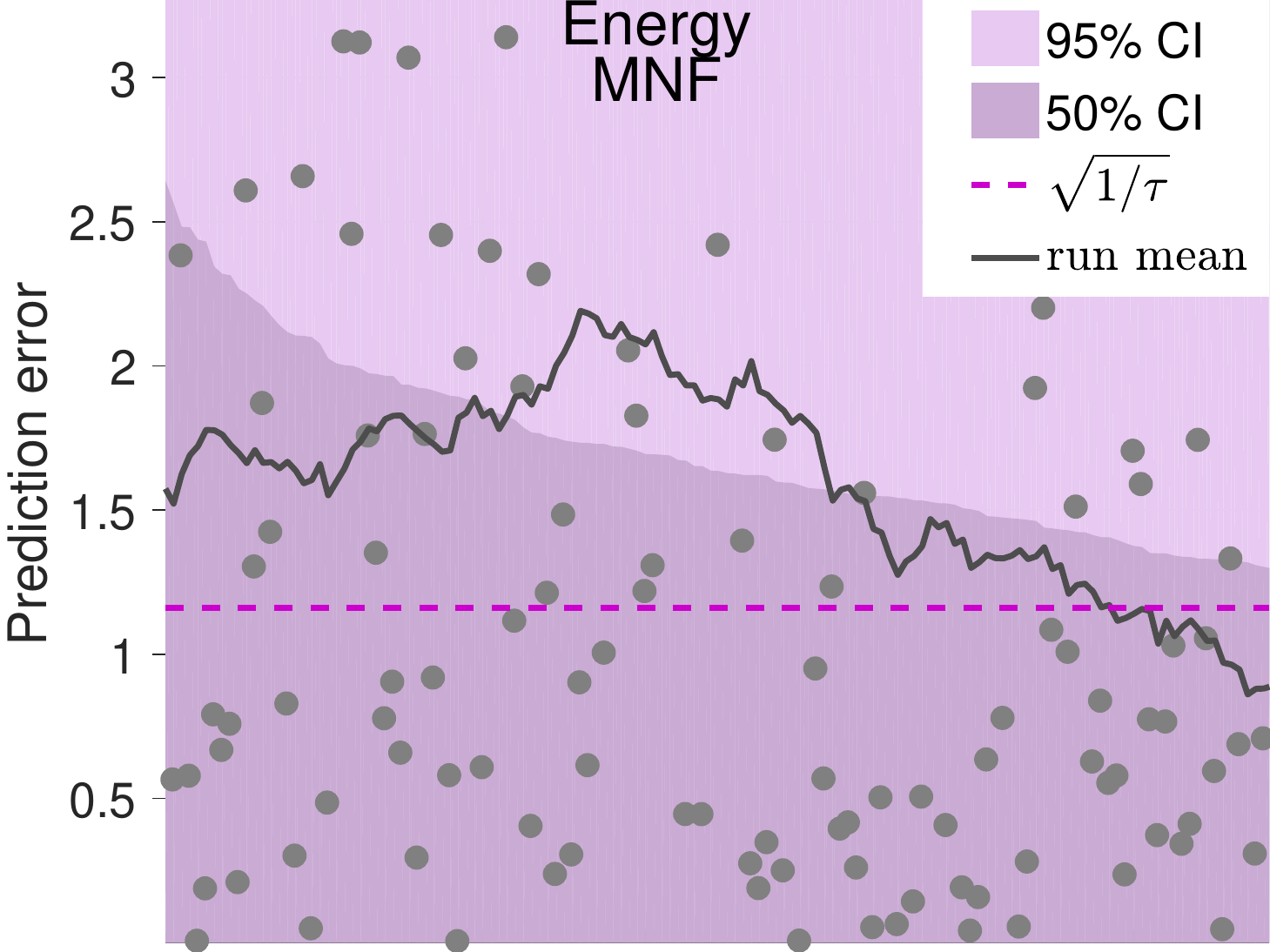} &
		\includegraphics[width=0.3\linewidth]{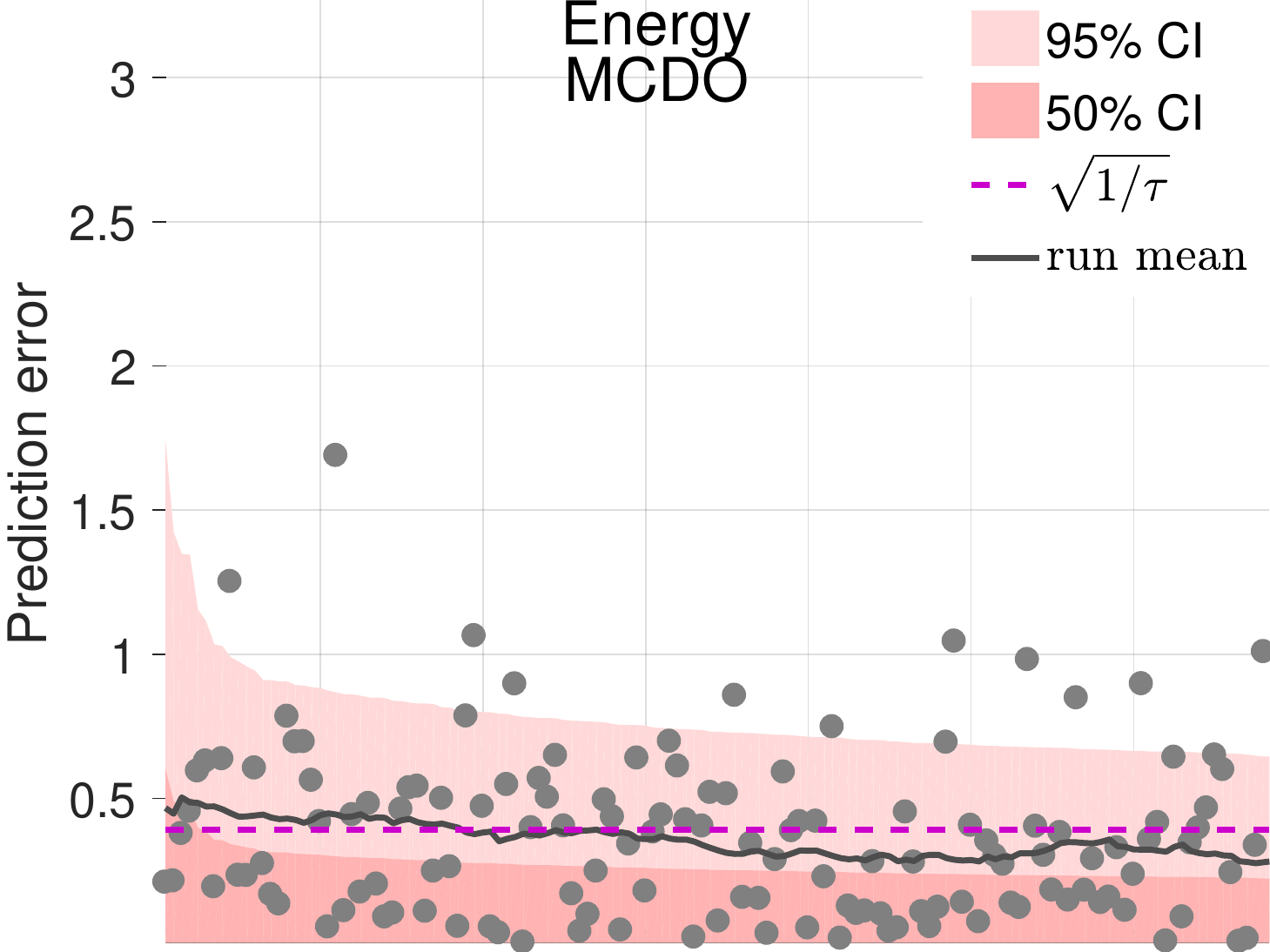} \\
		\vspace{2mm}
		\includegraphics[width=0.3\linewidth]{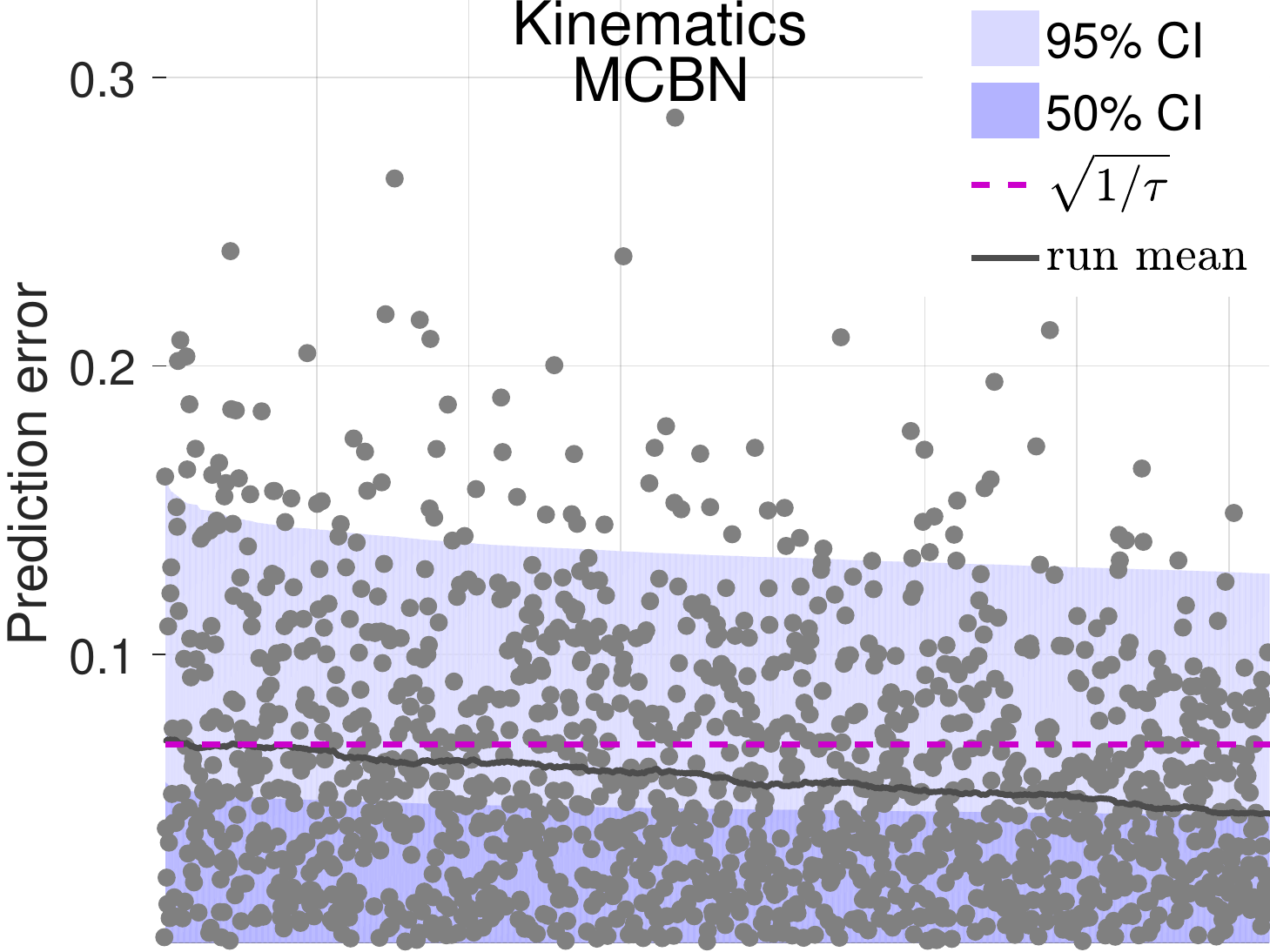} &
		\includegraphics[width=0.3\linewidth]{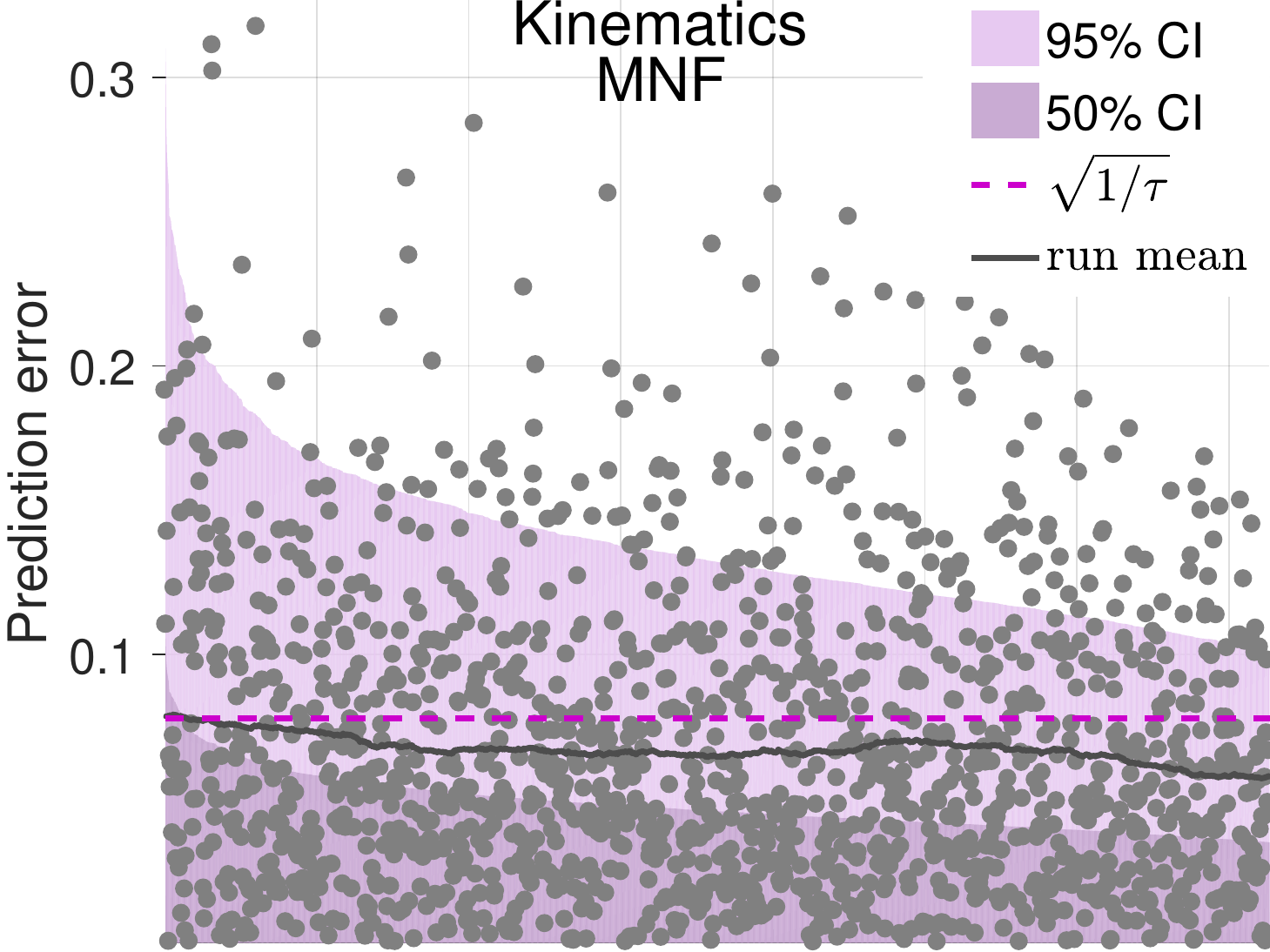} &
		\includegraphics[width=0.3\linewidth]{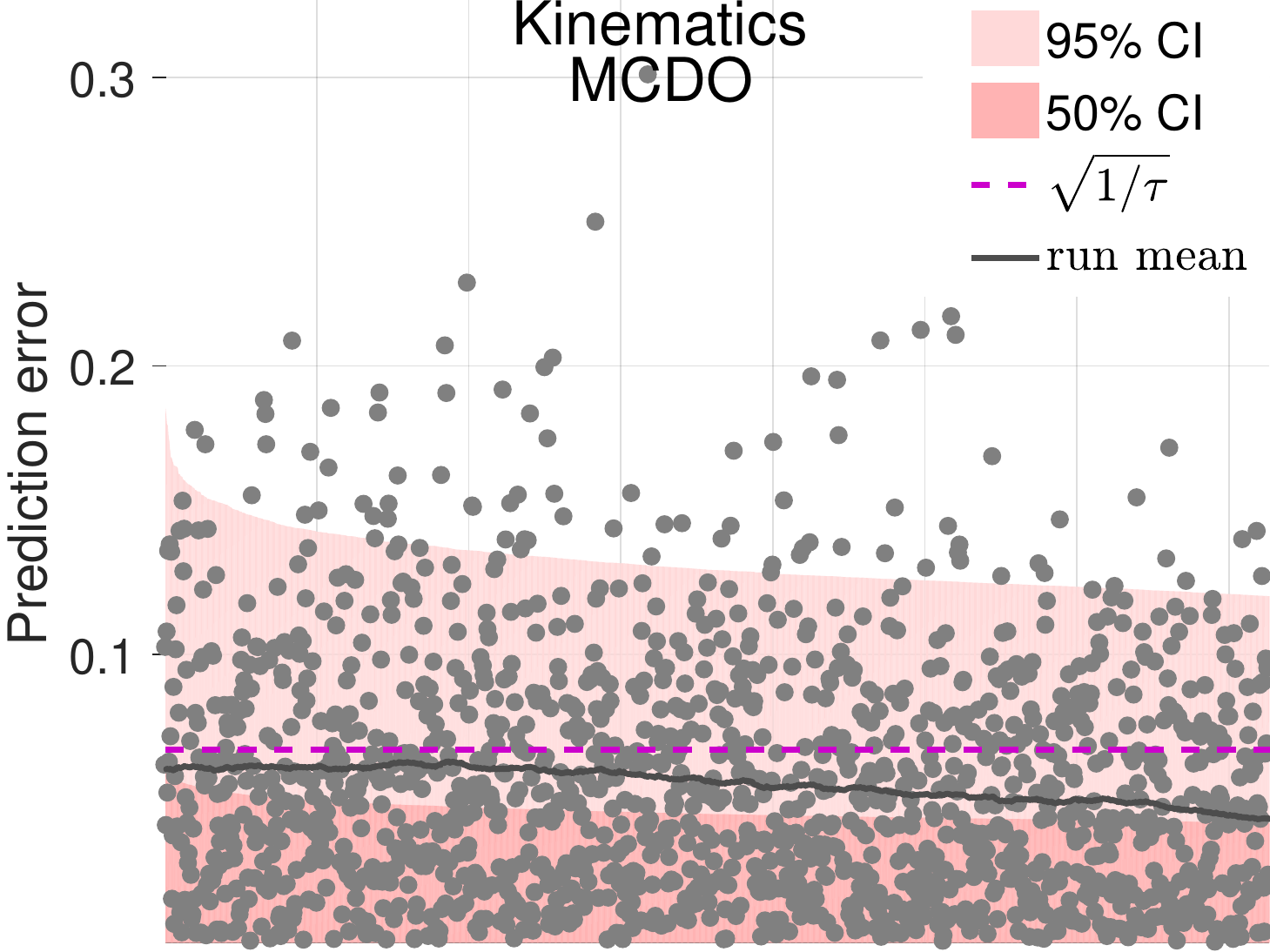} \\
	\end{tabular}
	\vspace{-3mm}
	\caption{Errors in predictions (gray dots) sorted by estimated uncertainty on select datasets. The shaded areas show model uncertainty (light area 95\% CI, dark area 50\% CI). Gray dots show absolute prediction errors on the test set, and the gray line depicts a running mean of the errors. The dashed line indicates the optimized constant uncertainty. A correlation between estimated uncertainty (shaded area) and mean error (gray) indicates the uncertainty estimates are meaningful for estimating errors.}
	\label{fig:sortplotsA1}
	\vspace{-3mm}
\end{figure}

\begin{figure}[H]
	\begin{tabular}{@{}c@{\hskip 1mm}c@{\hskip 1mm}c@{}}
		\includegraphics[width=0.3\linewidth]{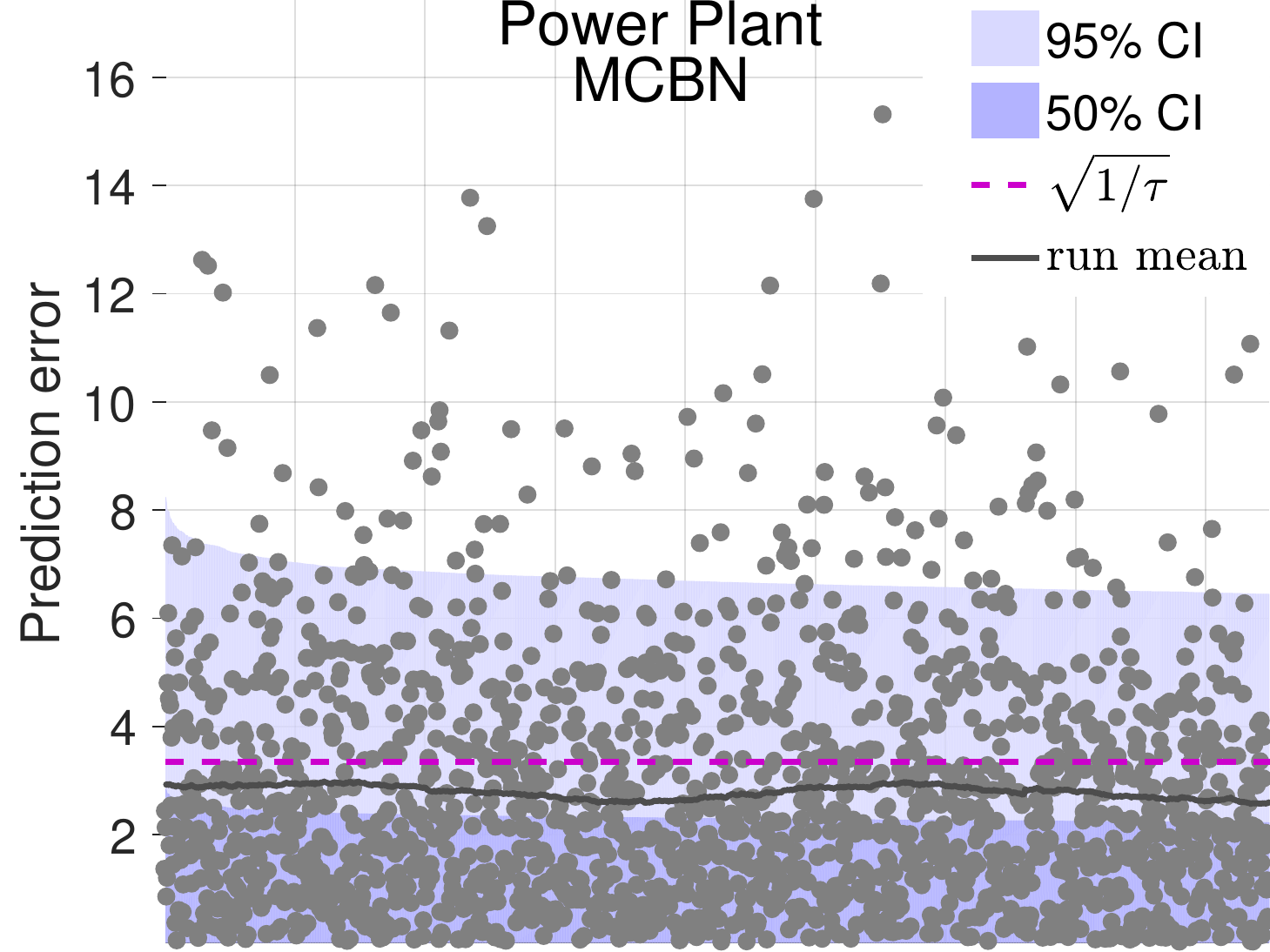} &
		\includegraphics[width=0.3\linewidth]{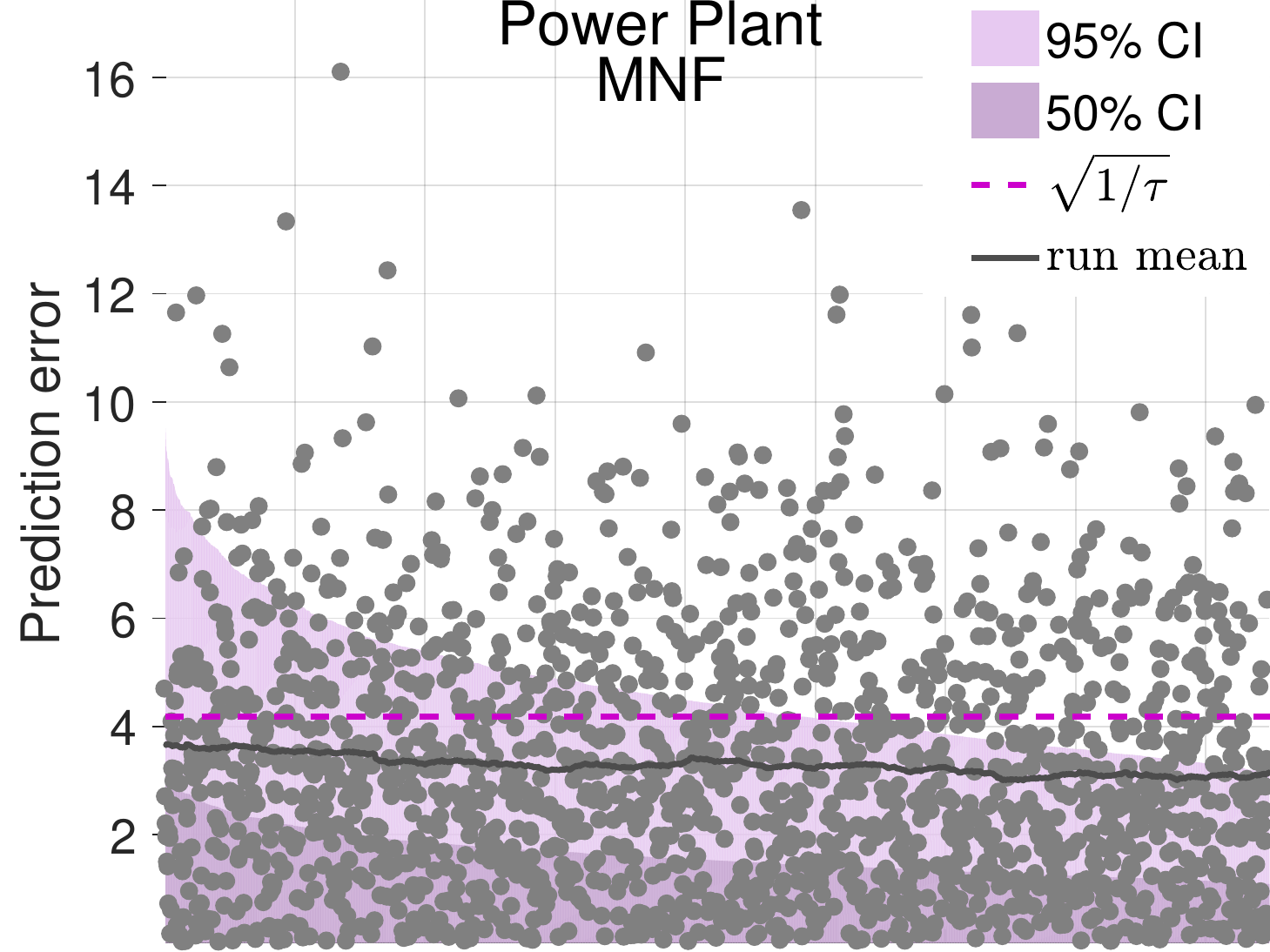}  &
		\includegraphics[width=0.3\linewidth]{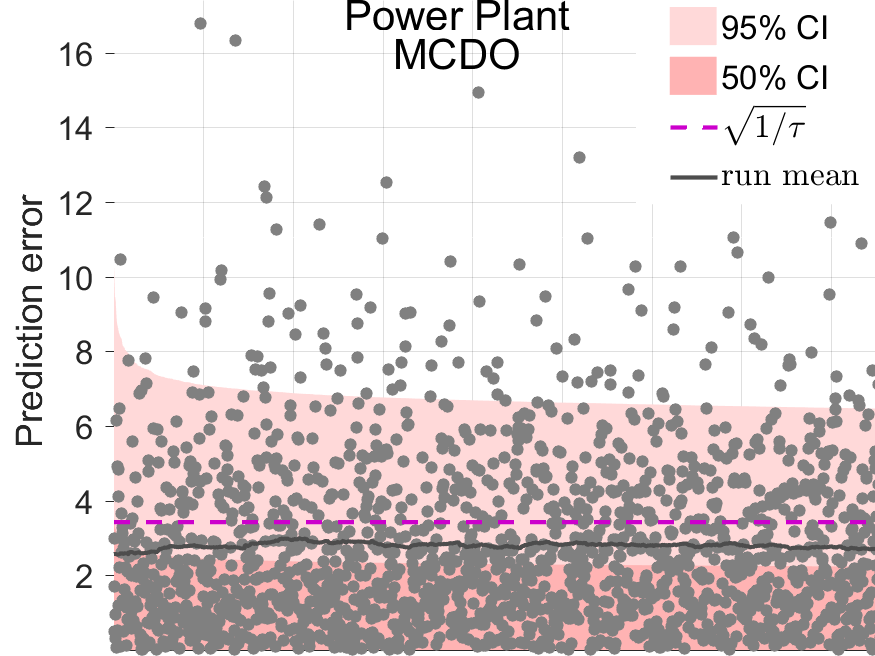} \\
		\vspace{2mm}
		\includegraphics[width=0.3\linewidth]{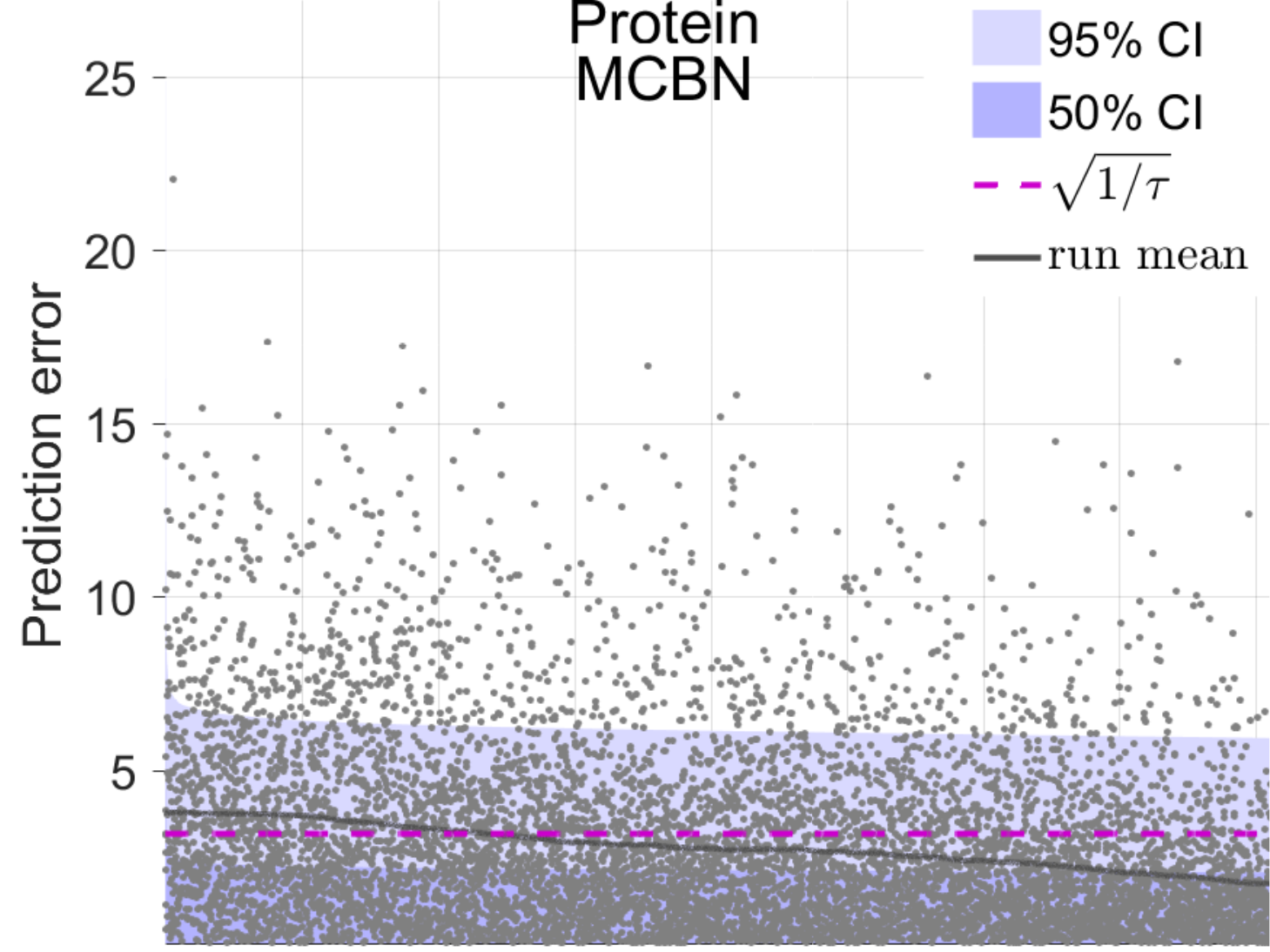} &
		\includegraphics[width=0.3\linewidth]{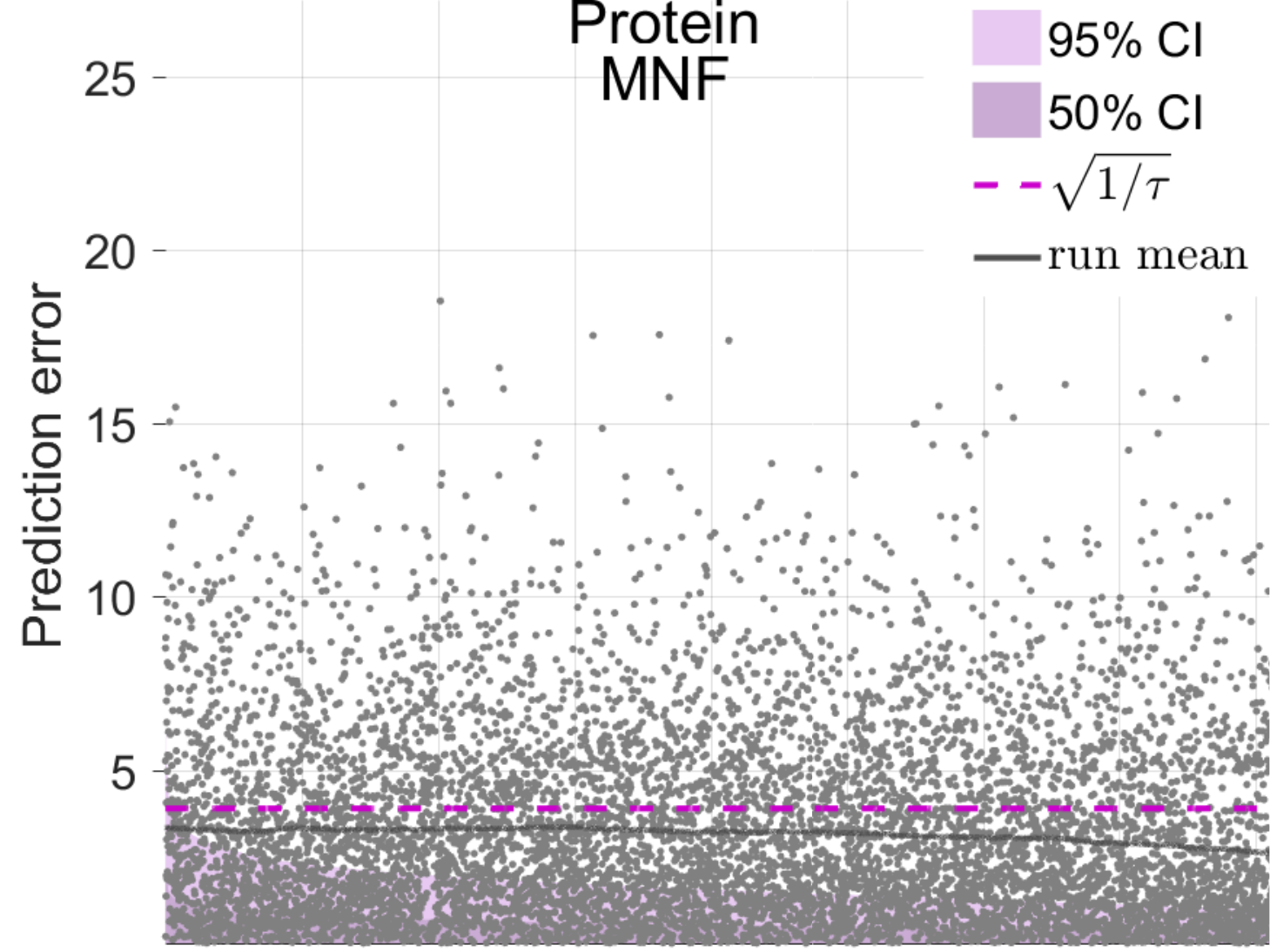} &
		\includegraphics[width=0.3\linewidth]{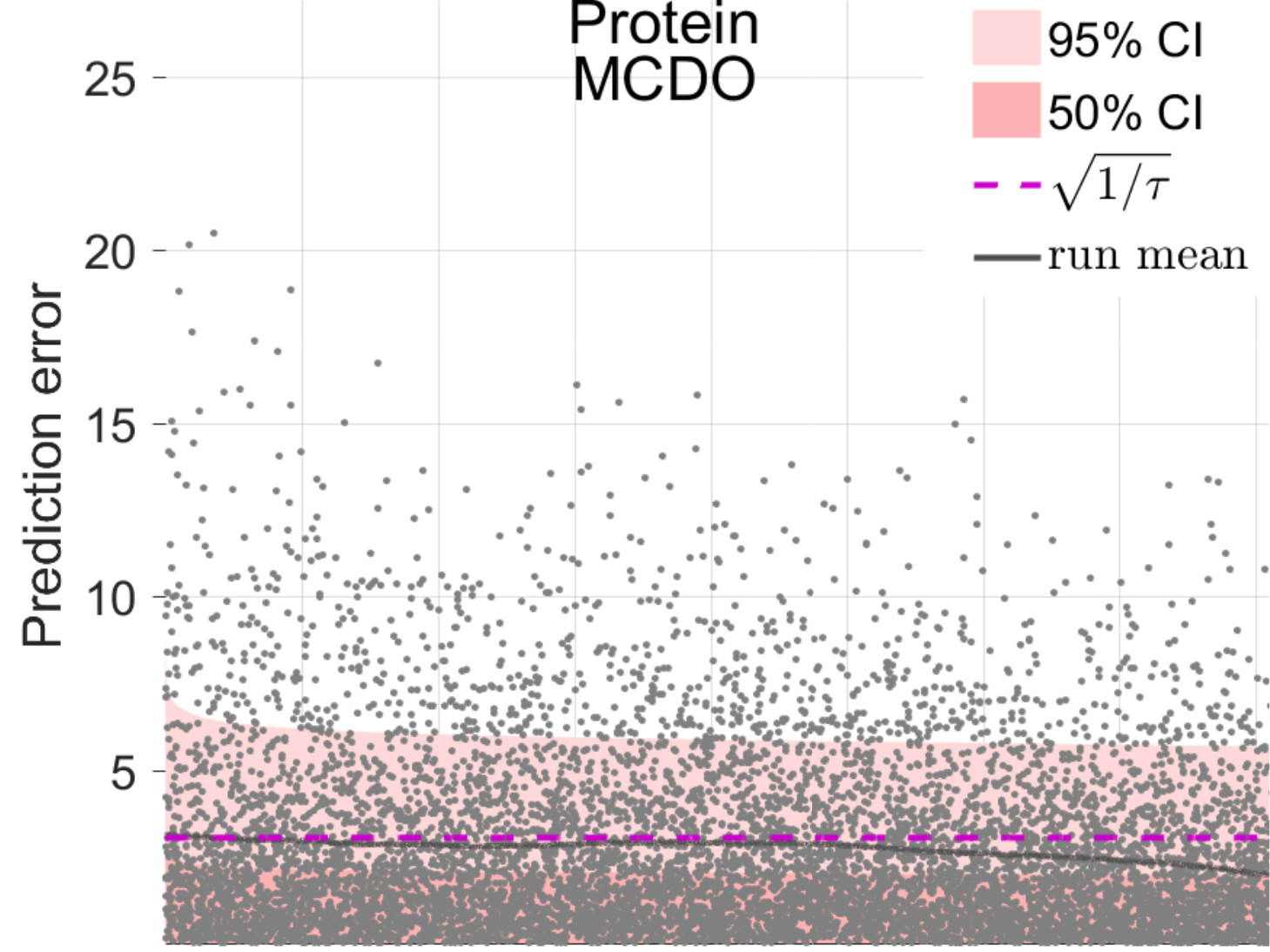} \\
		\vspace{2mm}
		\includegraphics[width=0.3\linewidth]{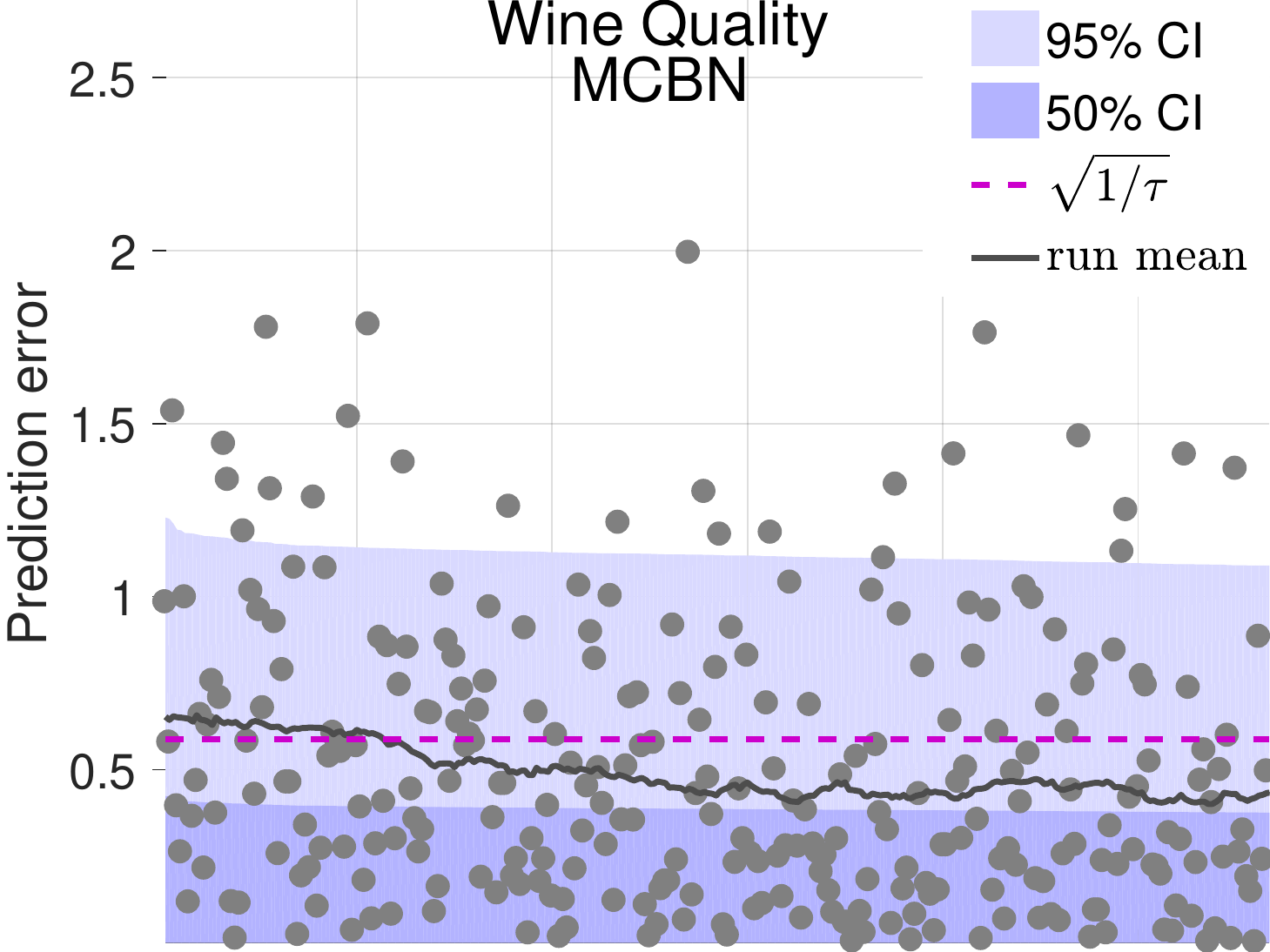} &
		\includegraphics[width=0.3\linewidth]{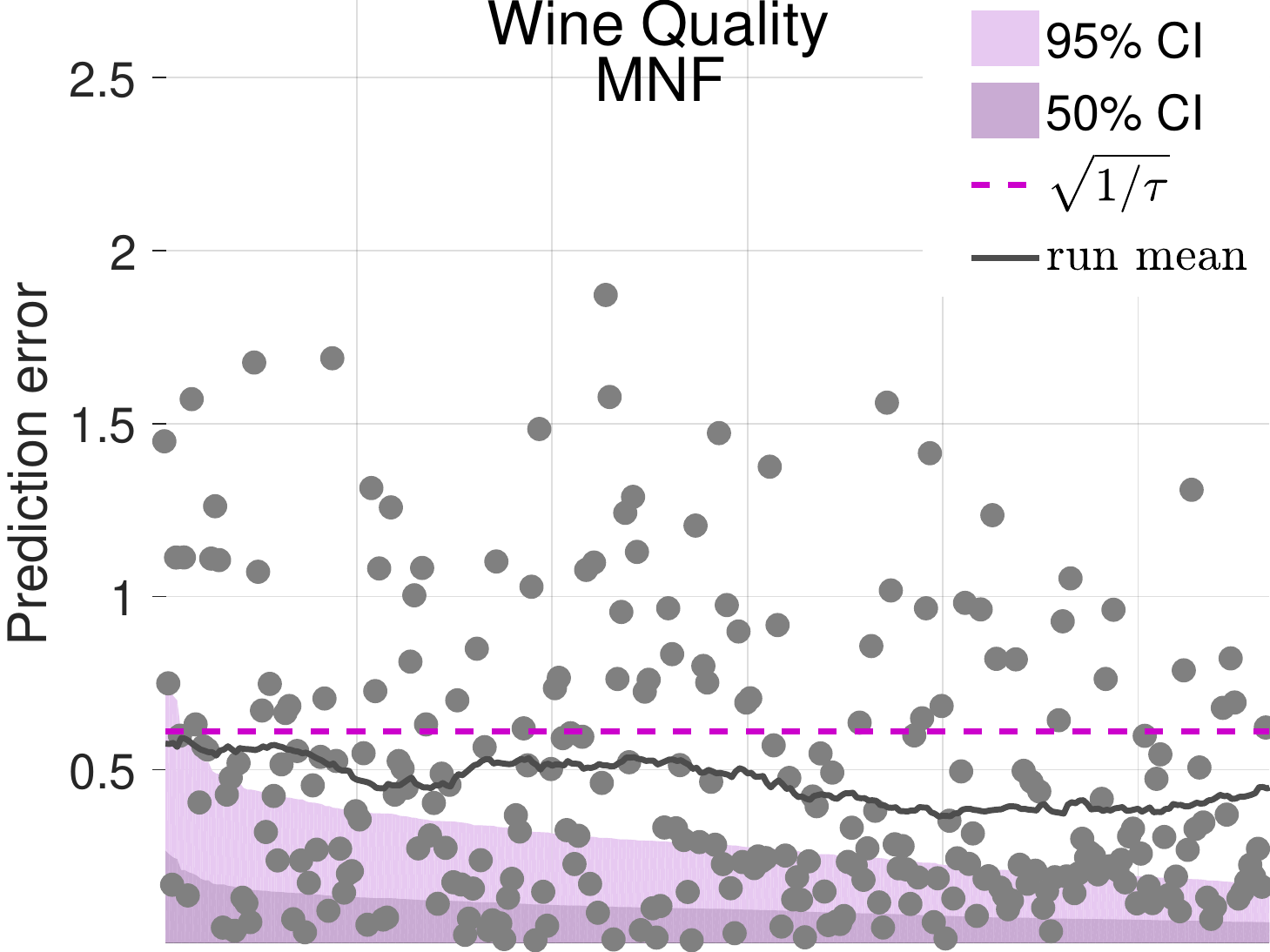} &
		\includegraphics[width=0.3\linewidth]{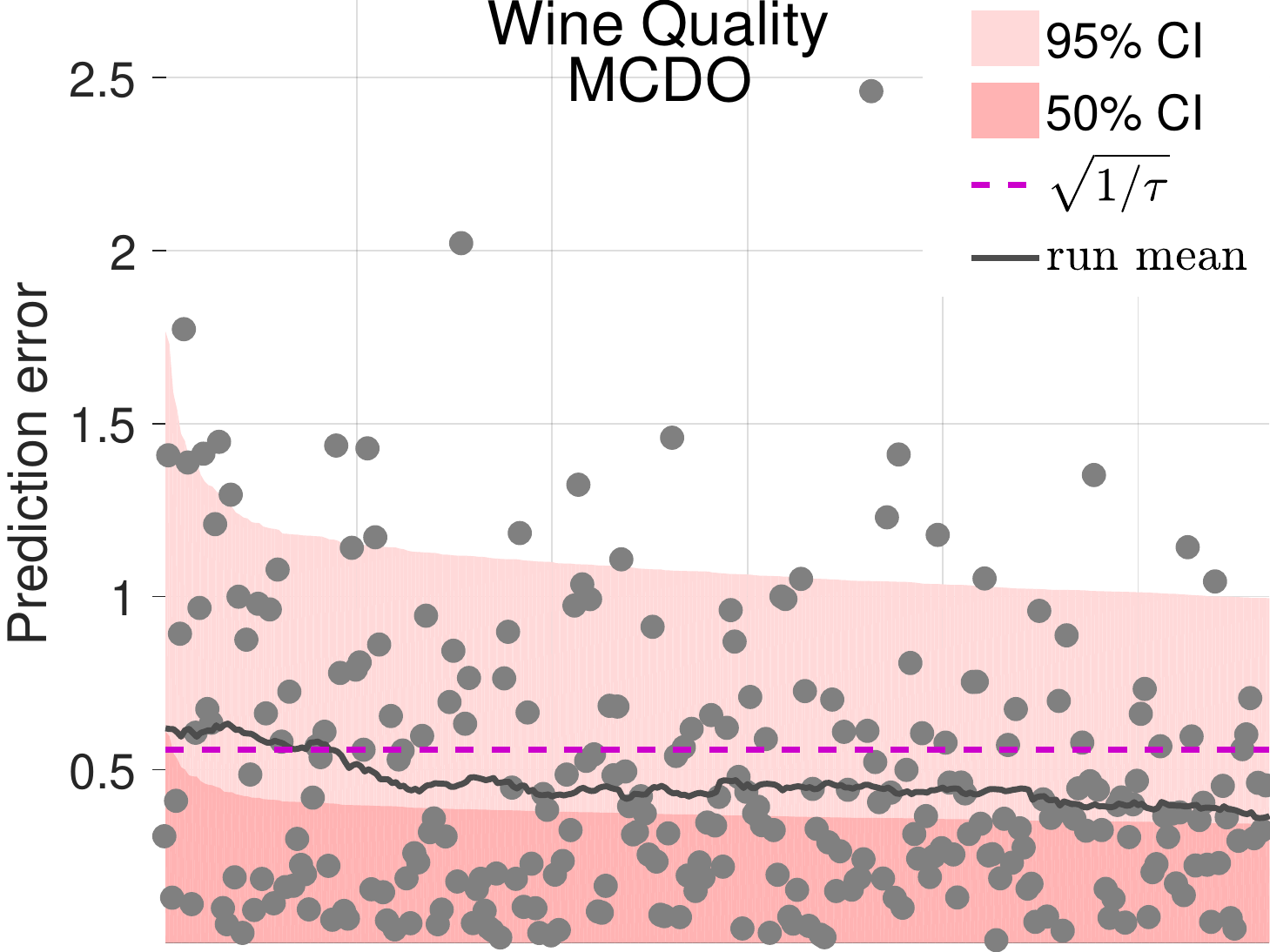} \\
		\vspace{2mm}
		\includegraphics[width=0.3\linewidth]{figures/r_yacht_MCBN} &
		\includegraphics[width=0.3\linewidth]{figures/r_yacht_MNF} &
		\includegraphics[width=0.3\linewidth]{figures/r_yacht_MCDO} \\
		
	\end{tabular}
	\vspace{-3mm}
	\caption{Errors in predictions (gray dots) sorted by estimated uncertainty on select datasets. The shaded areas show model uncertainty (light area 95\% CI, dark area 50\% CI). Gray dots show absolute prediction errors on the test set, and the gray line depicts a running mean of the errors. The dashed line indicates the optimized constant uncertainty. A correlation between estimated uncertainty (shaded area) and mean error (gray) indicates the uncertainty estimates are meaningful for estimating errors.}
	\label{fig:sortplotsA2}
	\vspace{-3mm}
\end{figure}

In Figure \ref{fig:sortplotsA1} and Figure \ref{fig:sortplotsA2}, we provide a full set of our uncertainty quality visualization plots, where errors in predictions are sorted by estimated uncertainty. The shaded areas show the model uncertainty and gray dots show absolute prediction errors on the test set. A gray line depicts a running mean of the errors. The dashed line indicates the optimized constant uncertainty. In these plots, we can see a correlation between estimated uncertainty (shaded area) and mean error (gray). This trend indicates that the model uncertainty estimates can recognize samples with larger (or smaller) potential for predictive errors.

We also conduct a sensitivity analysis to estimate how the uncertainty quality varies with batch size $M$ and the number of stochastic forward passes $T$. In tables \ref{table:batch_size_CRPS} and \ref{table:batch_size_PLL} we evaluate $\overline{\textrm{CRPS}}$ and $\overline{\textrm{PLL}}$ respectively for the regression datasets when trained and evaluated with varying batch sizes, but other hyperparameters fixed ($T$ was fixed at 100). The results show that results deteriorate when batch sizes are too small, likely stemming from the large variance of the approximate posterior. In tables \ref{table:forward_pass_CRPS} and \ref{table:forward_pass_PLL} we evaluate $\overline{\textrm{CRPS}}$ and $\overline{\textrm{PLL}}$ respectively for the regression datasets when trained and evaluated with varying n.o. stochastic forward samples, but other hyperparameters fixed ($M$ was fixed at 128). The results are indicative of performance improvements with larger $T$, although we see improvements over baseline for some datasets already with $T=50$ ($1/10$:th of the $T$ used in our main experiments).

\begin{table}[ht]
	\centering
	\caption{\textbf{Uncertainty quality measured by $\overline{\textrm{CRPS}}$ on regression dasets.} $\overline{\textrm{CRPS}}$ measured on eight datasets over 5 random 80-20 splits of the data with 5 different random seeds each split. Mean values for MCBN, MCDO and MNF are reported along with standard error. A significance test was performed to check if $\overline{\textrm{CRPS}}$ significantly exceeds the baseline. The $p$-value from a one sample t-test is reported.}
	\begin{tabular}{lllllll}
		\toprule
		& \multicolumn{6}{c}{$\overline{\textrm{CRPS}}$}                                                                                                                                                 \\
		Dataset             & \multicolumn{1}{c}{MCBN} & \multicolumn{1}{c}{$p$-value} & \multicolumn{1}{c}{MCDO} & \multicolumn{1}{c}{$p$-value} & \multicolumn{1}{c}{MNF} & \multicolumn{1}{c}{$p$-value}\\ 
		\midrule
		Boston Housing      & 8.50$\pm$0.86            & 6.39E-10                    & 3.06$\pm$0.33            & 1.64E-09                    & 5.88$\pm$1.09           & 2.01E-05                    \\
		Concrete            & 3.91$\pm$0.25            & 4.53E-14                    & 0.93$\pm$0.41            & 3.13E-02                    & 3.13$\pm$0.81           & 6.43E-04                    \\
		Energy Efficiency   & 5.75$\pm$0.52            & 6.71E-11                    & 1.37$\pm$0.89            & 1.38E-01                    & 1.10$\pm$2.63           & 6.45E-01                    \\
		Kinematics 8nm      & 2.85$\pm$0.18            & 2.33E-14                    & 1.82$\pm$0.14            & 1.64E-12                    & 0.52$\pm$0.26           & 7.15E-02                    \\
		Power Plant         & 0.24$\pm$0.05            & 2.32E-04                    & -0.44$\pm$0.05           & 2.17E-08                    & -0.89$\pm$0.15          & 3.36E-06                    \\
		Protein             & 2.66$\pm$0.10            & 2.77-12                     & 0.99$\pm$0.08            & 2.34E-12                    & 0.57$\pm$0.03           & 8.56E-16                    \\
		Wine Quality (Red)  & 0.26$\pm$0.07            & 1.26E-03                    & 2.00$\pm$0.21            & 1.83E-09                    & 0.93$\pm$0.12           & 6.19E-08                    \\
		Yacht Hydrodynamics & -56.39$\pm$14.27         & 5.94E-04                    & 21.42$\pm$2.99           & 2.16E-07                    & 24.92$\pm$3.77          & 9.62E-06                    \\
		\bottomrule
	\end{tabular}
	\label{table:crps_results}
\end{table}

\begin{table}[ht]
	\centering
	\caption{\textbf{Uncertainty quality measured by $\overline{\textrm{PLL}}$ on regression dasets.} $\overline{\textrm{PLL}}$ measured on eight datasets over 5 random 80-20 splits of the data with 5 different random seeds each split. Mean values for MCBN, MCDO and MNF are reported along with standard error. A significance test was performed to check if $\overline{\textrm{PLL}}$ significantly exceeds the baseline. The $p$-value from a one sample t-test is reported.}
	\begin{tabular}{lllllll}
		\toprule
		& \multicolumn{6}{c}{$\overline{\textrm{PLL}}$}                                                                                                                                                 \\
		Dataset             & \multicolumn{1}{c}{MCBN} & \multicolumn{1}{c}{$p$-value} & \multicolumn{1}{c}{MCDO} & \multicolumn{1}{c}{$p$-value} & \multicolumn{1}{c}{MNF} & \multicolumn{1}{c}{$p$-value}\\ 
		\midrule
		Boston Housing      & 10.49$\pm$1.35           & 5.41E-08                    & 5.51$\pm$1.05            & 2.20E-05                    & 1.76$\pm$1.12           & 1.70E-01                    \\
		Concrete            & -36.36$\pm$12.12         & 6.19E-03                    & 10.92$\pm$1.78           & 2.34E-06                    & -2.16$\pm$4.19          & 6.79E-01                    \\
		Energy Efficiency   & 10.89$\pm$1.16           & 1.79E-09                    & -14.28$\pm$5.15          & 1.06E-02                    & -33.88$\pm$29.57        & 2.70E-01                    \\
		Kinematics 8nm      & 1.68$\pm$0.37            & 1.29E-04                    & -0.26$\pm$0.18           & 1.53E-01                    & 0.42$\pm$0.43           & 2.70E-01                    \\
		Power Plant         & 0.33$\pm$0.14            & 2.72E-02                    & 3.52$\pm$0.23            & 1.12E-13                    & -0.86$\pm$0.15          & 7.33E-06                    \\
		Protein             & 2.56$\pm$0.23            & 4.28E-11                    & 6.23$\pm$0.19            & 2.57E-21                    & 0.52$\pm$0.07           & 1.81E-07                    \\
		Wine Quality (Red)  & 0.19$\pm$0.09            & 3.72E-02                    & 2.91$\pm$0.35            & 1.84E-08                    & 0.83$\pm$0.16           & 2.27E-05                    \\
		Yacht Hydrodynamics & 45.58$\pm$5.18           & 5.67E-09                    & -41.54$\pm$31.37         & 1.97E-01                    & 46.19$\pm$4.45          & 2.47E-07 \\
		\bottomrule
	\end{tabular}
	\label{table:pll_results}
\end{table}

\begin{table}[ht]
	\centering
	\caption{\textbf{Raw (unnormalized) CRPS and PLL scores on regression datasets.} CRPS and PLL measured on eight datasets over 5 random 80-20 splits of the data with 5 different random seeds each split. Mean values and standard errors are reported for MCBN, MCDO and MNF.}
	\begin{tabular}{lllllll}
		\toprule
		& \multicolumn{3}{c}{CRPS}                                                      & \multicolumn{3}{c}{PLL}                                    \\
		Dataset             & \multicolumn{1}{c}{MCBN} & \multicolumn{1}{c}{MCDO} & \multicolumn{1}{c}{MNF} & \multicolumn{1}{c}{MCBN} & MCDO           & MNF            \\ \midrule
		Boston Housing      & 1.45$\pm$0.02            & 1.41$\pm$0.02            & 1.57$\pm$0.02           & -2.38$\pm$0.02           & -2.35$\pm$0.02 & -2.51$\pm$0.06 \\
		Concrete            & 2.40$\pm$0.04            & 2.42$\pm$0.04            & 3.61$\pm$0.02           & -3.45$\pm$0.11           & -2.94$\pm$0.02 & -3.35$\pm$0.04 \\
		Energy Efficiency   & 0.33$\pm$0.01            & 0.26$\pm$0.00            & 1.33$\pm$0.04           & -0.94$\pm$0.04           & -0.80$\pm$0.04 & -3.18$\pm$0.07 \\
		Kinematics 8nm      & 0.04$\pm$0.00            & 0.04$\pm$0.00            & 0.05$\pm$0.00           & 1.21$\pm$0.01            & 1.24$\pm$0.00  & 1.04$\pm$0.00  \\
		Power Plant         & 2.00$\pm$0.01            & 2.00$\pm$0.01            & 2.31$\pm$0.01           & -2.75$\pm$0.00           & -2.72$\pm$0.01 & -2.86$\pm$0.01 \\
		Protein             & 1.95$\pm$0.01            & 1.95$\pm$0.00            & 2.25$\pm$0.01           & -2.73$\pm$0.00           & -2.70$\pm$0.00 & -2.83$\pm$0.01 \\
		Wine Quality (Red)  & 0.34$\pm$0.00            & 0.33$\pm$0.00            & 0.34$\pm$0.00           & -0.95$\pm$0.01           & -0.89$\pm$0.01 & -0.93$\pm$0.00 \\
		Yacht Hydrodynamics & 0.68$\pm$0.02            & 0.32$\pm$0.01            & 0.94$\pm$0.01           & -1.39$\pm$0.03           & -2.57$\pm$0.69 & -1.96$\pm$0.05 \\ \bottomrule
	\end{tabular}
	\label{table:rawcrpspll}
\end{table}

\begin{table}[H]
	\centering
	\caption{\textbf{Prediction accuracy measured by RMSE on regression datasets.} RMSE measured on eight datasets over 5 random 80-20 splits of the data with 5 different random seeds each split. Mean values and standard errors are reported for for MCBN, MCDO and MNF as well as conventional non-Bayesian models BN and DO.}
	\begin{tabular}{llllll}
		\toprule
		& \multicolumn{5}{c}{RMSE}                                                                                              \\
		Dataset             & \multicolumn{1}{c}{MCBN} & \multicolumn{1}{c}{BN} & \multicolumn{1}{c}{MCDO} & \multicolumn{1}{c}{DO} & MNF           \\ \midrule
		Boston Housing      & 2.75$\pm$0.05            & 2.77$\pm$0.05          & 2.65$\pm$0.05            & 2.69$\pm$0.05          & 2.98$\pm$0.06 \\
		Concrete            & 4.78$\pm$0.09            & 4.89$\pm$0.08          & 4.80$\pm$0.10            & 4.99$\pm$0.10          & 6.57$\pm$0.04 \\
		Energy Efficiency   & 0.59$\pm$0.02            & 0.57$\pm$0.01          & 0.47$\pm$0.01            & 0.49$\pm$0.01          & 2.38$\pm$0.07 \\
		Kinematics 8nm      & 0.07$\pm$0.00            & 0.07$\pm$0.00          & 0.07$\pm$0.00            & 0.07$\pm$0.00          & 0.09$\pm$0.00 \\
		Power Plant         & 3.74$\pm$0.01            & 3.74$\pm$0.01          & 3.74$\pm$0.02            & 3.72$\pm$0.02          & 4.19$\pm$0.01 \\
		Protein             & 3.66$\pm$0.01            & 3.69$\pm$0.01          & 3.66$\pm$0.01            & 3.68$\pm$0.01          & 4.10$\pm$0.01 \\
		Wine Quality (Red)  & 0.62$\pm$0.00            & 0.62$\pm$0.00          & 0.60$\pm$0.00            & 0.61$\pm$0.00          & 0.61$\pm$0.00 \\
		Yacht Hydrodynamics & 1.23$\pm$0.05            & 1.28$\pm$0.06          & 0.75$\pm$0.03            & 0.72$\pm$0.04          & 2.13$\pm$0.05 \\ \bottomrule
	\end{tabular}
	\label{table:rmse_results}
\end{table}

\begin{table}[H]
	\centering
	\caption{\textbf{Uncertainty quality sensitivity to batch size.} A sensitivity analysis to determine how MCBN uncertainty quality varies with batch size is measured on eight regression datasets using $\overline{\textrm{CRPS}}$ as the quality measure. Results are measured over 3 random 80-20 splits of the data with 5 different random seeds each split.}
	\begin{tabular}{lcccccccc}
		\toprule
		& \multicolumn{8}{c}{$\overline{\textrm{CRPS}}$} \\
		Batch size&8&16&32&64&128&256&512&1024\\ 
		\midrule
		Boston Housing&-7.1&\textbf{16.6}&11.8&7.2&2.5&0.9&-&-\\
		Concrete&-34.5&\textbf{6.0}&5.0&5.1&2.9&1.4&0.6&0.0\\
		Energy Efficiency&-61.6&-3.0&2.7&9.8&\textbf{11.1}&0.8&4.9&-\\
		Kinematics 8nm&-1.4&-4.3&0.2&2.8&\textbf{2.7}&1.7&0.9&0.5\\
		Power Plant&-10.5&0.8&0.0&-0.1&0.0&0.0&\textbf{0.2}&0.0\\
		Protein&\textbf{14.5}&4.8&3.6&2.8&2.5&1.6&1.0&0.5\\
		Wine Quality (Red)&\textbf{2.2}&1.6&0.6&0.6&0.3&0.0&0.2&0.0\\
		Yacht Hydrodynamics&15.1&-23.0&-30.4&21.0&\textbf{34.4}&-&-&-\\
		\bottomrule
	\end{tabular}
	\label{table:batch_size_CRPS}
\end{table}

\begin{table}[h!]
	\centering
	\caption{\textbf{Uncertainty quality sensitivity to batch size.} A sensitivity analysis to determine how MCBN uncertainty quality varies with batch size is measured on eight regression datasets using $\overline{\textrm{PLL}}$ as the quality measure. Results are measured over 3 random 80-20 splits of the data with 5 different random seeds each split.}
	\begin{tabular}{lcccccccc}
		\toprule
		& \multicolumn{8}{c}{$\overline{\textrm{PLL}}$} \\
		Batch size&8&16&32&64&128&256&512&1024\\ 
		\midrule
		Boston Housing&\textbf{13.9}&-36.7&10.0&7.9&3.7&1.5&-&-\\
		Concrete&-113.3&-528.4&-10.0&\textbf{2.9}&0.0&1.4&0.2&0.0\\
		Energy Efficiency&-64.4&5.2&-0.2&-9.6&-14.5&1.4&\textbf{10.4}&-\\
		Kinematics 8nm&-4.9&-5.4&-3.1&1.6&\textbf{2.3}&1.5&0.7&0.4\\
		Power Plant&-135.0&-1.4&-1.0&-1.1&-0.4&0.1&-0.1&\textbf{0.4}\\
		Protein&\textbf{44.9}&15.7&4.6&2.9&2.8&2.2&1.2&0.6\\
		Wine Quality (Red)&\textbf{2.2}&2.0&0.0&0.5&0.6&0.4&0.0&0.0\\
		Yacht Hydrodynamics&\textbf{99.6}&74.9&76.8&48.5&44.9&-&-&-\\
		\bottomrule
	\end{tabular}
	\label{table:batch_size_PLL}
\end{table}

\begin{table}[H]
	\centering
	\caption{\textbf{Uncertainty quality sensitivity to n.o. stochastic forward passes.} A sensitivity analysis to determine how MCBN uncertainty quality varies with the n.o. stochastic forward passes measured on eight regression datasets using $\overline{\textrm{CRPS}}$ as the quality measure. Results are measured over 3 random 80-20 splits of the data with 5 different random seeds each split.}
	\begin{tabular}{lccc}
		\toprule
		& \multicolumn{3}{c}{$\overline{\textrm{CRPS}}$} \\
		Forward passes&250&100&50\\ 
		\midrule
		Boston Housing&\textbf{6.1}&2.7&3.2\\
		Concrete&3.3&2.3&\textbf{3.3}\\
		Energy Efficiency&\textbf{13.2}&4.2&7.9\\
		Kinematics 8nm&3.2&2.7&\textbf{4.2}\\
		Power Plant&0.2&\textbf{0.5}&0.1\\
		Protein&2.3&\textbf{2.7}&2.4\\
		Wine Quality (Red)&\textbf{0.9}&-0.4&0.6\\
		Yacht Hydrodynamics&\textbf{32.9}&32.2&32.1\\
		\bottomrule
	\end{tabular}
	\label{table:forward_pass_CRPS}
\end{table}

\begin{table}[H]
	\centering
	\caption{\textbf{Uncertainty quality sensitivity to n.o. stochastic forward passes.} A sensitivity analysis to determine how MCBN uncertainty quality varies with the n.o. stochastic forward passes measured on eight regression datasets using $\overline{\textrm{PLL}}$ as the quality measure. Results are measured over 3 random 80-20 splits of the data with 5 different random seeds each split.}
	\begin{tabular}{lccc}
		\toprule
		& \multicolumn{3}{c}{$\overline{\textrm{PLL}}$} \\
		Forward passes&250&100&50\\ 
		\midrule
		Boston Housing&\textbf{7.8}&1.9&2.6\\
		Concrete&3.8&\textbf{7.1}&0.1\\
		Energy Efficiency&\textbf{15.7}&-30.5&-47.3\\
		Kinematics 8nm&2.5&2.2&\textbf{3.4}\\
		Power Plant&-0.9&\textbf{0.7}&-0.9\\
		Protein&1.8&2.0&\textbf{2.4}\\
		Wine Quality (Red)&\textbf{1.7}&-0.9&1.1\\
		Yacht Hydrodynamics&\textbf{38.0}&35.9&35.5\\
		\bottomrule
	\end{tabular}
	\label{table:forward_pass_PLL}
\end{table}

\subsection{Uncertainty in image segmentation}\label{appendix:segmentation}

\begin{figure}[t]
	\centering
	\begin{tabular}{@{}c@{\hskip 1mm}c@{\hskip 1mm}c@{}}
		\includegraphics[width=0.3\linewidth]{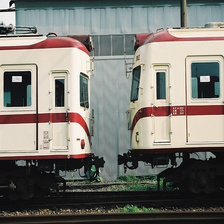} &
		\includegraphics[width=0.3\linewidth]{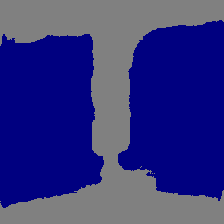} &
		\includegraphics[width=0.3\linewidth]{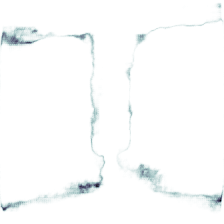} \\
		
		\includegraphics[width=0.3\linewidth]{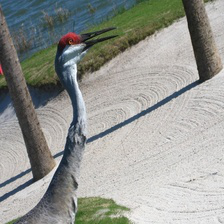} &
		\includegraphics[width=0.3\linewidth]{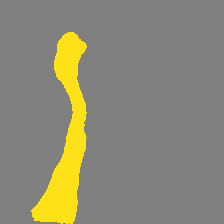} &
		\includegraphics[width=0.3\linewidth]{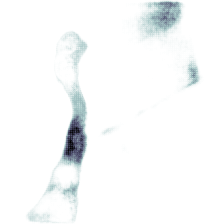} \\
		
		\includegraphics[width=0.3\linewidth]{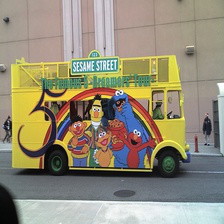} &
		\includegraphics[width=0.3\linewidth]{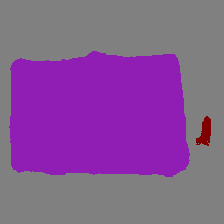} &
		\includegraphics[width=0.3\linewidth]{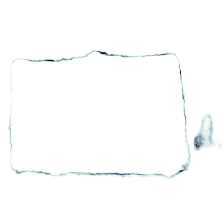} \\
		
		\includegraphics[width=0.3\linewidth]{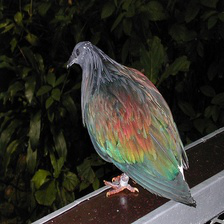} &
		\includegraphics[width=0.3\linewidth]{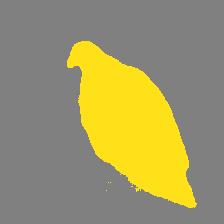} &
		\includegraphics[width=0.3\linewidth]{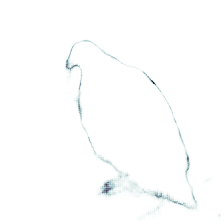}
		
	\end{tabular}
	\vspace{-3mm}
	\caption{\textbf{Uncertainty in image segmentation}. Results applying MCBN to Bayesian SegNet \citep{KendallBC15} on images from PASCAL-VOC (right). Left: original. Middle: the Bayesian estimated segmentation. Right: estimated uncertainty using MCBN for all classes. Mini-batches of size 36 were used for PASCAL-VOC on images of size 224x224. 20 inferences were conducted to estimate the mean and variance of MCBN.}
	\label{fig:segresults}
	\vspace{-3mm}
\end{figure}

We applied MCBN to an image segmentation task using Bayesian SegNet with the main CamVid and PASCAL-VOC models in \cite{KendallBC15}. Here, we provide more image from Pascal VOC dataset in Figure \ref{fig:segresults}.

\subsection{Batch normalization statistics}\label{appendix:extendedstatistics}

\begin{figure}[t]
	\centering
	\begin{tabular}{@{}c@{\hskip 1mm}c@{}}
		\includegraphics[width=0.35\linewidth]{figures/r-batchmean-l1_bs32_epoch=10} &
		\includegraphics[width=0.35\linewidth]{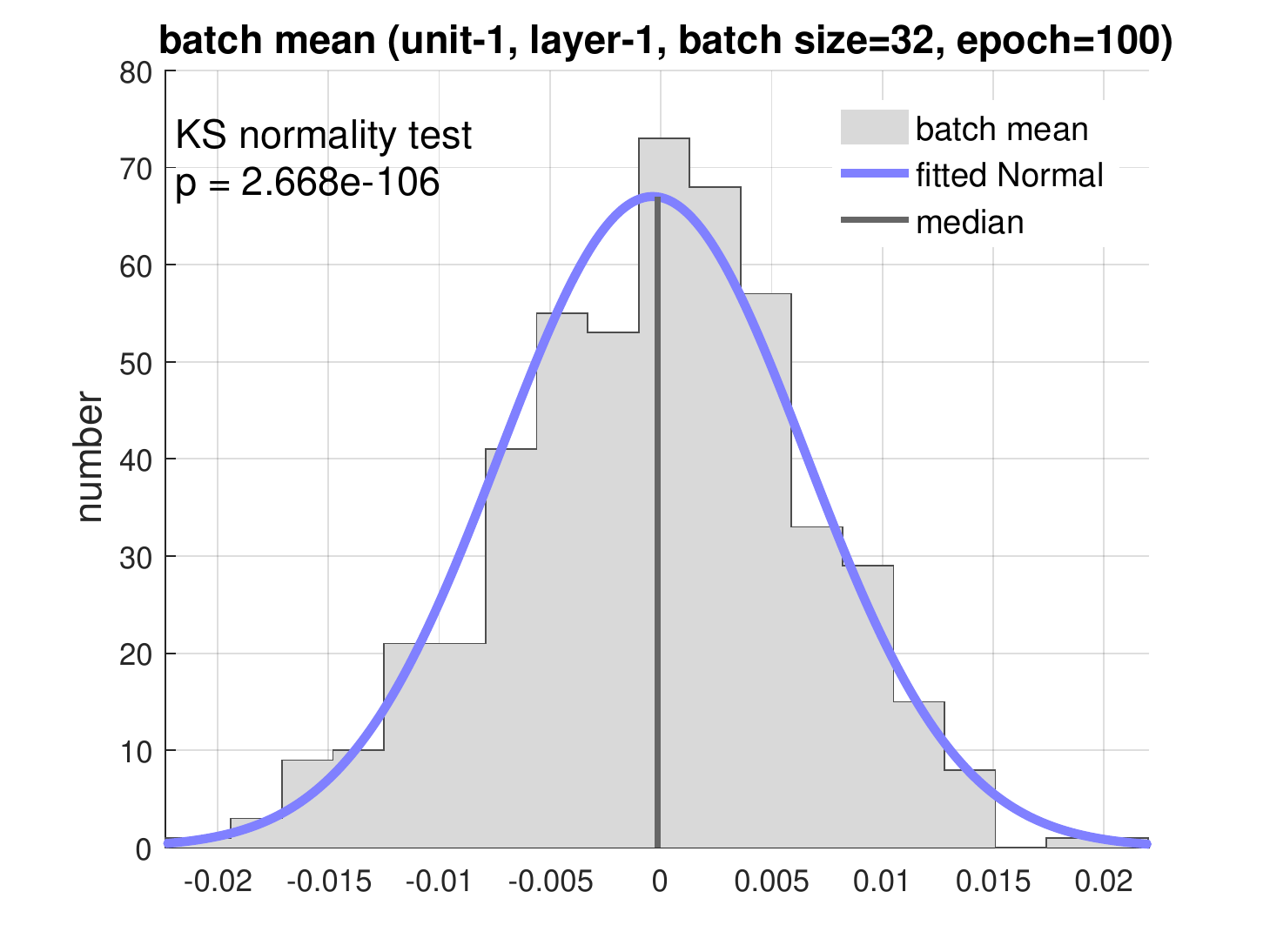} \\
		\includegraphics[width=0.35\linewidth]{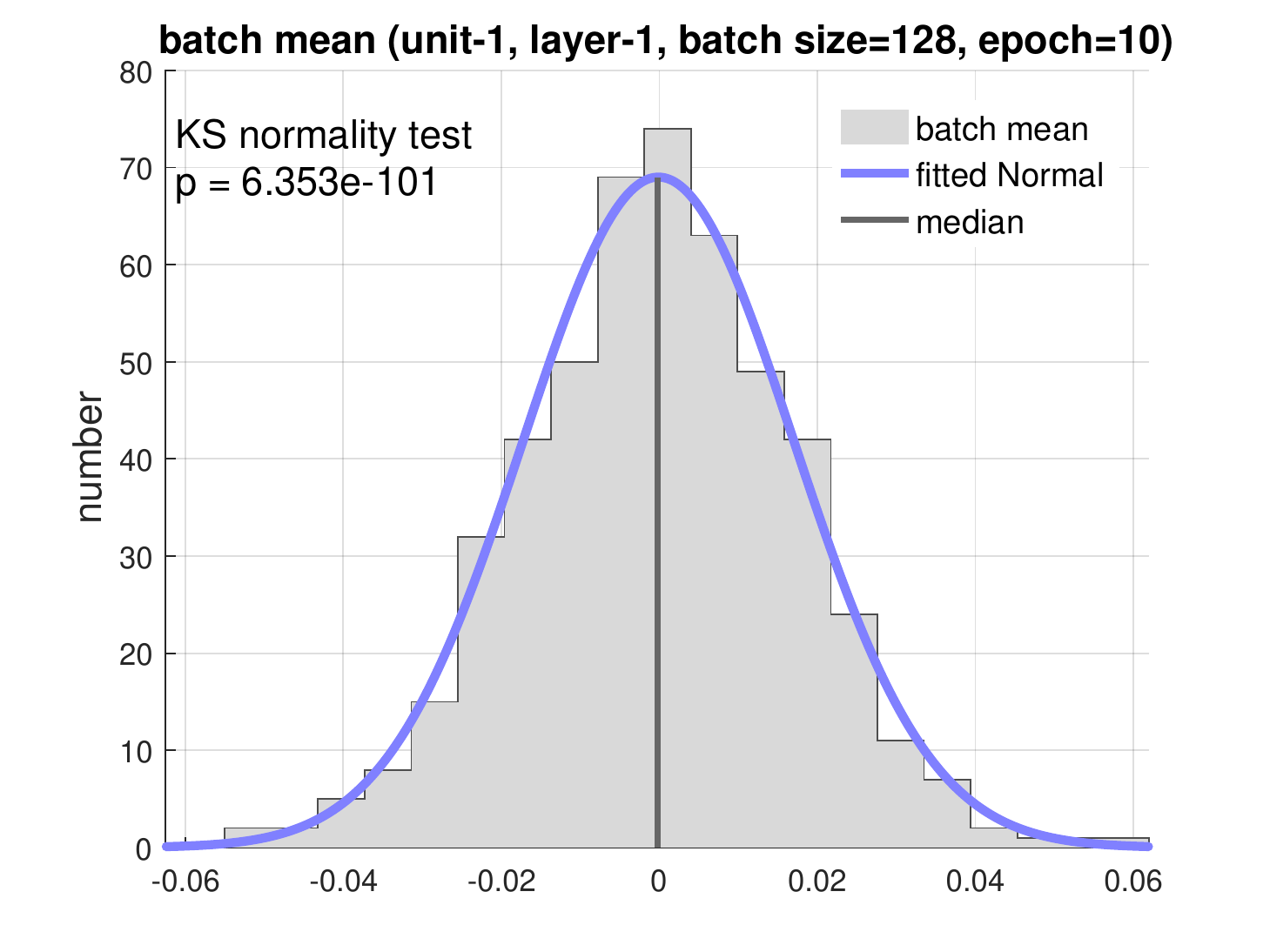} &
		\includegraphics[width=0.35\linewidth]{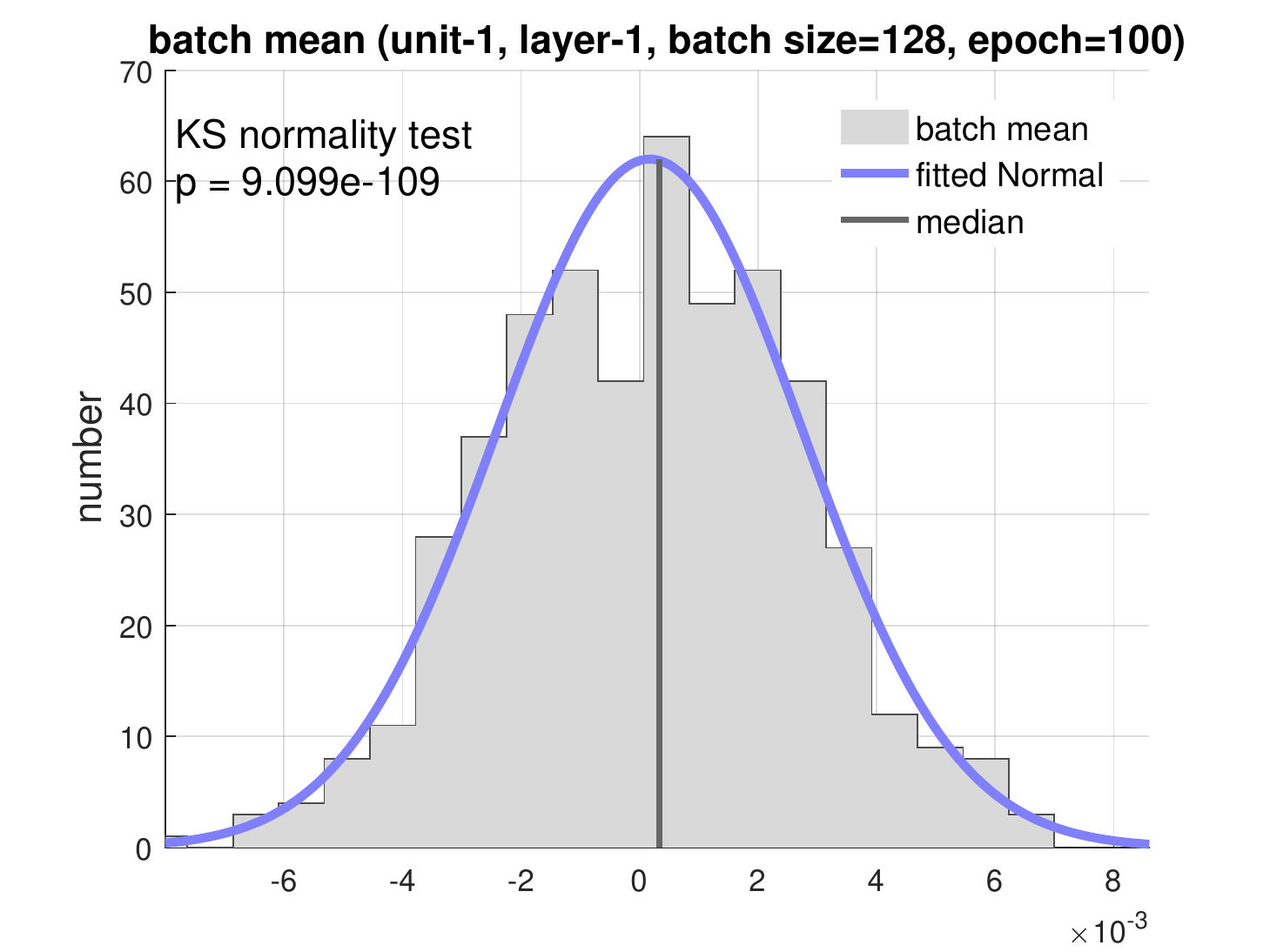} \\
		\vspace{2mm}
		\includegraphics[width=0.35\linewidth]{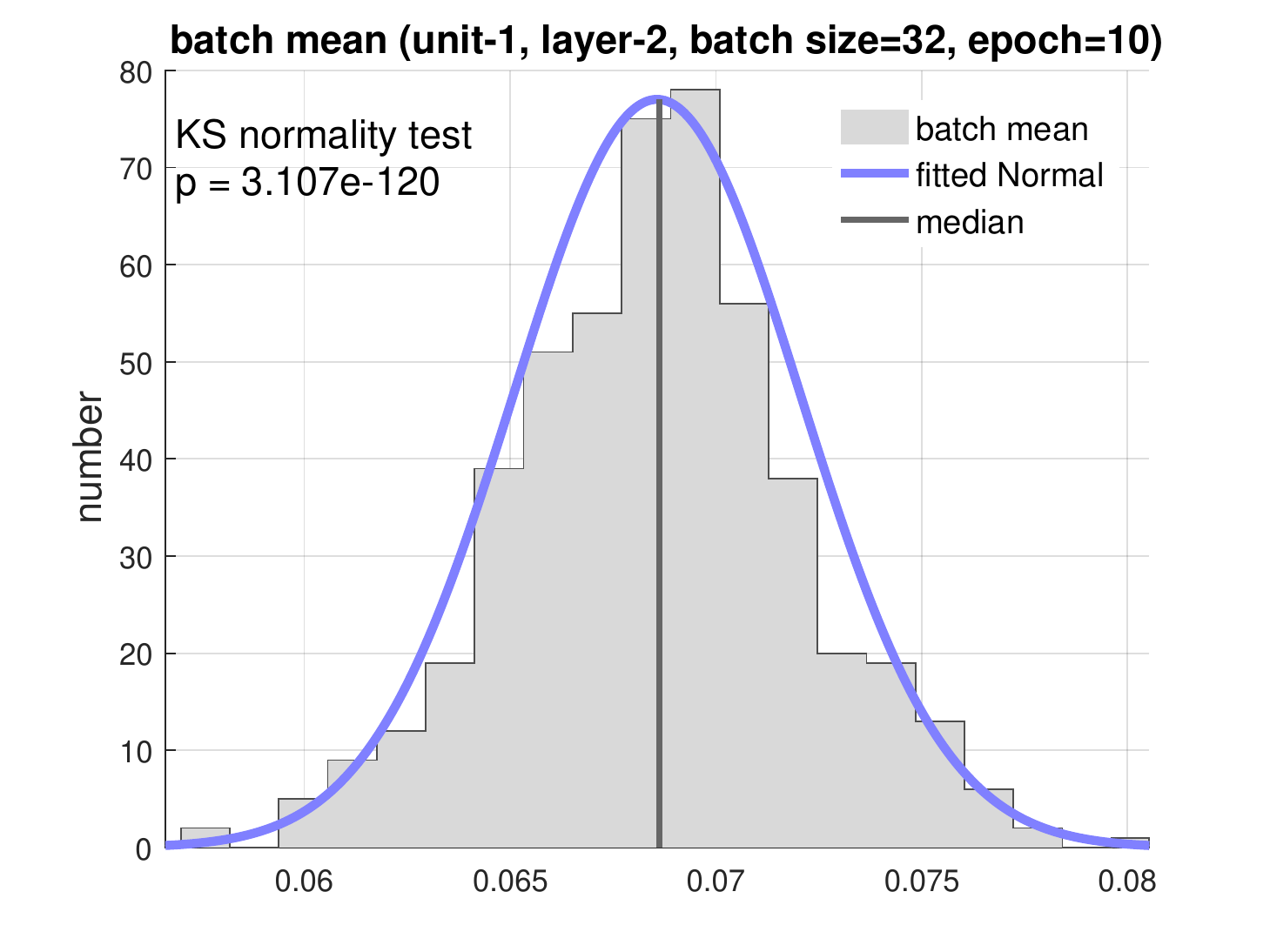} &
		\includegraphics[width=0.35\linewidth]{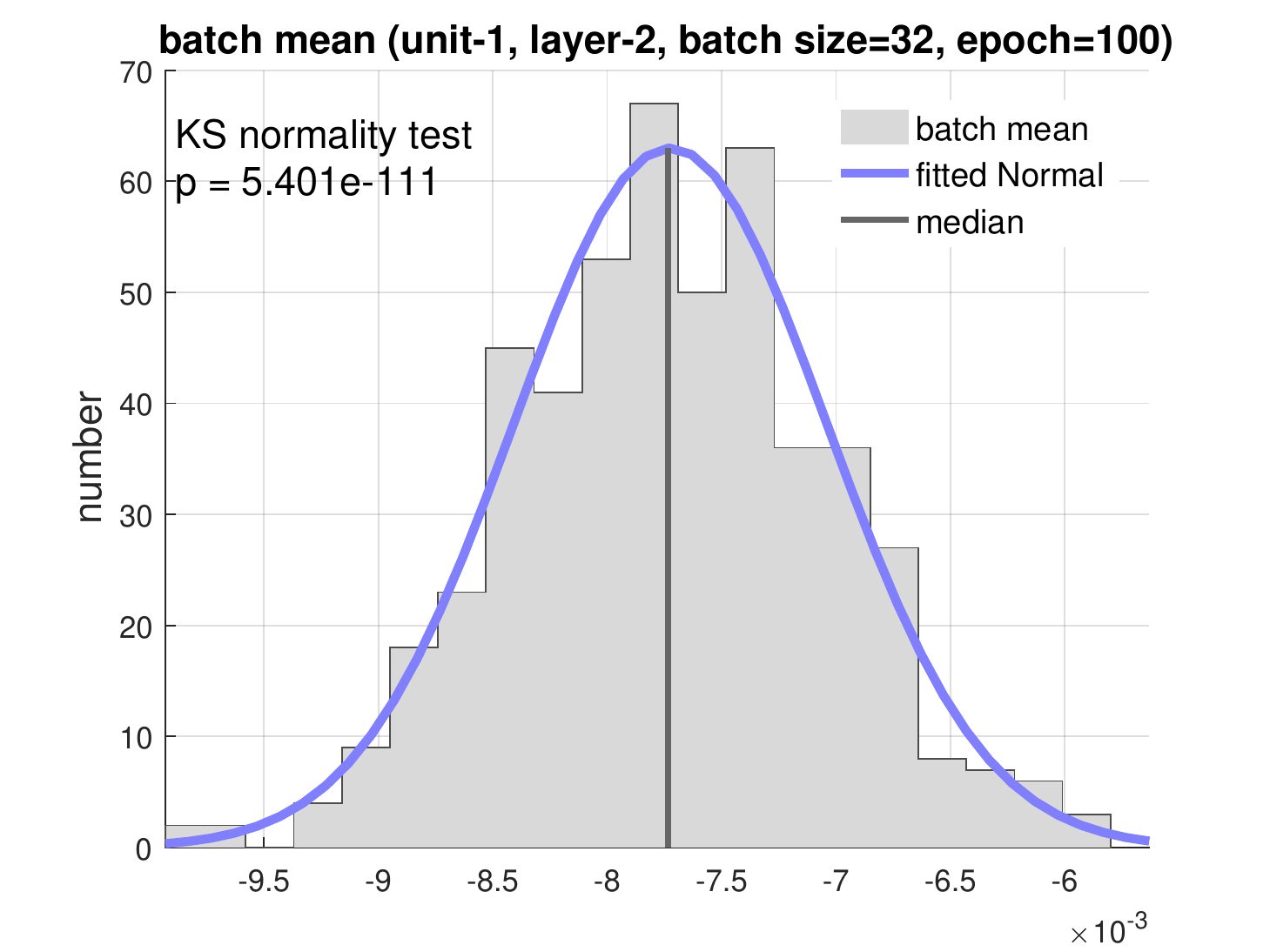} \\
		\includegraphics[width=0.35\linewidth]{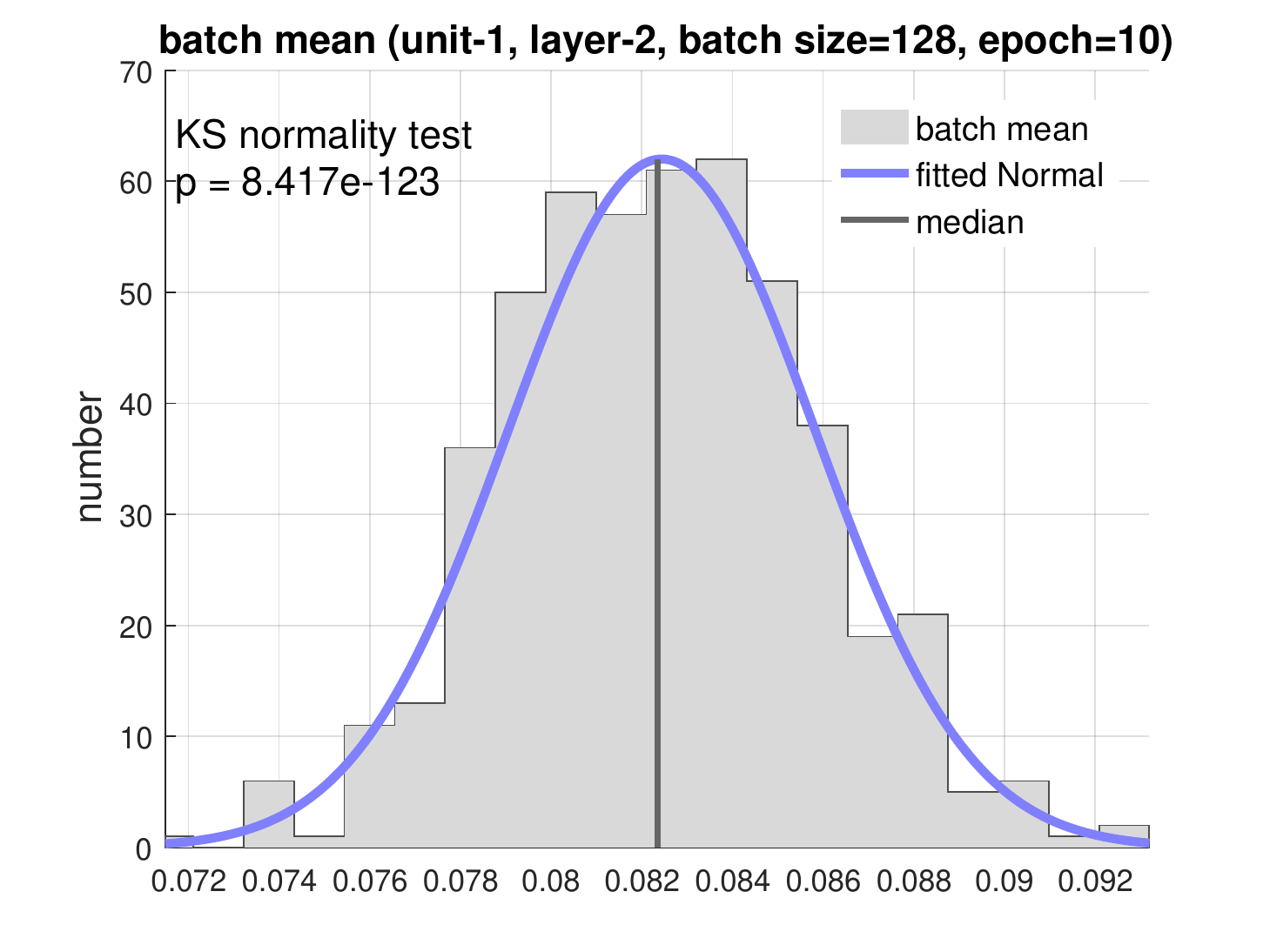} &
		\includegraphics[width=0.35\linewidth]{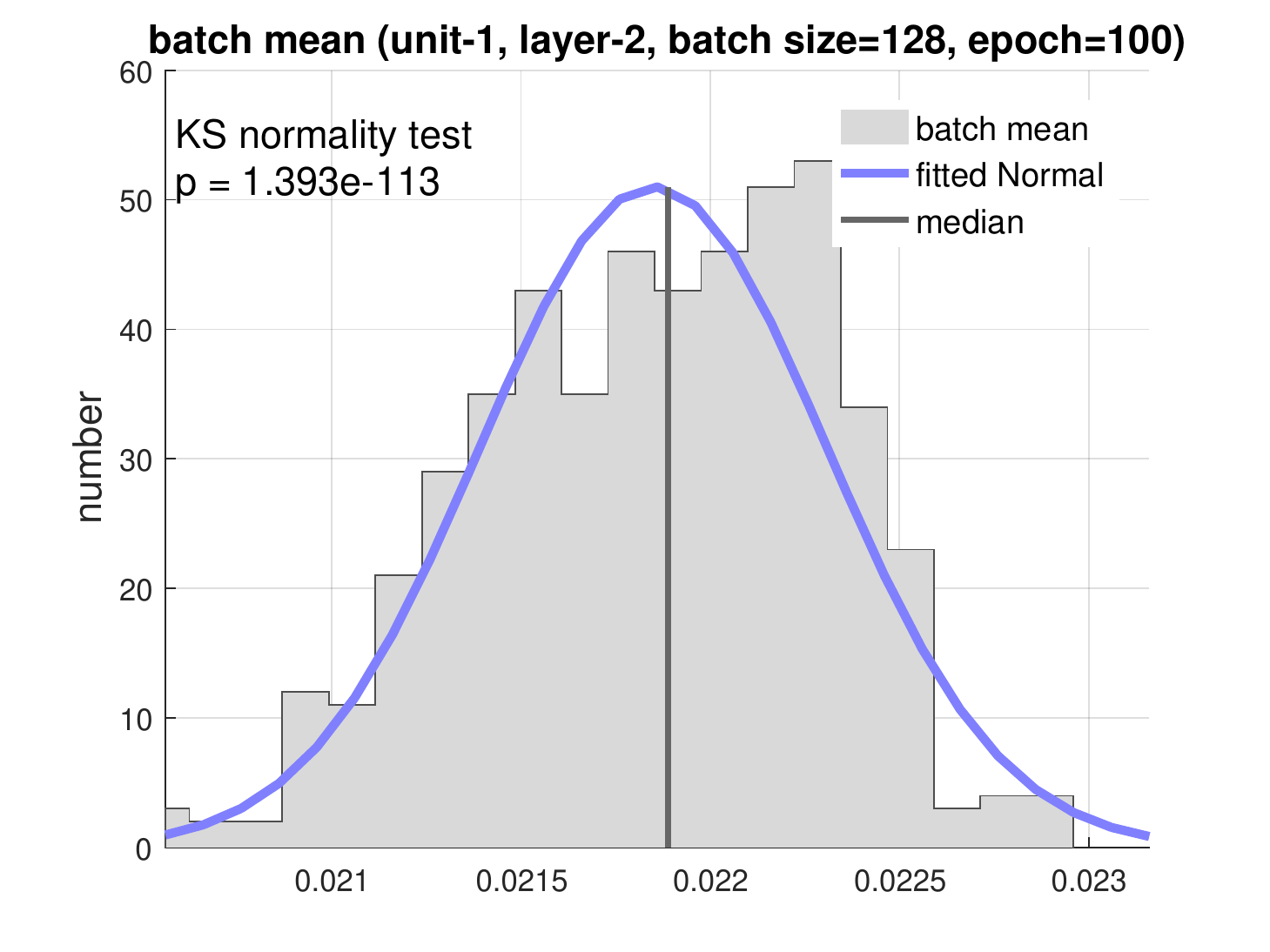} \\
	\end{tabular}
	\vspace{-3mm}
	\caption{\textbf{The distribution of means} of mini-batches during training of one of our datasets. The distribution closely follows our analytically approximated Gaussian distribution. The data is collected for one unit of each layer and is provided for different epochs and for different batch sizes.}
	\label{fig:batchmeans}
	\vspace{-3mm}
\end{figure}

\begin{figure}[t]
	\centering
	\vspace{-7mm}
	\begin{tabular}{@{}c@{\hskip 1mm}c@{}}
		\includegraphics[width=0.35\linewidth]{figures/r-batchstdev-l1_bs32_epoch=10} &
		\includegraphics[width=0.35\linewidth]{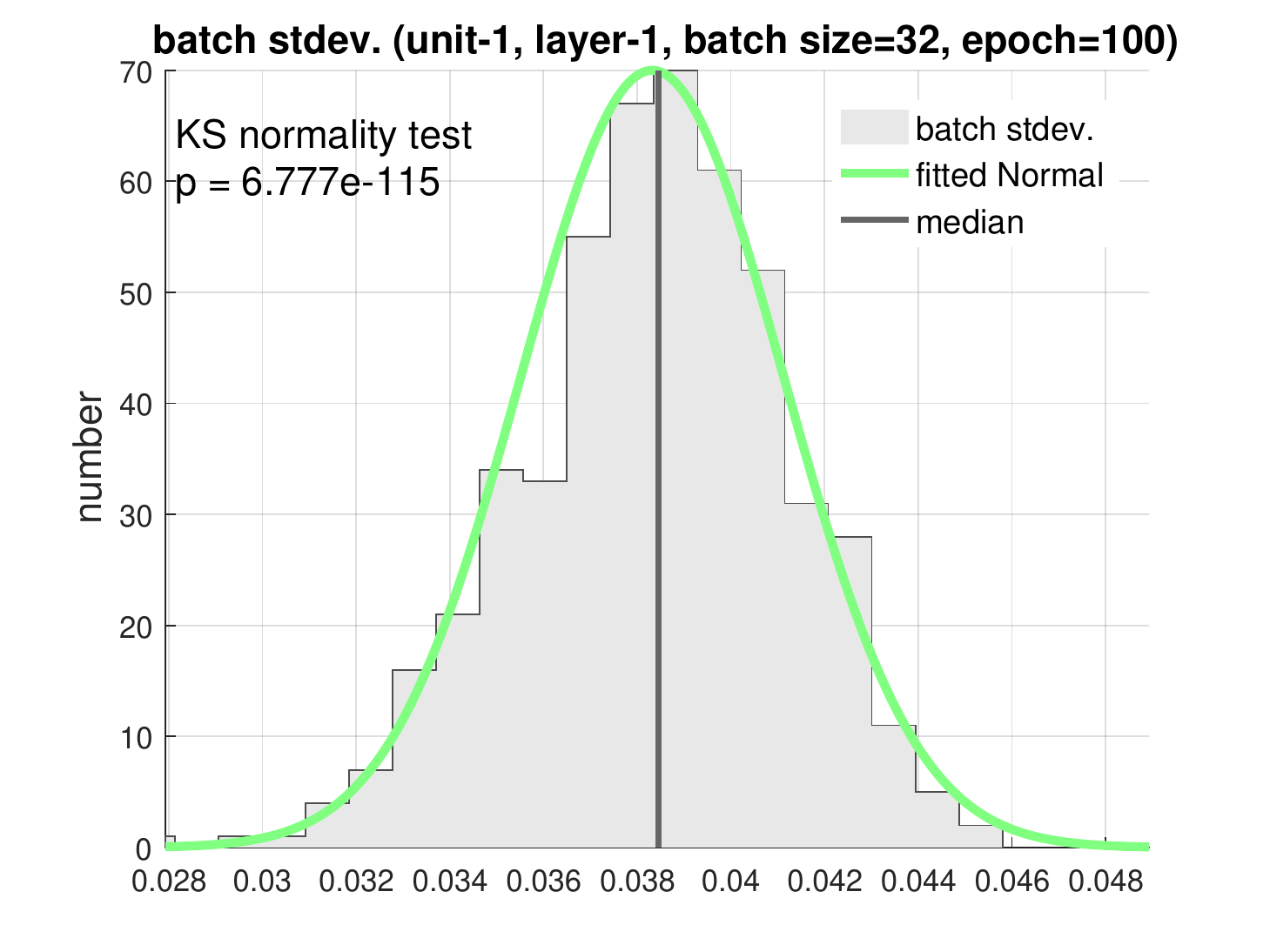} \\
		\includegraphics[width=0.35\linewidth]{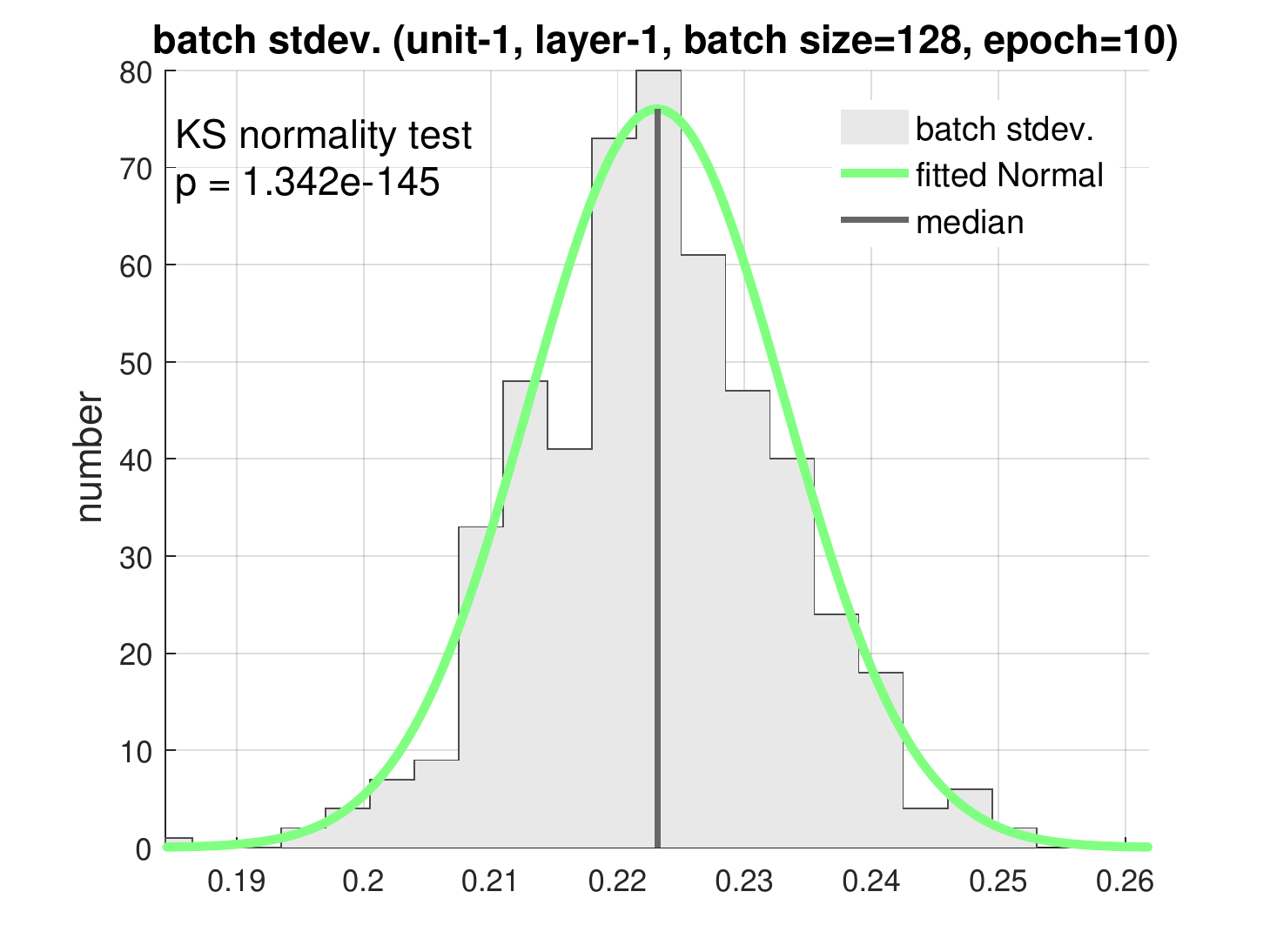} &
		\includegraphics[width=0.35\linewidth]{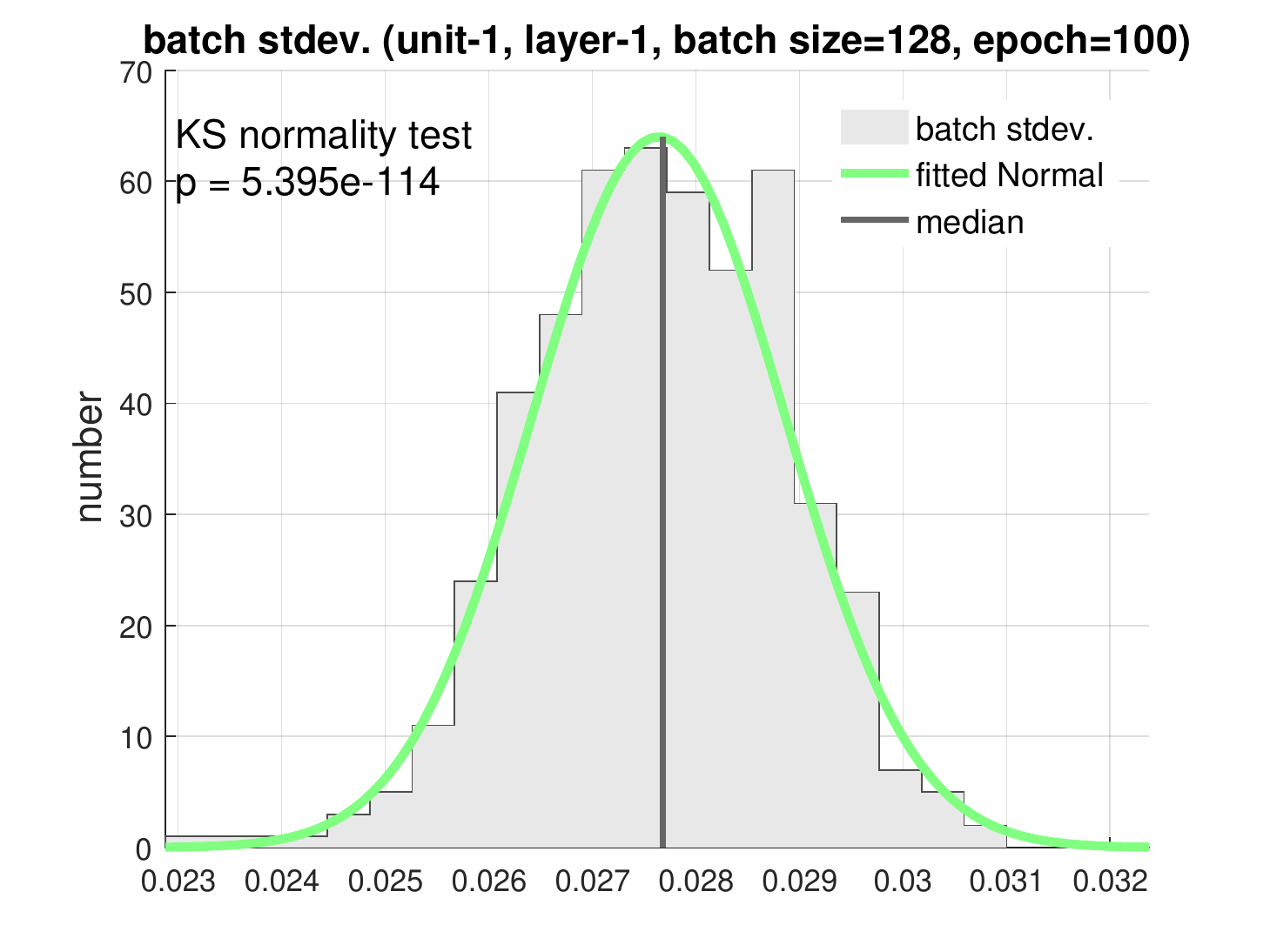} \\
		\vspace{2mm}
		\includegraphics[width=0.35\linewidth]{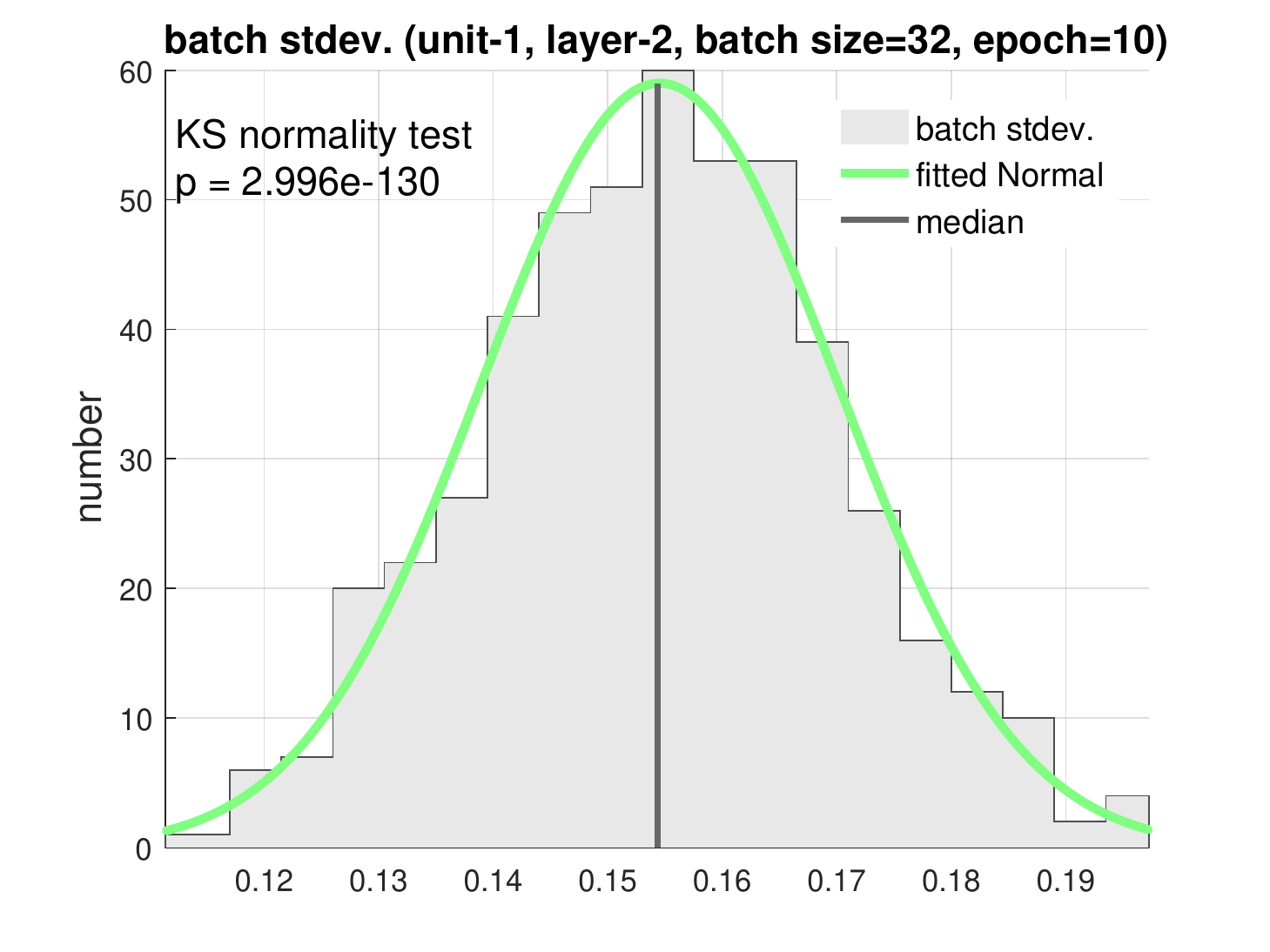} &
		\includegraphics[width=0.35\linewidth]{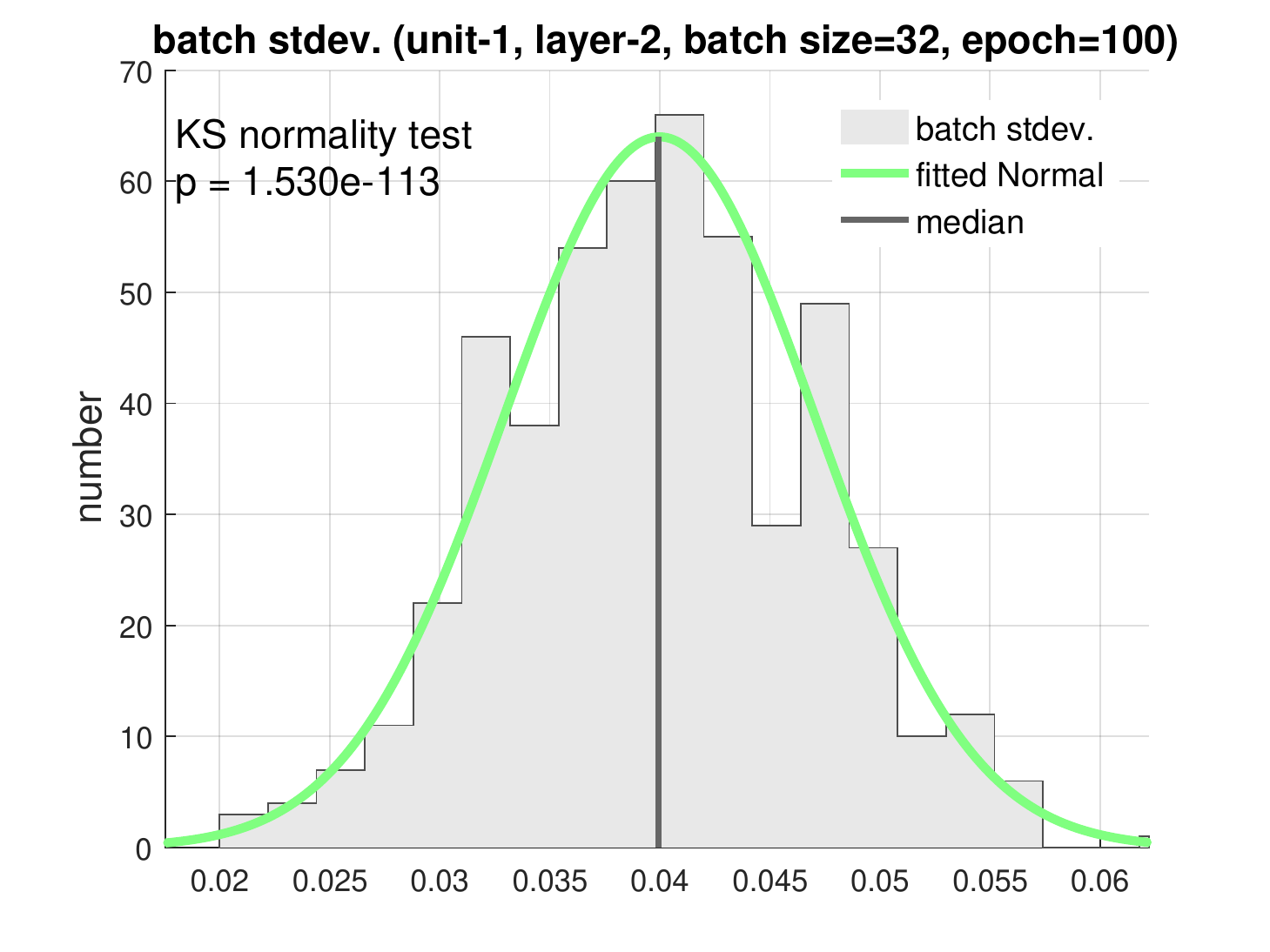} \\
		\includegraphics[width=0.35\linewidth]{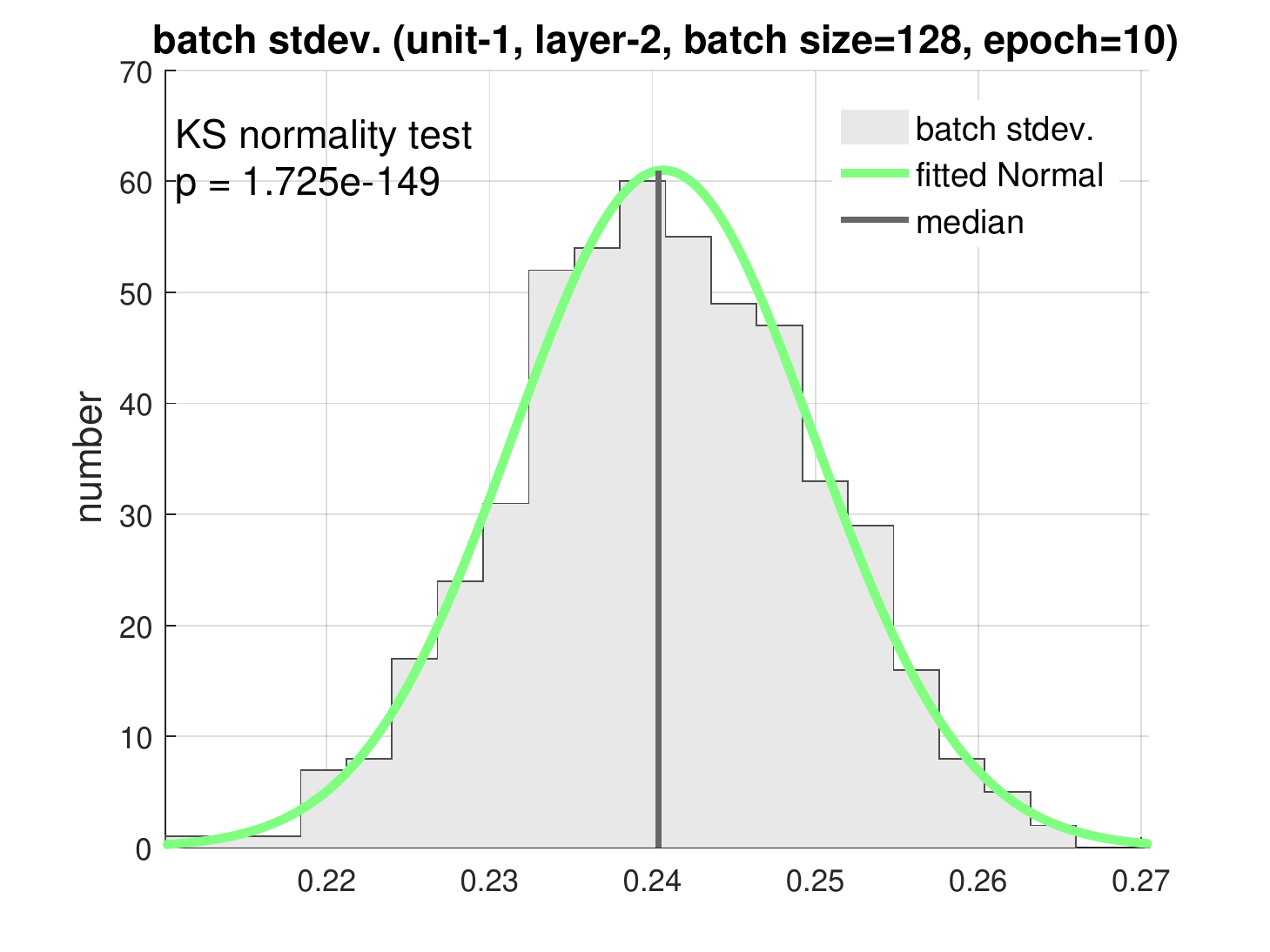} &
		\includegraphics[width=0.35\linewidth]{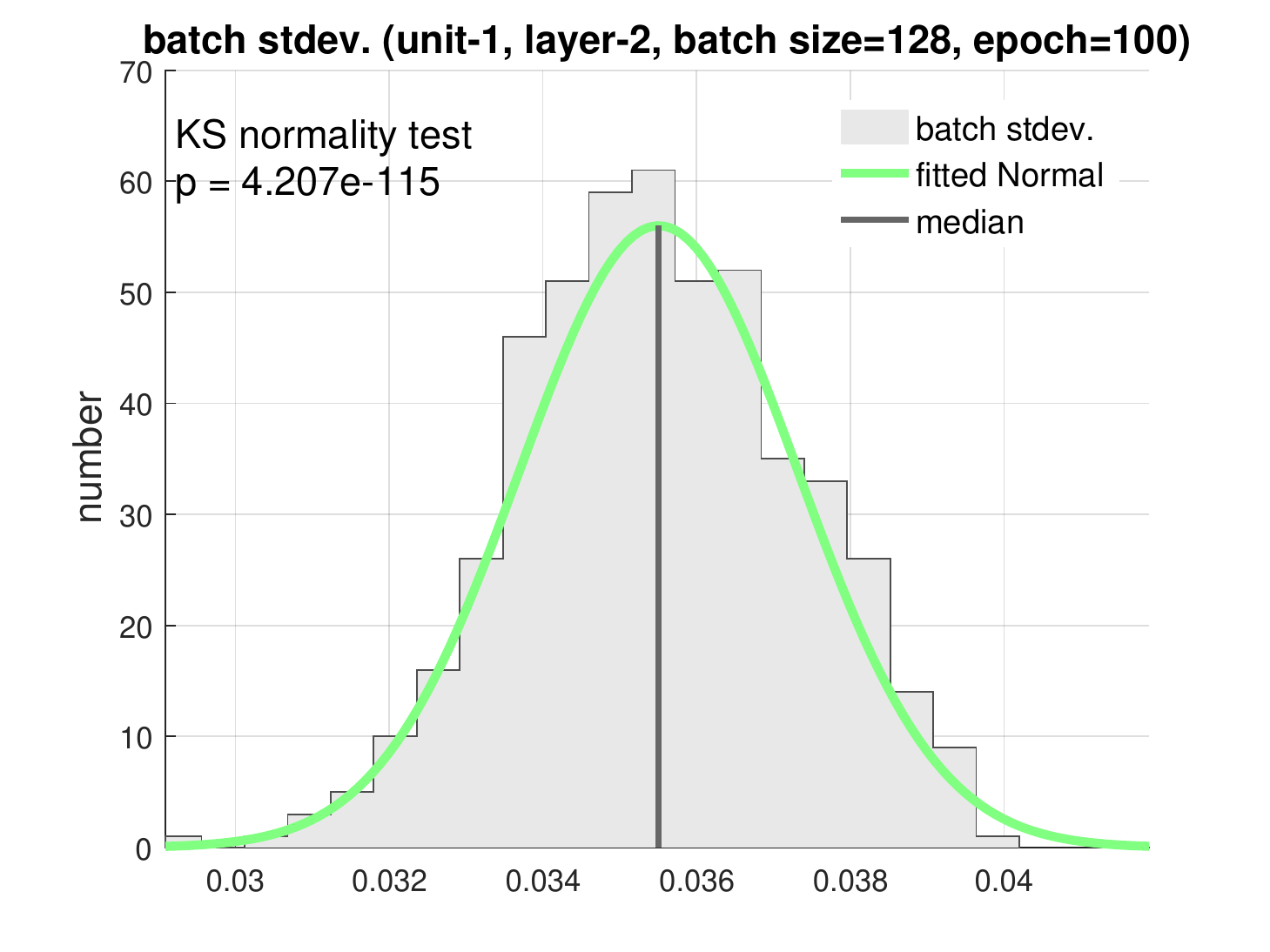} \\
	\end{tabular}
	\vspace{-3mm}
	\caption{\textbf{The distribution of standard deviation} of mini-batches during training of one of our datasets. The distribution closely follows our analytically approximated Gaussian distribution. The data is collected for one unit of each layer and is provided for different epochs and for different batch sizes.}
	\label{fig:batchstdev}
	\vspace{-3mm}
\end{figure}

In Figure \ref{fig:batchmeans} and Figure \ref{fig:batchstdev}, we provide statistics on the batch normalization parameters used for training. The plots show the distribution of BN mean and BN variance over different mini-batches of an actual training of Yacht dataset for one unit in the first hidden layer and the second hidden layer. Data is provided for different epochs and for different batch sizes.

\clearpage
\bibliography{mcbn_icml}
\bibliographystyle{icml2018}

\end{document}